\documentclass[sigconf,screen]{acmart}        % This is the version on Arxiv
\setcopyright{none}
\renewcommand\footnotetextcopyrightpermission[1]{}

\usepackage[utf8]{inputenc}
\usepackage{xcolor}
\usepackage{url}

\usepackage{amsmath}
\usepackage{booktabs}
\usepackage{tabularx}
\usepackage{cleveref}
\usepackage{color}
\usepackage{graphics}
\usepackage{graphicx}
\usepackage{enumitem}
\usepackage{balance}
\usepackage{xspace}
\usepackage{multirow}
\usepackage{array}
\usepackage{mathtools}

\usepackage{footnotehyper}      % for clicking the footnote link. 
\makesavenoteenv{tabular}
\makesavenoteenv{table*}
\usepackage{footnotebackref}
\usepackage{hyperref}
\usepackage{longtable}

\hypersetup{
    colorlinks=true,
    linkcolor=blue,
    filecolor=magenta,
    urlcolor=cyan,
    breaklinks=true
}
\definecolor{amber}{rgb}{1.0, 0.75, 0.0}
\newcommand{\topic}[1]{}%{\color{amber} Topic: #1}}
\newcommand{\vbarmc}[2]{\multicolumn{#1}{|c}{#2}}
\newcommand{\functionname}[1]{\texttt{#1}}
\newcommand{\norm}[1]{\left\|#1\right\|_1}
\newcommand{\twonorm}[1]{\|#1\|_2^2}
\newcommand{\papers}{more than 350\xspace}
\newcommand{\progress}[0]{\vspace{4pt}\textcolor{blue}{\textbf{Progress: }} }

\newtheorem{challenge}{Research Challenge}
\newcommand{\RCref}[1]{\textbf{RC\ref{#1}}}

\title{Counterfactual Explanations and Algorithmic Recourses for Machine Learning: A Review}

\author{Sahil Verma}
\email{vsahil@cs.washington.edu}
\affiliation{%
 \institution{University of Washington}
 \city{Seattle}
 \state{WA}
 \country{USA}}

\author{Varich Boonsanong}
\email{varicb@cs.washington.edu}
\affiliation{%
 \institution{University of Washington}
 \city{Seattle}
 \state{WA}
 \country{USA}}

 \author{Minh Hoang}
\email{minh257@cs.washington.edu}
\affiliation{%
 \institution{University of Washington}
 \city{Seattle}
 \state{WA}
 \country{USA}}

 \author{Keegan E. Hines}
\email{keegan@arthur.ai}
\affiliation{%
 \institution{Arthur AI}
 \city{Washington D.C.}
 \country{USA}}

 \author{John P. Dickerson}
\email{john@arthur.ai}
\affiliation{%
 \institution{Arthur AI}
 \city{Washington D.C.}
 \country{USA}}

 \author{Chirag Shah}
\email{chirags@uw.edu}
\affiliation{%
 \institution{University of Washington}
 \city{Seattle}
 \state{WA}
 \country{USA}}

\begin{document}

\begin{abstract}
%Machine learning is increasingly taking hold of life-impacting decisions. Therefore, it is essential to hold machine learning models accountable, and explainable to the people it is affecting. \jpd{We don't need either of these sentences.}
Machine learning plays a role in many deployed decision systems, often in ways that are difficult or impossible to understand by human stakeholders.  Explaining, in a human-understandable way, the relationship between the input and output of machine learning models is essential to the development of trustworthy machine learning based systems.  A burgeoning body of research seeks to define the goals and methods of \emph{explainability} in machine learning.  In this paper, we seek to review and categorize research on \emph{counterfactual explanations}, a specific class of explanation that provides a link between what could have happened had input to a model been changed in a particular way.  Modern approaches to counterfactual explainability in machine learning draw connections to the established legal doctrine in many countries, making them appealing to fielded systems in high-impact areas such as finance and healthcare.  Thus, we design a rubric with desirable properties of counterfactual explanation algorithms and comprehensively evaluate all currently proposed algorithms against that rubric.  Our rubric provides easy comparison and comprehension of the advantages and disadvantages of different approaches and serves as an introduction to major research themes in this field.  We also identify gaps and discuss promising research directions in the space of counterfactual explainability. 

\end{abstract}

\maketitle

\section{Introduction}
\label{sec:intro}

% Counterfactual Explanations $\subset$ Explanations $\subset$ FATE $\subset$ ML

Machine learning is increasingly accepted as an effective tool to enable large-scale automation in many domains. In lieu of hand-designed rules, algorithms are able to learn from data to discover patterns and support decisions. %Its widespread applicability has garnered the interest of academia and industry alike. 
Those decisions can, and do, directly or indirectly impact humans; high-profile cases include applications in credit lending~\cite{credit-risk-ml}, talent sourcing~\cite{hiring-ml}, parole~\cite{parole-ml}, and medical treatment~\cite{medical-treatment-ml}.
%Due to the life-impacting ability of such decisions, it is of utmost importance to ensure that the decision process is ethically sound.  \jpd{I'd prefer to keep `ethics' out of this -- that's a very different community}
%
The nascent Fairness, Accountability, Transparency, and Ethics (FATE) in machine learning community has emerged as a multi-disciplinary group of researchers and industry practitioners interested in developing techniques to detect bias in machine learning models, develop algorithms to counteract that bias, generate human-comprehensible explanations for the machine decisions, hold organizations responsible for unfair decisions, etc.   %\jpd{Stopping here for the night -- still not totally sure about the intro, but it definitely needs work.}

Human-understandable explanations for machine-produced decisions are advantageous in several ways. For example, focusing on a use case of applicants applying for loans, the benefits would include:   %\jpd{``In the case of an applicant participating using [a particular system..]'', ...}  
\begin{itemize}[leftmargin=10pt]
    \item An explanation can be beneficial to the applicant whose life is impacted by the decision. For example, it helps an applicant understand which of their attributes were strong drivers in determining a decision.
    \item Various forms of explanations can serve as a proxy for transparency in the system, which could increase its trustworthiness.
    \item Further, it can help an applicant challenge a decision if they feel an unfair treatment has been meted out, e.g., if one's race was crucial in determining the outcome. This can also be useful for organizations to check for bias in their algorithms.  
    \item In some instances, an explanation provides the applicant with feedback that they can act upon to receive the desired outcome at a future time.
    \item Explanations can help the machine learning model developers identify, detect, and fix bugs and other performance issues. 
    \item Explanations help adhere to laws surrounding machine-produced decisions, e.g., GDPR~\cite{GDPR}. 
    % \item \jpd{Something about the law? e.g. right to explanation}
\end{itemize}

Explainability in machine learning is broadly about using inherently interpretable and transparent models or generating post-hoc explanations for opaque models. 
Examples of the former include linear/logistic regression, decision trees, rule sets, etc. Examples of the latter include random forests, support vector machines (SVMs), and neural networks. Post-hoc explanation approaches can either be model-specific or model-agnostic. 
Explanations by feature importance and model simplification are two broad kinds of model-specific approaches. 
Model-agnostic approaches can be categorized into visual explanations, local explanations, feature importance, and model simplification. 

\vspace{5pt}

Feature importance finds the most influential features contributing to the model's overall accuracy or for a particular decision, e.g., SHAP~\cite{shap-paper}, QII~\cite{QII_SA1}. 
Model simplification finds an interpretable model that imitates the opaque model closely. 
Dependency plots are a popular kind of visual explanation, e.g., Partial Dependence Plots~\citep{intro_PDP}, Accumulated Local Effects Plot~\citep{intro_ALE}, Individual Conditional Expectation~\citep{ICE_PDP1}. 
They plot the change in the model's prediction as one or multiple features are changed. 
Local explanations differ from other methods because they only explain a single prediction. 
Local explanations can be further categorized into approximation and example-based approaches. 
Approximation approaches sample new datapoints in the vicinity of the datapoint whose prediction from the model needs to be explained (hereafter called the explainee datapoint), and then fit a linear model (e.g., LIME~\citep{ribeiro_why_2016}) or extracts a rule set from them (e.g., Anchors~\citep{Ribeiro2018Anchors}).
Example-based approaches seek to find datapoints in the vicinity of the explainee datapoint. 
They either offer explanations in the form of datapoints that have the same prediction as the explainee datapoint or the datapoints whose prediction differs from the explainee datapoint. 
Note that the latter kind of datapoints are still close to the explainee datapoint and are termed as ``counterfactual explanations'' (CFE).

\vspace{5pt}

Recall the use case of applicants applying for a loan. For an individual whose loan request has been denied, counterfactual explanations provide them with \emph{actionable} feedback that could help them make changes to their features in order to transition to the desirable side of the decision boundary, i.e., get the loan. This feedback is termed as an \emph{algorithmic recourse}. 
Unlike several other explainability techniques, CFEs (or recourses) do not explicitly answer the ``why'' the model made a prediction; instead, they provide suggestions to achieve the desired outcome. 
CFEs are also applicable to black-box models (when only the \functionname{predict} function of the model is accessible), and therefore place no restrictions on model complexity and do not require model disclosure. 
They also do not necessarily approximate the underlying model, producing accurate feedback. 
Owing to their intuitive nature, CFEs are also amenable to legal frameworks (see~\cref{sec:legal}). 
% \jpd{amenable to legal frameworks? Generally applicable? something other than success} 
% Due to increasing legal frameworks (see~\cref{sec:legal}), focus on ethical aspects of machine learning, and general curiosity to better understand machines, counterfactual explanations have gained the attention of academics and industry alike. 

% Due to the recent GDPR regulation~\citep{GDPR} which mandates explainable decisions in case of full or partial automation, counterfactual explanations has gained the attention of academics and industry alike. 

In this work, we collect, review and categorize \papers %\footnote{If we have missed some work, please contact us to add it to this review paper.} 
recent papers that propose algorithms to generate counterfactual explanations for machine learning models. Many of these methods have focused on datasets that are either tabular or image-based. 
We describe our methodology for collecting papers for this survey in~\cref{sec:method}. 
We describe recent research themes in this field and categorize the collected papers among a fixed set of desiderata for effective counterfactual explanations (see~\cref{tab:main-table}).

The contributions of this review paper are:
\begin{enumerate}
    \item We examine a set of \papers recent papers on the same set of parameters to allow for an easy comparison of the techniques these papers propose and the assumptions they work under. 
    \item The categorization of the papers achieved by this evaluation helps a researcher or a developer choose the most appropriate algorithm given the set of assumptions they have and the speed and quality of the generation they want to achieve. 
    \item Comprehensive and lucid introduction for beginners in the area of counterfactual explanations for machine learning. 
\end{enumerate}

% Say something about how CEs are increasingly important due to the legal requirements of deploying automated decision.

% Comments: Add that we are going to view it through the lens of - add in the setting for a loan, or somebody up for parole. 
\section{Background}
\label{sec:background}
% write the motivation here
This section gives the background about the social implications of machine learning, explainability research in machine learning, and some prior studies about counterfactual explanations.

\subsection{Social Implications of Machine Learning}
Establishing fairness and making an automated tool's decision explainable are two broad ways in which we can ensure equitable social implications of machine learning. 
Fairness research aims at developing algorithms that can ensure that the decisions produced by the system are not biased against a particular demographic group of individuals, which are defined with respect to sensitive or protected features, such as race, sex, and religion.
Anti-discrimination laws make it illegal to use sensitive features as the basis of any decision (see \Cref{sec:legal}). Biased decisions can also attract widespread criticism and are therefore crucial to avoid~\cite{prime-racist,apple-sexist}. 
Fairness has been captured in several notions based on a demographic grouping or individual capacity. 
\citet{verma_fairness} have enumerated and intuitively explained many fairness definitions using a unifying dataset. 
\citet{dunkelau_fairness-aware} provide an extensive overview of the major categorization of research efforts in ensuring fair machine learning and enlists important works in all categories. 
Explainable machine learning has also seen interest from other communities, specifically healthcare~\citep{tjoa2019survey1}, having huge social implications. 
Several works have summarized and reviewed other research in explainable machine learning~\citep{xai-survey2,carvalho2019:survey3,xai-survey4}. 

\subsection{Explainability in Machine Learning}
% \Kee{there's a weird transition here into the next section. Maybe we need as section 2.3 with an XAI overview, of which CF is but one part. But that might be redundant with a similar paragraph in a preceding section.}
% As the introduction section already outlines the importance of explainability in AI and its broad categorizations, 
This section gives some concrete examples that emphasize the importance of explainability and give further details of the research in this area. 
In a real-world example, the US military trained a classifier to distinguish enemy tanks from friendly tanks. Although the classifier performed well on the training and test dataset, its performance was abysmal on the battlefield. 
Later, it was found that the photos of friendly tanks were taken on sunny days, while for enemy tanks, photos clicked only on overcast days were available~\citep{xai-survey4}. 
The classifier found it much easier to use the difference between the background as the distinguishing feature. 
In a similar case, a husky was classified as a wolf because of the presence of snow in the background, which the classifier had learned as a feature associated with wolves~\citep{ribeiro_why_2016}. 
The use of an explainability technique helped discover these issues. 
% Had an explainability technique been used to understand the working of the classifier, the problems on the battlefield could have been averted. 

The explainability problem can be divided into model explanation and outcome explanation problems~\citep{xai-survey4}. %and model inspection problems. 

\emph{Model explanation} searches for an interpretable and transparent global explanation of the original model. 
% The interpretable model should have high fidelity to the original model and be understandable to humans, such as linear models, rule sets, and decision trees. 
Various papers have developed techniques to explain neural networks and tree ensembles using single decision tree~\citep{craven_exp1,KRISHNAN_exp2,Pedro_exp4} and rule sets~\citep{Deng_exp5,Andrews_exp6}.
Some approaches are model-agnostic, such as Golden Eye and  PALM~\citep{Andreas_exp7,Krishnan_exp8,Zien_exp9}. 

\emph{Outcome explanation} needs to provide an explanation for a specific prediction from the model. 
This explanation need not be a global explanation or explain the internal logic of the model. 
Model-specific approaches for deep neural networks (CAM, Grad-CAM~\citep{Khosla_cam,grad-cam}), and model agnostic approaches (LIME, MES~\citep{ribeiro_why_2016,Turner2016_MES}) have been proposed. 
These are either feature attribution or model simplification methods. 
Example-based approaches are another kind of explainability technique used to explain a particular outcome. 
This work focuses on counterfactual explanations (CFEs), which is an example-based approach. 

By definition, CFEs are applicable to supervised machine learning setups where the desired prediction has not been obtained for a datapoint. 
The majority of research in this area has applied CFEs to classification settings, which consists of several labeled datapoints that are given as input to the model, and the goal is to learn a function mapping from the input datapoints (with, say, m features) to labels. 
In classification, the labels are discrete values. 
$\mathcal{X}^m$ is used to denote the input space of the features, and $\mathcal{Y}$ is used to denote the output space of the labels. 
The learned function is the mapping $f: \mathcal{X}^m \to \mathcal{Y}$, which is used to predict labels for unseen datapoints in the future. 

% This can be further categorized into classification and regression., whereas in regression the labels are continuous values. 

% Fairness-y stuff
% Explainability vs interpretability vs transparency vs ...
\vspace{-2pt}
\subsection{History of Counterfactual Explanations}

Counterfactual explanations have a long history in other fields like philosophy, psychology, and the social sciences. Philosophers like David Lewis published articles on the ideas of counterfactuals back in 1973~\citep{Lewis1973:phil2}. 
\citet{Woodward2003:phil4} said that a satisfactory explanation must follow patterns of counterfactual dependence. 
Psychologists have demonstrated that counterfactuals elicit causal reasoning in humans~\citep{Byrne:psycho1,Byrne2019:psycho2,Kahneman1986:psycho3}. 
Philosophers have also validated the concept of causal thinking due to counterfactuals~\citep{VanFraassenBas1980:phil5,Woodward2003:phil4}. 

%Counterfactual explanations are surfacing as a popular means of providing explanations to the receivers of a machine generated decision. % \jpd{Stronger intro.  Something about the legal right to explanation.}

Studies have compared the likeability of CFEs with other explanation approaches. 
\citet{Binns:2018} and \citet{Dodge-explaining:2019} performed user studies that showed that users prefer CFEs over case-based reasoning, which is another example-based approach. 
% This was also corroborated in the user study performed by . \todo{expand}
The work by \citet{fernandez-loria_explaining_2020} provides three interesting examples where the feature importance explanation methods fail to capture the underlying model, whereas CFEs do. 
\citet{cfe-fair-adequate-Asher} argue that the partiality and locality of CFEs make them epistemically accessible and an adequate form of explanations. 

%\jpd{IMO this entire para should be in prelims or relwork or something, maybe move the para elsewhere, condense into one sentence and include in the intro as well.} 

\section{Counterfactual Explanations}
This section illustrates counterfactual explanations by giving an example and then outlines the major aspects of the problem. 

\subsection{An Example}
Suppose Alice walks into a bank and seeks a home mortgage loan. The decision is impacted in large part by a machine learning classifier that considers Alice's feature vector of \{\emph{Income}, \emph{CreditScore}, \emph{Education}, \emph{Age}\}. Unfortunately, Alice is denied the loan she seeks and is left wondering (1) why the loan was denied? and (2) what can she do differently so that the loan will be approved in the future? The former question might be answered with explanations like: ``CreditScore was too low'', and is similar to the majority of traditional explainability methods. The latter question forms the basis of a \emph{counterfactual explanation}: what small changes could be made to Alice's feature vector in order to end up on the other side of the classifier's decision boundary? Let us suppose the bank provides Alice with exactly this advice (through a CFE) of what she might change in order to be approved next time. 
A possible counterfactual recommended by the system might be to increase her \emph{Income} by $\$10$K or get a new master's degree or a combination of both. The answer to the former question does not tell Alice what action to take, while the CFE explicitly helps her. 
\Cref{fig:cf_fig} illustrates how the datapoint representing an individual, which originally got classified in the negative class, can take two paths to cross the decision boundary into the positive class region. 

The assumption in a CFE is that the underlying classifier would not change when the applicant applies in the future. And if the assumption holds, the counterfactual guarantees the desired outcome in the future time. 

     % maintained modularity
\subsection{Desiderata and Major Themes of Research}
\label{sec:themes}
% \topic{In this section, we'll hit on a small subset of papers and try to give a pedagogical overview of this field and the primary developments. I imagine this section starts with the Wachter paper, defines the loss function, and then proceeds by adding more and more desirable properties of counterfactuals, properties such as sparsity, data manifold, actionability, causality.}

The previous example alludes to many desirable properties of an effective counterfactual explanation. For Alice, the counterfactual should quantify a relatively small change, which will lead to the desired alternative outcome. Alice might need to increase her income by $\$10$K to get approved for a loan, and even though an increase of $\$50$K would do the job, it is most pragmatic for her if she can make the smallest possible change. Additionally, Alice might care about a simpler explanation -- it is easier for her to focus on changing a few things (such as only \emph{Income}) instead of trying to change many features. Alice certainly also cares that the counterfactual she receives is giving her advice, which is realistic and actionable. It would be of little use if the recommendation were to decrease her age by ten years. 

These desiderata, among others, have set the stage for recent developments in the field of counterfactual explainability. As we describe in this section, major themes of research have sought to incorporate increasingly complex constraints on counterfactuals, all in the spirit of ensuring the resulting explanation is truly actionable and helpful. 
% These constraints include sparsity, closeness, realism (similarity to other data), causality, and more. 
Development in this field has focused on addressing these desiderata in a way that is generalizable across algorithms and is computationally efficient. 

\belowcaptionskip=-3pt
\abovecaptionskip=3pt
\begin{figure}
    \centering
    \includegraphics[width=0.7\columnwidth]{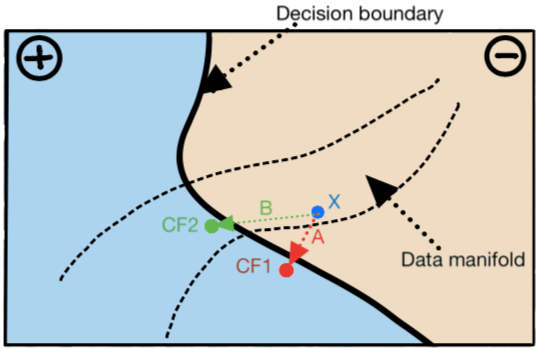}
    \caption{Two possible paths for a datapoint (shown in \textcolor{blue}{blue}), originally classified in the negative class, to cross the decision boundary. The endpoints of both the paths (shown in \textcolor{red}{red} and \textcolor{green}{green}) are valid counterfactuals for the original point. Note that the \textcolor{red}{red} path is the shortest, whereas the \textcolor{green}{green} path adheres closely to the manifold of the training data, but is longer.}
    \label{fig:cf_fig}
\end{figure}

\begin{enumerate}[leftmargin=*]
    \item \textit{Validity}: \citet{wachter_counterfactual_2017} first proposed counterfactual explanations in 2017. They posed CFE as an optimization problem. 
    \Cref{eq:cfe1_primal} states the optimization objective, which is to minimize the distance between the counterfactual ($x^{\prime}$) and the original datapoint ($x$) subject to the constraint that the output of the classifier on the counterfactual is the desired label ($y^{\prime} \in \mathcal{Y}$). Converting the objective into a differentiable, unconstrained form yields two terms (see \Cref{eq:cfe1}). 
    The first term encourages the output of the classifier on the counterfactual to be close to the desired class, and the second term forces the counterfactual to be close to the original datapoint. A metric $d$ is used to measure the distance between two datapoints $x, x^{\prime} \in \mathcal{X}$, which can be the L1/L2 distance, or quadratic distance, or distance functions which take as input the CDF of the features~\citep{Ustun19:Actionable}, or pairwise feature costs as perceived by users~\citep{hima-beyond-recourse-globalcfe}. Thus, this original definition already emphasized that an effective counterfactual must be \emph{small change} relative to the starting point. 
    
    \begin{equation}
        {\mathrm{arg \ }}\underset{x^{\prime}}{\mathrm{min \ }} d(x, x^{\prime}) \text{ subject to } f(x^{\prime}) = y^{\prime}
        \label{eq:cfe1_primal}
    \end{equation}
    
    \begin{equation}
        {\mathrm{arg \ }}\underset{x^{\prime}}{\mathrm{min \ }}\underset{\lambda}{\mathrm{max \ }} \lambda(f(x^{\prime}) - y^{\prime})^2 + d(x, x^{\prime})
        \label{eq:cfe1}
    \end{equation}
    % where $\lambda$ is used for trade-off between the two terms and $d(x, x^{\prime})$ is any distance metric between and input $x$ and a potential counterfactual $x^{\prime}$. 
    A counterfactual that indeed is classified in the desired class is a valid counterfactual. As illustrated in \cref{fig:cf_fig}, the points shown in \textcolor{red}{red} and \textcolor{green}{green} are valid counterfactuals, as they are in the positive class region. The distance to the \textcolor{red}{red} counterfactual is smaller than the distance to the \textcolor{green}{green} counterfactual. 
    
    \item \textit{Actionability}: An important consideration while making a recommendation is about which features are mutable (e.g., income, age) and which are not (e.g., race, country of origin).
    A recommended counterfactual should never change the immutable features. In fact, if a change to a legally sensitive feature produces a change in prediction, it shows inherent bias in the model. Several papers have also mentioned that an applicant might have a preference order amongst the mutable features (which can also be hidden.) The optimization problem is modified to take this into account. We might call the set of actionable features $\mathcal{A}$, and update our loss function to be,
    \begin{equation}
        {\mathrm{arg \ }}\underset{x^{\prime }\in \mathcal{A}}{\mathrm{min \ }}\underset{\lambda}{\mathrm{max \ }} \lambda(f(x^{\prime}) - y^{\prime})^2 + d(x, x^{\prime})
        \label{eq:cfe2}
    \end{equation}
    
    \item \textit{Sparsity}: There can be a trade-off between the number of features changed and the total amount of change made to obtain the counterfactual. A counterfactual ideally should change a smaller number of features in order to be the most effective. It has been argued that people find it easier to understand shorter explanations~\citep{Miller-xai:2019,naumann2021consequenceaware}, making sparsity an important consideration. We update our loss function to include a penalty function that encourages sparsity in the difference between the modified and the original datapoint, $g(x^{\prime} - x)$, e.g., L0/L1 norm.   
    \begin{equation}
        {\mathrm{arg \ }}\underset{x^{\prime }\in \mathcal{A}}{\mathrm{min \ }}\underset{\lambda}{\mathrm{max \ }} \lambda(f(x^{\prime}) - y^{\prime})^2 + d(x, x^{\prime}) + g(x^{\prime} - x)
        \label{eq:cfe3}
    \end{equation}
    
    \item \textit{Data Manifold closeness}: It would be hard to trust a counterfactual if it resulted in a combination of features that were utterly unlike any observations the classifier has seen before. In this sense, the counterfactual would be ``unrealistic", not easy to realize, and anomalous to the training datapoints \citep{Brown_Talbert_cfe-anomalous}. Therefore, a generated counterfactual should be realistic in the sense that it is near the training data and adheres to observed correlations among the features. Many papers have proposed various ways of quantifying this. We might update our loss function to include a penalty for adhering to the data manifold defined by the training set $\mathcal{X}$, denoted by $l(x^{\prime} ; \mathcal{X})$
    % \small
    \begin{equation}
    % \begin{small}
        % \scalebox{0.9\columnwidth}{
        \resizebox{.85\columnwidth}{!}{$ {\mathrm{arg \ }}\underset{x^{\prime }\in \mathcal{A}}{\mathrm{min \ }}\underset{\lambda}{\mathrm{max \ }} \lambda(f(x^{\prime}) - y^{\prime})^2 + d(x, x^{\prime}) + g(x^{\prime} - x) + l(x^{\prime} ; \mathcal{X}) $ }
        \label{eq:cfe4}
    % \end{small}
    \end{equation}
    % \normalsize
    In \cref{fig:cf_fig}, the region between the dashed lines shows the data manifold. There are two possible paths to cross the decision boundary for the \textcolor{blue}{blue} datapoint. The shorter, red path takes it to a counterfactual that is outside the data manifold, whereas a bit longer, the green path takes it to a counterfactual that follows the data manifold. Adding the data manifold loss term encourages the algorithm to choose the green path over the red path, even if it is slightly longer. 
    
    \item \textit{Causality}: Features in a dataset are rarely independent, therefore, changing one feature in the real world affects other features. For example, getting a new educational degree necessitates increasing the individual's age by at least some amount. In order to be realistic and actionable, a counterfactual should maintain any known causal relations between features. 
    % Another important consideration is to encode any strong prior knowledge we might have about the features. Specifically, causal relations between features should be taken into account while generating counterfactuals.
    % The second term in \cref{eq:cfe1} is modified to account for such relations.
    % We add the loss term for causal relations in the optimization objective. 
    Generally, our loss function now accounts for (1) counterfactual validity, (2) sparsity in feature vector (and actionability of features); (3) similarity to the training data; and (4) causal relations. 
    \end{enumerate}
    
    The following research themes are not added as terms in the optimization objective; they are properties of the algorithm generating the CFEs. 
    
    \begin{enumerate}[resume]
    \item \textit{Amortized inference}: Generating a counterfactual is expensive, which involves solving an optimization process for each datapoint. 
    \citet{mahajan_preserving_2020} proposed generative technique for ``amortized inference'' of CFEs. Learning to predict a CFE allows the algorithm to quickly compute a counterfactual (or several) for any new input $x$, without requiring to solve an optimization problem. \citet{verma2021amortized} proposed another approach that uses RL to generate amortized CFEs. 
    
    \item \textit{Black-box access}: If a CFE generating approach can work with the black-box access to an ML model, i.e., with only accessing its `predict' function, it can then be used in settings where the access to the ML model cannot be given due to proprietary or legal reasons. \citet{dandl_multi-objective_2020} propose a genetic algorithm and \citet{verma2021amortized} propose a RL-based algorithm to this end. 
    
    \item \textit{Model Agnosticity}: A closely linked concept is model agnosticity. An approach that is model agnostic can work with different kinds of ML models and hence is more desirable than a model-specific approach. An approach that requires black-box access to the model is model-agnostic by definition. 
    
    % \item \textit{Alternative methods}: Finally, several papers solve the counterfactual generation problem using linear programming, mixed-integer programming, Answer Set Programming, or SMT solvers. These approaches give guarantees and optimize fast but are limited to classifiers with linear (or piece-wise linear) structures. 
    
\end{enumerate}

\subsection{Relationship to other related terms} 

Out of the papers collected, different terminology often captures the basic idea of counterfactual explanations, although subtle differences exist between the terms. Several terms worth noting include:
\begin{itemize}[leftmargin=*]
    \item \emph{Algorithmic Recourse}: \citet{Ustun19:Actionable} point out that counterfactuals do not take into account the actionability of the prescribed changes, which recourse does. Works taking a causal view of the problem further fortify this claim \citep{karimi_algorithmic_2020,karimi-imperfect:2020}. 
    Recent papers in counterfactual generation take actionability and feasibility of the prescribed changes, and therefore the difference with recourse has blurred. In this work, we use the term counterfactual explanation, its abbreviation CFE, and recourse interchangeably. 
    
    \item \emph{Inverse classification}: Inverse classification aims to perturb an input in a meaningful way in order to classify it into its desired class~\citep{inverse-classification1,inverse-classification2}. Such an approach prescribes the actions to be taken in order to get the desired classification. Therefore inverse classification has the same goals as CFEs. 
    
    \item \emph{Contrastive explanation}: Contrastive explanations generate explanations of the form ``an input x is classified as y because features $f_1, f_2,\dots, f_k$ are present and $f_n,\dots,f_r$ are absent''. 
    The features that are minimally sufficient for a classification are called {\em pertinent positives}, and the features whose absence is necessary for the final classification are termed {\em pertinent negatives}. 
    To generate both pertinent positives and pertinent negatives, one needs to solve the optimization problem to find the minimum perturbations needed to maintain the same class label or change it, respectively. Therefore contrastive explanations (specifically pertinent negatives) are related to CFEs. 
    
    \item \emph{Adversarial learning}: Adversarial learning is closely related, but the terms are not interchangeable. Adversarial learning aims to generate the least amount of change in a given input to classify it differently, often with the goal of far-exceeding the decision boundary and resulting in a highly-confident misclassification. While the optimization problem is similar to the one posed in a counterfactual generation, the desiderata are different. For example, in adversarial learning (often applied to images), the goal is an imperceptible change in the input image. This is often at odds with the CFE's goal of sparsity and parsimony (though single-pixel attacks are an exception). Further, notions of data manifold and actionability/causality are rarely considerations in adversarial learning. A few works point to the similarity and synergy between the two domains: 
    \citet{pawelczyk2021_CFE_AE_connection} explore the connection between the optimization objectives and results of the adversarial and CFE generating techniques.
    \citet{freiesleben2020CFE_AE1} state that the differences in the desired class label and distance from the original datapoint distinguish CFEs from adversarial examples.
    \citet{Elliott2021_adversarial_cfe_images} propose generating semantically meaningful adversarial perturbations to generate CFEs for images. 
    \citet{semantic-explanation_adversarial_cfe_diff} point out that the constraint of producing plausible datapoints distinguishes CFEs from adversarial examples. 
      
\end{itemize}

       % replaced by the tex file for modularity
\section{Assessment of the approaches on counterfactual properties}
\label{sec:table}

For easy comprehension and comparison, we identify several properties that are important for a counterfactual generation algorithm. 
For all the collected papers which propose an algorithm to generate counterfactual explanations, we assess the algorithm they propose against these properties. 
The results are presented in~\cref{tab:main-table}. 
For papers that do not propose new algorithms and discuss related aspects of counterfactual explanations or modifications to previous methods are mentioned in \cref{sec:other_works}. 
The methodology we used to collect the papers is given in~\cref{sec:method}.

\subsection{Properties of counterfactual algorithms}
This section expounds on the key properties of a counterfactual explanation generation algorithm. The properties form the columns of ~\cref{tab:main-table}. 
\begin{enumerate}[leftmargin=*]
    
    \item \emph{Model access}: The counterfactual generation algorithms require different levels of access to the underlying model for which they generate counterfactuals. We identify three distinct access levels -- access to complete model internals, access to gradients, and access to only the prediction function (\emph{black-box}). 
    Access to the complete model internals is required when the algorithm uses a solver-based method like, mixed integer programming~\citep{russell_efficient_2019,Ustun19:Actionable,karimi_model-agnostic_2020,karimi_algorithmic_2020,Kanamori2020:DACE} or if they operate on decision trees~\citep{Tolomei2017:Interpretable,fernandez-random:2020,lucic-actionable:2020, oblique-tree-cfe, scaling_Nearest_CFE} which requires access to all internal nodes of the tree.
    A majority of the methods use a gradient-based algorithm to solve the optimization objective, modifying the loss function proposed by~\citet{wachter_counterfactual_2017}, but this is restricted to differentiable models only. 
    % One constraint that comes with using this method is the restriction to differentiable models only. 
    % The third class of methods only require access to the prediction function of the model. 
    Black-box approaches use gradient-free optimization algorithms such as Nelder-Mead~\citep{grath_interpretable_2018}, growing spheres~\citep{medina_comparison-based_2018}, FISTA~\citep{dhurandhar_model_2019,van_looveren_interpretable_2020} ASP ~\citep{declarative-CFE}, or genetic algorithms~\citep{inverse-classification2,sharma_certifai_2019,dandl_multi-objective_2020} to solve the optimization problem. 
    Finally, some approaches do not cast the goal into an optimization problem and solve it using heuristics~\citep{guidotti_local_2018,rathi-generating:2019,white_measurable_2019,keane2020good}.
    \citet{poyiadzi_face_2020} propose FACE, which uses Dijkstra's algorithm~\citep{dijkstra1959} to find the shortest path between existing training datapoints to find counterfactual for a given input. Hence, this method does not generate new datapoints. \citet{Fraunhofer-pure-region-sampling} and \citet{Pierre-tree-ensemble-pure-region} divide the feature space into `pure' regions where all datapoints (by sampling) belong to one class and then use graph traversing techniques to find the closest CFEs.  
    
    Distinct from the three levels of model access, there exist approaches that propose new training routines. \citet{alexis-ross-training-methodology-cfe} propose adding adversarial loss during training of the ML model to have a higher probability of having a recourse for the training datapoints. (After training, any CFE generating method can be used.) 
    \citet{guo-cfe-counternet} propose CounterNet, a novel architecture that predicts the class and generates the CFE of a datapoint when trained from scratch. \citep{sum-product-networks-cfe} train a sum-product network that acts as both a classifier and density estimator and uses that to generate CFEs. 
    
    \item \emph{Model agnostic}: This column describes the domain of models a given algorithm can operate on. 
    For example, gradient-based algorithms can only handle differentiable models, and the algorithms based on solvers require linear or piece-wise linear models~\citep{russell_efficient_2019,Ustun19:Actionable,karimi_model-agnostic_2020,karimi_algorithmic_2020,Kanamori2020:DACE}, some algorithms are model-specific and only work for those models like tree ensembles~\citep{Tolomei2017:Interpretable,Kanamori2020:DACE,fernandez-random:2020,lucic-actionable:2020}. 
    Black-box methods have no restriction on the underlying model and are, therefore, model-agnostic. 

    \item \emph{Optimization amortization}:  
    Among the collected papers, the proposed algorithm mostly returned a single counterfactual for a given input datapoint.
    %Among all the papers we collected, we found that algorithms proposed by most of the papers, when given an input datapoint, returned a single counterfactual. 
    Therefore these algorithms require solving the optimization problem for each counterfactual that was generated, that too, for every input datapoint. 
    A smaller number of the methods are able to generate multiple counterfactuals (generally diverse by some metric of diversity) for a single input datapoint; therefore, they require to be run once per input to get several counterfactuals~\citep{guidotti_local_2018,russell_efficient_2019,sharma_certifai_2019,mothilal_explaining_2020,mahajan_preserving_2020,karimi_model-agnostic_2020,dandl_multi-objective_2020,fernandez-random:2020,oblique-tree-cfe}. 
    \citet{mahajan_preserving_2020}'s approach learns the mapping of datapoints to counterfactuals using a variational auto-encoder (VAE)~\citep{doersch-autoencoder}. 
    Therefore, once the VAE is trained, it can generate multiple counterfactuals for all input datapoints, without solving the optimization problem separately and is thus very fast. 
    \citet{verma2021amortized} and \citet{rl_cfe_approach2_amortized} train a reinforcement learning model to learn the actions that need to be taken to generate CFEs for a data distribution. Hence, these approaches are also amortized. 
    \citep{gan_cfe_amortized} trains a CGAN to synthesize CFEs with umbrella sampling; hence, their approach is also amortized. \citet{conditional_gan_cfe_looveren} also train a GAN-based model that is amortized. 
    \citet{schleich2021geco} partially evaluate (amortize) the classifier for the static features, hence speeding up the CFE generation. 
    We report two aspects of optimization amortization in the table. 
    \begin{itemize}%[leftmargin=0.3em]
        \item  \emph{Amortized Inference}: This column is marked \texttt{Yes} if the algorithm can generate counterfactuals for multiple input datapoints without optimizing separately for them; otherwise, it is marked \texttt{No}. 
        \item  \emph{Multiple counterfactuals (CF)}: This column is marked \texttt{Yes} if the algorithm can generate multiple counterfactuals for a single input datapoint; otherwise, it is marked \texttt{No}. 
    \end{itemize}

    \item \emph{Counterfactual (CF) attributes}: These columns evaluate algorithms on sparsity, data manifold adherence, and causality. 
    
    Among the collected papers, methods using solvers explicitly constrain sparsity~\citep{Ustun19:Actionable,karimi_model-agnostic_2020}, black-box methods constrain L0 norm of counterfactual and the input datapoint~\citep{medina_comparison-based_2018,dandl_multi-objective_2020}. 
    Gradient-based methods typically use the L1 norm of counterfactual and the input datapoint. 
    Some of the methods change only a fixed number of features~\citep{white_measurable_2019,keane2020good}, change features iteratively~\citep{Grace:2019,Epistemic_and_Aleatoric_uncertainty, verma2021amortized, Gradual_Construction}, or flip the minimum possible split nodes in the decision tree~\citep{guidotti_local_2018} to induce sparsity. 
    Some methods also induce sparsity post-hoc~\citep{medina_comparison-based_2018,mothilal_explaining_2020}. 
    This is done by sorting the features in ascending order of relative change and greedily restoring their values to match the values in the input datapoint until the prediction for the CFE is still different from the input datapoint.
    
    % Some recent papers have raised concerns that counterfactual explanations ought to similar to the training data in order to be realistic. 
    Adherence to the data manifold has been addressed using several different approaches, like training VAEs on the data distribution~\citep{dhurandhar_explanations_2018,joshi_towards_2019,van_looveren_interpretable_2020,mahajan_preserving_2020}, constraining the distance of a counterfactual from the $k$ nearest training datapoints~\citep{dandl_multi-objective_2020,Kanamori2020:DACE, prototype_based_cfe}, directly sampling points from the latent space of a VAE trained on the data, and then passing the points through the decoder~\citep{pawelczyk_learning_2020}, using an ensemble of model to capture the predictive entropy~\citep{Epistemic_and_Aleatoric_uncertainty}, using an Kernel Density Estimator (KDE) to estimate PDF of underlying data manifold \citep{coherent_CFEs}, using cycle consistency loss in GAN~\citep{conditional_gan_cfe_looveren}, mapping back to the data domain~\citep{Grace:2019}, using a combination of existing datapoints~\citep{keane2020good}, using Gaussian Mixture Models to approximate the probability of in-distributionness~\citep{efficient-contrastive}, or by using feature correlations~\citep{convex_optimization_cfe}, or by simply not generating any new datapoint~\citep{poyiadzi_face_2020}. 
    
    % Another important consideration is to encode any strong prior knowledge we might have about the features. Specifically, causal relations between features should be taken into account while generating counterfactuals. 
    The relation between different features is represented by a directed graph between them, which is termed as a causal graph~\citep{causality:Pearl}. 
    Out of the papers that have addressed this concern, most require access to the complete causal graph~\citep{karimi_algorithmic_2020,karimi-imperfect:2020} (which is rarely available in the real world), while~\citet{mahajan_preserving_2020,verma2021amortized, gan_cfe_amortized,prototype_based_cfe} can work with partial causal graphs.
    
    These three properties are reported in the table. 
    \begin{itemize}%[leftmargin=0.3em]
        \item  \emph{Sparsity}: This column is marked \texttt{No} if the algorithm does not consider sparsity, else it specifies the sparsity constraint. 
        \item  \emph{Data manifold}: This column is marked \texttt{Yes} if the algorithm forces the generated counterfactuals to be close to the data manifold by some mechanism; otherwise, it is marked \texttt{No}. 
        \item  \emph{Causal relation}: This column is marked \texttt{Yes} if the algorithm considers the causal relations between features when generating counterfactuals; otherwise, it is marked \texttt{No}. 
    \end{itemize}

    \item \emph{Counterfactual (CF) optimization (opt.) problem attributes}: These are a few attributes of the optimization problem. 
    
    % An important fact for a counterfactual to consider is the mutability and actionability of a feature.
    % As we previously mentioned, immutable features should not be subject to change in a counterfactual. 

    Out of the papers that consider feature actionability, most classify the features into immutable and mutable types. 
    \citet{karimi_algorithmic_2020} and \citet{inverse-classification2} categorize the features into immutable, mutable, and actionable types. 
    Actionable features are a subset of mutable features. They point out that certain features are mutable but not directly actionable by the individual, e.g., \emph{CreditScore} cannot be directly changed; it changes as an effect of changes in other features like income, credit amount. 
    \citet{mahajan_preserving_2020} uses an oracle to learn the user preferences for changing features (among mutable features) and can also learn hidden preferences. %which can be hidden as well. 
    
    Most tabular datasets have both continuous and categorical features. Performing arithmetic over continuous features is natural, but handling categorical variables in gradient-based algorithms can be complicated. 
    Some algorithms cannot handle categorical variables and filter them out~\citep{medina_comparison-based_2018,lucic-actionable:2020}. 
    \citet{wachter_counterfactual_2017} proposed clamping all categorical features to each of their values, thus spawning many processes (one for each value of each categorical feature), leading to scalability issues. 
    Some approaches convert categorical features to one-hot encoding and then treat them as numerical features. In this case, maintaining one-hotness can be challenging. 
    Some use a different distance function for categorical features, which is generally an indicator function (1 if a different value, else 0).  \citep{coherent_CFEs} use Markov chain transitions to encode categorical distances.
    \citet{gan_cfe_amortized} use Gaussian mixture models to normalize the continuous features and Gumbel-Softmax to relax categorical features into continuous ones.
    Genetic algorithms, evolutionary algorithms, and SMT solvers can naturally handle categorical features. 
    We report these properties in the table. 
    \begin{itemize}%[leftmargin=*]
        \item  \emph{Feature preference}: This column is marked \texttt{Yes} if the algorithm considers feature actionability, otherwise marked \texttt{No}. 
        \item  \emph{Categorical distance function}: This column is marked \texttt{-} if the algorithm does not use a separate distance function for categorical variables, else it specifies the distance function. 
    \end{itemize}

\end{enumerate}

\newcolumntype{R}[1]{>{\raggedleft\arraybackslash}p{#1}}

\belowcaptionskip=10pt
\begin{table*}
\caption{Assessment of the collected papers on the key properties, which are important for readily comparing and comprehending the differences and limitations of different counterfactual algorithms. Papers are sorted chronologically. Details about the full table is given in appendix~\ref{sec:full-table}.}
\centering
\label{tab:main-table}

\resizebox{\textwidth}{!}{%

 \begin{tabular}{m{0.04\textwidth}m{0.06\textwidth}m{0.095\textwidth}m{0.13\textwidth}m{0.04\textwidth}m{0.06\textwidth}m{0.12\textwidth}m{0.10\textwidth}m{0.11\textwidth}m{0.11\textwidth}m{0.125\textwidth}}

 \toprule
 &  & \vbarmc{2}{Assumptions} & \vbarmc{2}{Optimization amortization} & \vbarmc{3}{CF attributes} & \vbarmc{2}{CF opt. problem attributes} \\ \cmidrule{3-11}    
 Year & Paper & Model \mbox{access} & Model \mbox{domain} & Amortized \mbox{Inference} & Multiple CFEs & Sparsity & Data \mbox{manifold} & Causal \mbox{relation} & Feature \mbox{preference} & Categorical dist. func \\
 
 \midrule
 \multirow{3}{*}{2017 $\begin{dcases*} \\ \\ \\ \\  \end{dcases*}$} & \citep{inverse-classification2} & Black-box & Agnostic & No & No & Iteratively & No & No & Yes & - \\
 & \citep{wachter_counterfactual_2017} & Gradients & Differentiable & No & No & L1 & No & No & No & - \\
 & \citep{Tolomei2017:Interpretable} & Complete & Tree ensemble & No & No & No & No & No & No & - \\
 \multirow{5}{*}{2018 $\begin{dcases*} \\ \\ \\ \\ \\ \\ \end{dcases*}$} & \citep{medina_comparison-based_2018} & Black-box & Agnostic & No & No & L0 and post-hoc & No & No & No & - \\ % 18 May 2018   % growing spheres
 & \citep{guidotti_local_2018} & Black-box & Agnostic & No & Yes & Flips min. split nodes & No & No & No & Indicator \\ % 28 May 2018       % not an optimization problem
 & \citep{dhurandhar_explanations_2018} & Gradients & Differentiable & No & No & L1 & Yes & No & No & - \\ % october 2018
 & \citep{grath_interpretable_2018} & Black-box & Agnostic & No & No & No & No & No & No\footnote{It considers global and local feature importance, not preference.} & - \\ % Nov 2018 % Nelder-mead 
 \multirow{8}{*}{2019 $\begin{dcases*} \\ \\ \\ \\ \\ \\ \\ \\ \\ \\ \\ \\ \end{dcases*}$} & \citep{russell_efficient_2019} & Complete & Linear & No & Yes & L1 & No & No & No & N.A.\footnote{All features are converted to polytope type.} \\ % 2nd Jan 2019
 & \citep{Ustun19:Actionable} & Complete & Linear & No & No & Hard \mbox{constraint} & No & No & Yes & - \\ % Jan 29, 2019
 & \citep{sharma_certifai_2019} & Black-box & Agnostic & No & Yes & No & No & No & Yes & Indicator \\ % 19th May 2019   % genetic 
 & \citep{dhurandhar_model_2019} & Black-box or gradient & Differentiable & No & No & L1 & Yes & No & No & - \\ % 31st May 2019 % FISTA
 & \citep{rathi-generating:2019} & Black-box & Agnostic & No & No & No & No & No & No & - \\ % 21th June, 2019 % Not an optimization problem
 & \citep{joshi_towards_2019} & Gradients & Differentiable & No & No & No & Yes & No & No & - \\
 % 22nd July 2019
 & \citep{ramakrishnan_synthesizing_2019} & Gradients & Differentiable & No & No & No & No & No & No & - \\ % 9th Oct 2019
 & \citep{white_measurable_2019,white_measurable_2021_supporting} & Black-box & Agnostic & No & No & Changes one feature & No & No & No & - \\ % 23rd Nov 2019 % Not an optimization problem
 \multirow{17}{*}{2020 $\begin{dcases*} \\ \\ \\ \\ \\ \\ \\ \\ \\ \\ \\ \\ \\ \\ \\ \\ \\ \\ \\ \\ \end{dcases*}$} & \citep{mothilal_explaining_2020} & Gradients & Differentiable & No & Yes & L1 and post-hoc & No & No & No & Indicator \\ % 27th Jan 2020
 & \citep{poyiadzi_face_2020} & Black-box & Agnostic & No & No & No & Yes\footnote{Does not generate new datapoints} & No & No & - \\ % 7th Feb, 2020  % Dijkstra's algorithm
 & \citep{van_looveren_interpretable_2020} & Black-box or gradient & Differentiable & No & No & L1 & Yes & No & No & Embedding\\ % 18th Feb, 2020   % FISTA
 & \citep{mahajan_preserving_2020} & Gradients & Differentiable & Yes & Yes & No & Yes & Yes & Yes & - \\ % 22nd Feb, 2020
 & \citep{karimi_model-agnostic_2020} & Complete & Linear & No & Yes & Hard \mbox{constraint} & No & No & Yes & Indicator \\ % 28th Feb, 2020 This is claimed to be model-agnostic, but it is not. It requires linear models, can't do sigmoid activation. 
 & \citep{pawelczyk_learning_2020} & Gradients & Differentiable & No & No & No & Yes & No & Yes & N.A.\footnote{The distance is calculated in latent space.} \\ % 20th Apr, 2020
 & \cite{keane2020good} & Black-box & Agnostic & No & No & Yes & Yes & No & No & - \\ % 26th May, 2020
 & \citep{karimi_algorithmic_2020} & Complete & Linear and causal graph & No & No & L1 & No & Yes & Yes & - \\ % 11th Jun, 2020
 & \citep{karimi-imperfect:2020} & Gradients & Differentiable & No & No & No & No & Yes & Yes & - \\ % 16th June 2020
 & \citep{Grace:2019} & Gradients & Differentiable & No & No & Iteratively & Yes & No  & No\footnote{It considers feature importance not user preference.} & - \\ % 21st June, 2020
 & \citep{dandl_multi-objective_2020} & Black-box & Agnostic & No & Yes & L0 & Yes & No & Yes & Indicator \\ % 24th June 2020 % Genetic
 & \citep{Kanamori2020:DACE} & Complete & Linear and tree ensemble & No & No & No & Yes & No & Yes & - \\ % July, 2020
 & \citep{fernandez-random:2020} & Complete & Random \mbox{Forest} & No & Yes & L1 & No & No & No & - \\ % 3rd July 2020
 & \citep{lucic-actionable:2020,lucic-actionable:2020-update1} & Complete & Tree ensemble & No & No & L1 & No & No & No  & - \\ % 9th July 2020 
 %% Second version papers
%  \midrule - there were 5 second version papers here, shifted down for journal submission
  
 \bottomrule
 \end{tabular}

 }

\end{table*}

\begin{table*}
    \caption{Continued from \Cref{tab:main-table}}
    \centering
    \label{tab:main-table2}
    
    \resizebox{\textwidth}{!}{%
    
    \begin{tabular}{m{0.04\textwidth}m{0.06\textwidth}m{0.095\textwidth}m{0.13\textwidth}m{0.04\textwidth}m{0.06\textwidth}m{0.12\textwidth}m{0.10\textwidth}m{0.11\textwidth}m{0.11\textwidth}m{0.125\textwidth}}

    \toprule
    &  & \vbarmc{2}{Assumptions} & \vbarmc{2}{Optimization amortization} & \vbarmc{3}{CF attributes} & \vbarmc{2}{CF opt. problem attributes} \\ \cmidrule{3-11}    
    Year & Paper & Model \mbox{access} & Model \mbox{domain} & Amortized \mbox{Inference} & Multiple CFEs & Sparsity & Data \mbox{manifold} & Causal \mbox{relation} & Feature \mbox{preference} & Categorical dist. func \\
    \midrule
    % \citep{efficient-contrastive} & Complete & Linear & No & No & L1 & Yes & No & No & - \\ %4 Jan 2021
    \multirow{23}{*}{2021 $\begin{dcases*} \\ \\ \\ \\ \\ \\ \\ \\ \\ \\ \\ \\ \\ \\ \\ \\ \\ \\ \\ \\ \\ \\ \\ \\ \\ \\ \\ \\ \end{dcases*}$} & \citep{conditional_gan_cfe_looveren} & Gradient & Differentiable & Yes & No & L1 & No\footnote{Maybe partially as it uses cycle consistency loss} & No & No & - \\ % 25 Jan 2021
    & \citep{oblique-tree-cfe,oblique-tree-cfe-copy} & Complete & Decision Tree & No & Yes & L1 & No & No & Yes & - \\ % 1st Mar, 2021 
    & \citep{Ordered-CFE-Kanamori} & Complete & Linear & No & Yes & Iteratively & No & Yes & No & - \\  % 14th March, 2021
    & \citep{Epistemic_and_Aleatoric_uncertainty} & Gradients & Differentiable & No & No & Iteratively & Yes & No & Yes & -  \\ %16 Mar 2021 
    & \citep{naumann2021consequenceaware} & Black-box & Agnostic & No & Yes & Gower & No & Yes & Yes & Gower \\  % 12th April, 2021
    & \citep{nice_cfe} & Black-box & Agnostic & No & No & Yes & Yes & No & No & Indicator \\ %15 Apr 2021
    & \citep{prototype_based_cfe} & Black-box & Agnostic & No & No & No & No & Yes & No & Latent space \\ % 3rd May, 2021
    & \citep{CFE-scorecard} & Complete & Linear & No & Yes & Hard \mbox{constraint} & Yes & No & Yes & - \\ %9 May 2021
    & \citep{convex_optimization_cfe} & Complete & Linear & No & No & No & Yes & No & No & - \\ % 17th May, 2021
    & \citep{schleich2021geco} & Black-box or complete & Agnostic if black-box & No & Yes & L0/L1 & No & Yes & Yes & Indicator \\ %19th May 2021
    & \citep{nemirovsky-hired-people-cfe-countergan} & Black-box or gradient & Agnostic if black-box & Yes & No & L1 & Yes & No & Yes & - \\ % 27th May, 2021
    & \citep{Pierre-tree-ensemble-pure-region} & Complete & Tree ensemble & Yes & No & Yes & No & No & Yes & - \\ % 31st May 2021
    & \citep{rl_cfe_approach2_amortized} & Black-box & Agnostic & Yes & Yes & L0/L1 & Yes & No & Yes & Indicator \\ % 4 Jun 2021
    & \citep{verma2021amortized} & Black-box & Agnostic & Yes & Yes & Iteratively & Yes & Yes & Yes & - \\ % 7th June, 2021
    & \citep{CFE-Tree-ensembles} & Complete & Tree ensemble & No & No & L0/L1 & Yes & No & Yes & Gower \\ %25 Jun 2021
    & \citep{scaling_Nearest_CFE} & Complete & Linear & No & Yes & Hard \mbox{constraint} & No & No & Yes & Indicator \\ % 30 July 2021
    & \citep{Fraunhofer-pure-region-sampling} & Black-box & Agnostic & Yes & Yes & No & No & No & No & - \\ % 11th August, 2021
    & \citep{gan_cfe_amortized} & Black-box & Agnostic & Yes & Yes & No & Yes & Yes & No & Not sure \\ % 14 August 2021
    & \citep{Gradual_Construction} & Gradient & Differentiable & No & No & No & No & No & No & - \\ % 6 Sep 2021 \\
    & \citep{coherent_CFEs} & Black-box & Agnostic & No & No & L1 & Yes & No & No & Markov Chains \\ % 30 Sep 2021
    & \citep{monte-carlo-cfe-technique} & Black-box & Agnostic & Partially & Yes & Hard constraint & No & No & Yes & Gower \\ % 18th Nov 2021
    \multirow{5}{*}{2022 $\begin{dcases*} \\ \\ \\ \\ \\ \\ \\ \end{dcases*}$} & \citep{guo-cfe-counternet} & Training from scratch & Differentiable & Yes & No & No & No & No & No & - \\ % Jan 30, 2022
    & \citep{Xiang2022-Realistic-VAE-CFE} & Gradient & Differentiable & No & No & No & Yes & Yes & No & - \\ % 15th Feb, 2022
    & \citep{user-specific-cost} & Black-box & Agnostic & No & Might & Yes & No & No & Yes & - \\ % 21st Feb 2022
    & \citep{hima-beyond-recourse-globalcfe} & Black-box & Agnostic & Yes & Might & Yes & No & No & Yes & Indicator \\ % 15th April, 2022
    & \citep{sum-product-networks-cfe} & Training from scratch & Differentiable & No & No & No & Yes & No & Yes & - \\  % 16th May, 2022
     \bottomrule
     \end{tabular}
     }
    
    \end{table*}

\section{Evaluation of counterfactual generation algorithms}
\label{sec:eval}
% \subsection{Commonly used algorithms to solve the optimization problem for generation of CFEs} % This has been partially covered in the "Optimization required section", and the methods are so varied that it is hard to summarize them. 
This section lists the common datasets used to evaluate counterfactual generation algorithms and the metrics on which they are typically evaluated and compared. 

\subsection{Commonly used datasets for evaluation}
The datasets used in the evaluation in the papers we review can be categorized into tabular and image datasets. Not all methods support image datasets. 
Some of the papers also used synthetic datasets for evaluating their algorithms, but we skip those in this review since they were generated for a specific paper and also might not be available. 
Common datasets in the literature include:
\begin{itemize}[leftmargin=*]
    \item \emph{Image}: MNIST~\citep{lecun-mnisthandwrittendigit-2010}, EMNIST~\citep{EMNIST-data}, CelebA~\citep{celebA_data}, CheXpert~\citep{chexpert-data}, ImageNet~\citep{imagenet-data}, ISIC Skin Lesion~\citep{isic-skin-data}, ADNI~\citep{alzheimer-data}, ChestX-ray8~\citep{chest-xray-data}.
    
    \item \emph{Tabular}: Adult income, German credit, Student Performance, Breast cancer, Default of credit, Shopping, Iris,  Wine, Spambee, Covertype, ICU~\citep{UCI-repo}, LendingClub~\citep{lendingclub-data}, Give Me Some Credit \citep{givemesomecredit-data}, COMPAS~\citep{compas-data}, LSAT~\citep{lsat-data}, Pima diabetes~\citep{pima-diabetes-data}, HELOC/FICO~\citep{fico-data}, Fannie Mae~\citep{fannie-data}, Portuguese Bank~ \citep{Portuguese_Bank_data}, Sangiovese~\citep{Sangiovese-data}, Bail dataset~\citep{bail-data}, 
    Simple-BN~\citep{mahajan_preserving_2020}, AllState~\citep{allstate-data}, WiDS Datathon~\citep{woman-in-cs-data}, Home Credit Default Risk~\citep{home-credit-data}, German Housing~\citep{forster-capturing-2021}, HospitalTriage~\citep{hospitaltriage-data}, MIMIC-IV~\citep{MMIC-IV}, Freddie Mac~\citep{freddiemac-data}, UK unsecured personal loans~\citep{stress-test-creditscore}, insurance dataset \citep{inverse-classification-multiple-algos}, BPIC2017 \citep{Loreley-huang-2022}. 
\end{itemize}

\subsection{Metrics for evaluation of counterfactual generation algorithms} 

Most of the counterfactual generation algorithms are evaluated on the desirable properties of counterfactuals.  
Counterfactuals are considered actionable feedback to individuals who have received undesirable outcomes from automated decision-makers, and therefore, a user study can be considered a gold standard. 
The ease of acting on a recommended counterfactual is thus measured by using quantifiable proxies:
% \vspace{-0.8em}
\begin{enumerate}[leftmargin=*]
    \item \emph{Validity}: Validity measures the ratio of the counterfactuals that actually have the desired class label to the total number of counterfactuals generated. 
    Higher validity is preferable. Most papers report it. 
    
    \item \emph{Proximity}: Proximity measures the distance of a counterfactual from the input datapoint. 
    For counterfactuals to be easy to act upon, they should be close to the input datapoint. Distance metrics like the L1 norm, L2 norm, Mahalanobis distance are common. 
    To handle the variability of range among different features, some papers standardize them in pre-processing or divide L1 norm by median absolute deviation of respective features~\citep{wachter_counterfactual_2017,mothilal_explaining_2020,russell_efficient_2019}, or divide L1 norm by the range of the respective features~\citep{karimi_algorithmic_2020,karimi_model-agnostic_2020,dandl_multi-objective_2020}. 
    Some papers term proximity as the average distance of the generated counterfactuals from the input. Lower values of average distance are preferable. 
    
    \item \emph{Sparsity}: Shorter explanations are more comprehensible to humans~\citep{Miller-xai:2019}, therefore, counterfactuals ideally should prescribe a change in a small number of features. Although a consensus on a hard cap on the number of modified features has not been reached, \citet{keane2020good} cap a sparse counterfactual to at most two feature changes. 
    
    \item \emph{Counterfactual generation time}: Intuitively, this measures the time required to generate counterfactuals. This metric can be averaged over the generation of a counterfactual for a batch of input datapoints or for the generation of multiple counterfactuals for a single input datapoint. 
    
    \item \emph{Diversity}: Some algorithms support the generation of multiple counterfactuals for a single input datapoint. The purpose of providing multiple counterfactuals is to increase the ease for applicants to reach at least one counterfactual state. Therefore, the recommended counterfactuals should be diverse, allowing applicants to choose the easiest one. 
    If an algorithm is strongly enforcing sparsity, there could be many different sparse subsets of the features that could be changed. Therefore, having a diverse set of counterfactuals is useful. 
    Diversity is encouraged by maximizing the distance between the multiple counterfactuals by adding it as a term in the optimization objective~\citep{mothilal_explaining_2020,dandl_multi-objective_2020} or as a hard constraint~\citep{Ustun19:Actionable,karimi_model-agnostic_2020,scaling_Nearest_CFE}, or by minimizing the mutual information between all pairs of modified features~\citep{Grace:2019}. \citet{mothilal_explaining_2020} reported diversity as the feature-wise distance between each pair of counterfactuals. A higher value of diversity is preferable. 
    
    \item \emph{Closeness to the training data}: Recent papers have considered the actionability and realisticness of the modified features by grounding them in the training data distribution. This has been captured by measuring the average distance to the k-nearest datapoints~\citep{dandl_multi-objective_2020}, or measuring the local outlier factor~\citep{Kanamori2020:DACE}, or measuring the reconstruction error from a VAE trained on the training data~\citep{mahajan_preserving_2020,van_looveren_interpretable_2020}, or measuring the PDF of such datapoints using KDE \citep{coherent_CFEs}, or measuring the maximum mean discrepancy (MMD) between the original and counterfactual points~\citep{conditional_gan_cfe_looveren}. A lower value of the distance and reconstruction error is preferable. 
    
    \item \emph{Causal constraint satisfaction (feasibility)}: This metric captures how realistic the modifications in the counterfactual are by measuring if they satisfy the causal relation between features. \citet{mahajan_preserving_2020} evaluated their algorithm on this metric.
    
    \item \emph{IM1 and IM2}: \citet{van_looveren_interpretable_2020} proposed two interpretability metrics specifically for algorithms that use auto-encoders. Let the counterfactual class be $t$, and the original class be $o$. $AE_t$ is the auto-encoder trained on training instances of class $t$, and $AE_o$ is the auto-encoder trained on training instances of class $o$. Let $AE$ be the auto-encoder trained on the full training dataset (all classes).
    \begin{equation}
        IM1 = \frac{\twonorm{x_{cf} - AE_t(x_{cf})}}{\twonorm{x_{cf} - AE_o(x_{cf})} + \epsilon}
    \end{equation}
    
    \begin{equation}
        IM2 = \frac{\twonorm{AE_t(x_{cf}) - AE(x_{cf})}}{\norm{x_{cf}} + \epsilon}
    \end{equation}
    %\emph{IM1} is the ratio of the reconstruction error of the counterfactual using an auto-encoder trained on the counterfactual class divided by the reconstruction error of the counterfactual using an auto-encoder trained on the original class. \emph{IM2} is a scaled L2 norm of the reconstruction from an auto-encoder trained on the counterfactual class and the reconstruction from an auto-encoder trained on all the classes. 
    A lower value of \emph{IM1} implies that the counterfactual ($x_{cf}$) can be better reconstructed by the auto-encoder trained on the counterfactual class ($AE_t$) compared to the auto-encoder trained on the original class ($AE_o$). Thus implying that the counterfactual is closer to the data manifold of the counterfactual class.  
    A lower value of \emph{IM2} implies that the reconstruction from the auto-encoder trained on the counterfactual class and the auto-encoder trained on all classes is similar. 
    Therefore, a lower value of \emph{IM1} and \emph{IM2} means a more interpretable counterfactual. 
    
    \item \emph{Label Variation Score and Oracle Score}: \citet{hvilshoj-cfe-new-metrics} point out that the previous metrics are unable to detect out-of-distribution CFEs (especially for high dimensional datasets) and propose two new metrics. \emph{Label Variation Score} applies when each datapoint has multiple labels, and the intuition is that CFE for a particular label should not affect the predictions for other labels (unless they are highly correlated). 
    
    \begin{equation}
        LVS = \sum_{l \in L} d_{div} [ p_{l}(x), p_{l}(CFE(x)) ]
    \end{equation}
    where L is the total number of labels for a datapoint and $p_{l}$ is the predicted probability for the specific label $l$, and $d_{div}$ measures the divergence between the predicted probability of label $l$ for the original datapoint $x$ and its CFE. 
    
    \emph{Oracle Score} is similar to validity, however, with an additional classifier trained on the same dataset as the original classifier. The intuition is that if a CFE is more like an adversarial example for a classifier, the CFE would not be classified in the desired class by the other classifier, and hence we use the prediction from the additional classifier as the ground truth validity. 
    
\end{enumerate}

Some of the reviewed papers did not evaluate their algorithm on any of the above metrics. They only showed a couple of example inputs and respective CFEs, details about which are available in the full table (see~\cref{sec:full-table}).

\subsection{Other works}
\label{sec:other_works}
This section enlists works that talk about the desirable properties of counterfactuals or point to their issues. We also talk about works that propose minor modifications to previous similar approaches. \\

\textbf{Works exploring desirable CFE properties: }
\citet{sokol_desiderata_2019} list several desirable properties of counterfactuals inspired from~\citet{Miller-xai:2019} and state how the method of flipping logical conditions in a decision tree satisfies most of them. 
\citet{issues_posthoc} enlist \emph{proximity}, \emph{connectedness}, and \emph{stability} as three desirable properties of a CFE and propose the metrics to measure them.

\textbf{Works pointing to issues with CFEs:} 
\citet{laugel_dangers_2019} says that if the explanation is not based on training data, but the artifacts of non-robustness of the classifier, it is unjustified.
They define justified explanations to be connected to training data by a continuous set of datapoints, termed $\mathcal{E}$-chainability. 
\citet{hidden_assumptions} state five reasons that have led to the success of counterfactual explanations and also point out the overlooked assumptions. They mention the unavoidable conflicts which arise due to the need for privacy invasion in order to generate helpful explanations. 
\citet{Atoosa-philosophy-cfe} provide philosophical insight into the implicit assumptions and choices made when generating CFEs.

\textbf{Causal CFEs: } 
\citet{cruds_cf} propose using conditional subspace VAEs (CSVAE), a variant of VAEs, to generate CFEs that obey correlations between features, causal relations between features, and personal preferences. This method builds a probabilistic data model of the training data using a CSVAE and uses it to generate CFEs. However, these CFEs are not with respect to a specific ML model. 
\citet{cf-causal-latent-ontop} propose a technique that can be used with any counterfactual generation approach to generate causality abiding CFEs. 
\citet{bernhard-causal-cfe-confounded} extend \citet{karimi-imperfect:2020}'s work to the setting where unobserved confounders may be present in the causal setting. 
\citet{causality-from-transport-cfe} show that optimal transport-based methods are an approximation of Pearl's CFEs and hence can be used to generate causal CFEs. 
\citet{beckers-causal-xai} delve further into the integration of causality, actual causation, and CFEs. 

\textbf{CFE for specific models: }
\citet{bayesian-network-CFE} propose a CFE generation approach targeted for Bayesian network classifiers. 
\citet{artelt_computation_2019, efficient-contrastive} enlists the counterfactual optimization problem formulation for several model-specific cases, like generalized linear model, gaussian naive bayes, and mention the general algorithm to solve them. 
\citet{Koopman2021PersuasiveCE} propose a BFS-based technique for generating CFEs for Bayesian networks. 

\textbf{Works considering multi-agent scenarios of CFEs: }
\citet{Manuel:game-theory} cast the counterfactual generation problem as a Stackelberg game between the decision maker and the person receiving the prediction. Given a ground set of CFEs, the proposed algorithm returns the top-k CFEs, which maximizes the utility of both the involved parties. 
\citet{post-hoc-CFE-not-good} point out that the interests of the provider and receiver of model explanations might be in conflict, and the ambiguous post-hoc explanations might be unsuitable for achieving the purpose of transparency as desired in GDPR. This also relates to fairwashing (see \RCref{ch:fairwashing}).

\textbf{Global CFEs: } 
\citet{hima-beyond-recourse-globalcfe} propose AReS to generate rule lists that act as global CFEs. \citet{ley-global-cfe-ares-improve} and \citet{kanamori-decision-tree-globalcfe} propose computationally more efficient implementation of \citet{hima-beyond-recourse-globalcfe}'s work.  \citet{carrizosa-global-cfe-tree-models} propose a mixed integer quadratic model to generate CFEs for a group of datapoints. 
\citet{inverse-classification-multiple-algos} propose generating CFEs for a set of datapoints using lagrangian and subgradient methods. 
\citet{Dhurandhar-global-model-consistent-with-cfe} propose a technique to train a globally interpretable model (for a black-box model) such that this model is consistent with the pertinent positives and pertinent negatives~\citep{dhurandhar_explanations_2018} of the training datapoints used to train the original model.

\textbf{Works proposing modifications to previous approaches: }
\citet{RL-for-CFE-relace-paper} and \citet{RL-for-CFE-mcts-paper} use RL to generate CFE as was also proposed by \citet{verma2021amortized}. 
\citet{rasouli2022CARE-CFE} propose a genetic algorithm to generate CFEs as was also proposed by \citet{dandl_multi-objective_2020}. 
\citet{CFE_genetic_creditscorecards} propose to use genetic algorithm for CFE generation similar to \citet{dandl_multi-objective_2020}'s work. 
\citet{another-genetic-algo-monteiro} propose extending \citet{dandl_multi-objective_2020}'s approach using U-NSGA-III evolutionary algorithm. 
\citet{mahajan-work-extension-linear-interpolation} extend \citet{mahajan_preserving_2020}'s work by interpolating between the input and CFE datapoint to generate CFEs closer to the input datapoint. 
\citet{semi-supervised-autoencoder-cfe} propose using a semi-supervised autoencoder instead of the traditional unsupervised autoencoder to generate CFEs close to the training data manifold. 
\citet{Loreley-huang-2022} propose LORELEY that extends LORE \citep{guidotti_local_2018} to generate CFEs for multi-class classification problems and account for flow constraints. 
\citet{student-moodle-cfe-cbr-technique} use feature importances provided by LIME to assist the case-based reasoning approach to generate CFEs. 
\citet{Delaney2021Uncertainty-CFE} propose using trust scores to measure the out-of-distributionness of the CFEs. 
\citet{guidotti-cfe-ensemble-of-explainers} propose using an ensemble of base CFE explainers to generate diverse CFEs. 

\textbf{Benchmark and dataset curation: }
\citet{cfe-algos-quantitative-comparison-survey} quantitatively compare 10 CFE generating approaches using 22 datasets and nine metrics. 
\citet{pawelczyk2021CARLA-toolbox} and \citet{artelt-ceml-toolbox} have developed extensible toolboxes where several CFE approaches can be plugged in and compared on specific datasets.

\textbf{Various uncategorized works: }
\citet{cfe-logic-programming-laurastate} talk about generating CFEs with real-world constraints on features and adaptability with updating ML models using constraint logic programming. 
\citet{action-cfe-tahoun} propose to disentangle actions from feature modifications to address the lack of intervention data and appropriate action costs. The users should already describe the actions they are willing to take, and a model should just choose the minimum cost action that generates the CFE. 
\citet{contrastive-xai-retail-forecast} propose a CFE approach to provide a lower and upper bound for the feature values that get a low prediction error from the ML model for a datapoint that originally had a high prediction error. 
\citet{korikov-cfe-inverse-optimization1, korikov-cfe-inverse-optimization2} show how CFEs can be generated by using the generalization of inverse combinatorial optimization and solve it under two objectives. 
\citet{predictive_multiplicity} provide a general upper bound on the cost of counterfactual explanations under the phenomenon of predictive multiplicity, wherein more than one trained models have the same test accuracy and there is no clear winner among them. 
\citet{cfe-biology-multiclass} propose a hierarchical decompositions-based method to obtain CFEs for multi-class classification problems. 
\citet{declarative-CFE} and \citet{cfe-for-monotone-obj} propose brute force approaches to generate CFEs. 
% and their approach is restricted to categorical features. 
% \citet{cfe-for-monotone-obj} propose another brute force approach to generate a set of CFEs provided that the objective function is monotonic. 

% -----------------------------------
% \citet{what-if-tool} developed a model-agnostic interactive visual tool for letting developers and practitioners visually examine the effect of changes in various features. 
% This tool has added functionality for exploring data like partial dependence plots, fairness, and performance evaluation.     % I can skip this tool. 
% \citet{fernandez-loria_explaining_2020} point out at the insufficiency of feature importance methods for explaining a model's predictions, and substantiate it with a synthetic example. 
% They generate counterfactuals by removing features instead of modifying feature values. 
% There exist papers which do not propose novel algorithms to generate counterfactuals, but explore other aspects about it -- we mention those papers in this section. (We also relegate discussion about papers that propose algorithms similar to ones proposed by previous approaches to this section.) 

\section{Counterfactual Explanations for Other Data Modalities}
\label{sec:otherdatamodality}

Since we restrict this survey to the papers that generate CFEs for tabular data, 
in this section we point the readers to the papers that propose algorithms targeted towards other data modalities: 
\begin{enumerate}
    \item \emph{Image data}: \citep{Hendricks-image-CFE, beyond_trivial_cfe_images, liu_generative_2019, cdeepex_images, attribute_based_images, cfe_knockoff_images, da-dgcex_images, fast_cfe_images, latent-cf_images, conditional_gan_cfe_looveren, contrastive-cfe-images-CLEAR, cfe-text+images-DiscGrad, cfe_images_onfly_scope_disc, Elliott2021_adversarial_cfe_images, cfe-images-STEEX, oh-cfe-images-BIN, Akula-CFE-CoCoX-images, conceptual-cfe-abid, Filandrianos-CFE-Images, Smith-CFE-images-robot-action, akula-cfe-CX-ToM-images-interactivity, khorram-cycle-cfe-images, li-cfe-chest-xray-images, holtgen-DeDUCE-images, sanchez-diffusion-causal-cfe-images, jeanneret-diffusion-cfe-images, ghandeharioun-DISSECT-images, Dash-evaluate-bias-images, vermeire-explainable-images, Hvilshoj-ECINN-images, Mertes-GANterfactual-Medical-Images, yang-attribute-perturbation-text+images, Alipour-visual-question-answering-cfe, oh-cfe-alzheimer-images, vandenhende-cfe-visual-images, cfe-visual-explanation-goyal2019, Kenny-cfe-images-semifactual, SCOUT_cfe_images, encoder_cfe_image, Gohar-images-cfe-bayesian-model}. 
    
    \item \emph{Text data}: \citep{Gradual_Construction, tolkachev-CFE-text, cfe-text+images-DiscGrad, ramon-cfe-comparison-text+behavior, CFE-COIN-VQA, ravfogel-2021-cfe-text, Madaan-cfe-text, yang-cfe-financial-text, robeer-cfe-realistic-text, yang-attribute-perturbation-text+images, chen-2021-kace-cfe-text}.
    
    \item \emph{Speech data}: \citep{Zhang-emotion-vocal-recognition-cfe}. 
    
    \item \emph{Time-series data}: \citep{cfe_multivariate_time_series, conditional_gan_cfe_looveren, cf-timeseries1, wang-cfe-time-series-ICU-cardio, sulem-cfe-time-series-anomaly-detection, Dice-time-series-ext, wang_cfe-time-series-latentcf++, karlsson-cfe-time-series, cfe-sequential-data-Tsirtsis}. 
    
    \item \emph{Graph data for graph neural networks}: \citep{CF-GNN1, CF-GNN2, contrastive-graph-GNN3, input_perturbation_GNN4, CF-GNN5, robust-CFE-GNN, abrate-cfe-graph-brain-networks}. A survey for CFE on graph neural networks: \citep{cfe-graph-nn-survey}. 
    
    \item \emph{Agent action (e.g. Reinforcement Learning or Planning)}: \citep{CFE_for_rl, stein-cfe-agent-navigation, brandao-cfe-agent-motion-planning}. 
    
    \item \emph{Recommender systems}: \citep{cfe-reco-approach1, cfe-reco-approach2, cfe-reco-approach3, cfe-reco-approach4, cfe-reco-approach5, cfe-reco-approach6, cfe-reco-approach7, cfe-reco-approach8, sarah_rec, anna-cfe-recos}. 
    
    \item \emph{Functional data}: \citep{kovalev-CFE-functional, carrizosa-cfe-functional} and 
    \emph{Behavioral data}: \citep{ramon-cfe-comparison-text+behavior}. 

\end{enumerate}

\section{Other Applications of Counterfactual Explanations}
\label{sec:otherapplications}

Here we refer the readers to other applications where counterfactual explanations are being used apart from explaining ML models:
\begin{enumerate}
    \item \emph{Anomaly and data-drift detection}: \citet{hinder-cfe-drift} propose to use CFEs to explain data drift. \citet{sulem-cfe-time-series-anomaly-detection} propose to use CFEs to explain anomalies in time-series datasets. \citet{ravi-cfe-anomaly-detection} wrote a survey on the explainability techniques for convolutional auto-encoders for anomaly detection of images. \citet{cfe_anomaly_detection} propose to use CFEs to explain anomaly detection when using autoencoders. \citet{antoran2021-cfe-for-uncertainty-estimation} use CFEs to find changes in a datapoint that would help a classifier have a higher confidence in its prediction. 

    \item \emph{Training dataset debugging}: \citet{debugging-ml-models-cfe} propose to use CFEs to debug ML models by diagnosing the behavior and using synthetic data to alter the decision boundaries. \citet{debugging-data-cfe2} propose to use CFEs to debug potentially mislabeled datasets. \citet{cfe-financial-risk-debugging} propose to use CFEs to detect bugs in financial models. \citet{debugging-data-cfe3-han} propose finding a minimal subset of training datapoints that are responsible for a particular prediction and hence can be used to debug training datasets. 
    
    \item \emph{Data augmentation}: \citet{CFE-earning-call-data-augment} propose to use CFEs to augment training data that is used to predict market volatility based on earning calls. 
    \citet{CFE-classimbalance} propose using CFEs to augment training data to tackle the class imbalance problem. 
    \citet{hasan-cfe-augmentation-proximity,rasouli-cfe-robustness-tabular} propose using CFEs for data augmentation of tabular datasets for increased robustness. 
    \citet{temraz-cfe-data-augmentation-climate-change} propose using CFEs to generate data points that can be used to train ML models that predict crop growth (afflicted by climate change). 
    
    \item \emph{Drug designing}: \citet{nguyen2021-cfe-dta} use CFEs to find changes in a drug and protein molecule that will increase their affinity for each other. They use multi-agent RL to this end. 
    
    \item \emph{ML model bias detection}: \citep{Ustun19:Actionable,cfe-model-bias1, cfe-model-bias2}. 
    
    \item \emph{Various applications}: \citet{mazzine-CFE-employment} propose to use CFEs in employment services to help job seekers get personalized advice for increasing their propensity for getting recommended for a job and to help the ML developers to detect potential bias and other issues in their ML model. \citet{Sadler-cfe-community-detection} propose to use CFEs for community detection in social networks. \citet{dimension_reduction_cfe} propose to use CFEs to understand interactive dimensionality reduction. \citet{student-performance-cfe} propose to use CFEs for providing actionable suggestions to improve student performance in a university course. \citet{Cong-CFE-KS-test} propose a CFE approach to explain why a test set fails the Kolmogorov-Smirnov test. \citet{Marchezini2022-dh-cfe-latent-mental-health} propose to use CFE for altering both observational and latent variables to reason about mental health. \citet{yao-cfe-evaluate-recsys} propose to use counterfactuals for evaluating the explanations for recommender systems. \citet{gupta-cfe-constraint-satisfaction-problem} use CFEs to propose changes to constraint satisfaction problems that have no solutions. \citet{cfe-for-entity-resolution} propose using CFEs to explain entity resolution models. \citet{artelt-cfe-model-adaptation} use CFEs to explain the differences between the learning of a pair of models. 
    \citet{CRASS-dataset-NLP} propose a new dataset, CRASS, to test reasoning and natural language understanding of LLMs. 
\end{enumerate}

There has been one case of real-world deployment of CFEs in a hiring platform, \href{https://hired.com/}{Hired}. \citet{nemirovsky-cfe-gan-hiring-application} use a GAN-based approach~\citep{nemirovsky-hired-people-cfe-countergan} to suggest changes in features like expected salary, years of experience, and skills to candidates in order to get them approved by the Hired Marketplace ML model.

\section{Open Questions and Research Progress for solving them}

In the first version of this survey paper, we delineated the open questions and challenges yet to be tackled by the existing works pertaining to CFEs \citep{cfe-challenges-paper}. In this version, we supplement this section with the research progress made towards solving them and new research challenges. 

\begin{challenge}\label{ch:unification}
Unify counterfactual explanations with traditional ``explainable AI.''
\end{challenge}
Although counterfactual explanations have been credited to eliciting causal thinking and providing actionable feedback to users, they do not tell which feature(s) was the principal reason for the original decision and why. It would be nice if, along with giving actionable feedback, counterfactual explanations also gave the reason for the original decision, which can help applicants understand the model's logic. This is addressed by traditional ``explainable AI'' methods like LIME~\citep{ribeiro_why_2016}, Anchors~\citep{Ribeiro2018Anchors}, Grad-CAM~\citep{grad-cam}. 

\progress
\citet{guidotti_local_2018} have attempted this unification, as they first learn a local decision tree and then interpret the inversion of decision nodes of the tree as counterfactual explanations. However, they do not show the CFEs they generate, and their technique also misses other desiderata of counterfactuals (see~\cref{sec:themes}). 
\citet{unifying-cfe1-divyat} propose \textit{necessity} and \textit{sufficiency} as the two important properties of an explanation. Feature attribution explanations find the feature values that are sufficient for a prediction, while CFEs find the feature values that are necessary for a prediction. They propose methods to find the necessity and sufficiency of any feature subset and discuss how that aligns with finding CFEs. 
\citet{galhotra-sigmod-cfe} propose \textsc{Lewis} that also emphasizes the \textit{necessity} and \textit{sufficiency} scores of a feature subset in finding its global importance and in generating a CFE for local explainability. 
\citet{unifying-cfe2-deeplift} propose to use DeepLIFT to assign contribution scores to the features that changed in a counterfactual datapoint. 
\citet{ramon-cfe-comparison-text+behavior} rank the feature importances using LIME and SHAP, and then remove the features in decreasing order of importance until a CFE is found. 
\citet{DisCERN-use-shap-for-cfe} propose to use methods like LIME and SHAP to find feature importances and then replace the features in decreasing order of importance with the values borrowed from the nearest unlike neighbor (case-based reasoning approach). 
\citet{CF-Shap-paper} propose to change the background distribution used to compute the Shapley values to make the feature attribution amount to the counterfactual-ability of the features, i.e., changing a feature with higher attribution would have a higher probability of changing the prediction. 
\citet{SCOUT_cfe_images} propose to use the discriminant attribution explanations as a way to produce CFEs for images. 
\citet{student-moodle-cfe-cbr-technique} use LIME to assist case-based reasoning techniques to generate CFEs. 
\citet{use-cfe-instead-of-erasure} propose using counterfactual-ability of features as a metric for their feature importance.

\begin{challenge}\label{ch:discrete}
Provide counterfactual explanations as discrete and sequential steps of actions. 
\end{challenge}
Most counterfactual generation approaches return the modified datapoint, which would receive the desired classification. 
The modified datapoint (state) reflects the idea of instantaneous and continuous actions, but in the real world, actions are discrete and often sequential. 
Therefore the counterfactual generation process must take the discreteness of actions into account and provide a series of actions that would take the individual from the current state to the modified state, which has the desired class label. 

\progress
\citet{naumann2021consequenceaware} argue that to help an individual achieve the desired goal, CFEs should be provided as a sequential step of actions instead of just providing the final goal. \citet{sequential-CFE} conduct a user study to show the high preference for a sequential step of actions steps over a single-step goal. 
\citet{ramakrishnan_synthesizing_2019} propose a program synthesis based technique to generate such sequences. 
\citet{Ordered-CFE-Kanamori} propose a mixed-integer based programming method and \citet{verma2021amortized} propose an RL-based method that generates ordered sequences of actions as a CFE.

\begin{challenge}
Extend counterfactual explanations beyond classification.
\end{challenge}

\progress
Recent work has been extending counterfactual explanations to different tasks and model architectures.
\citet{cf-regression} propose a Bayesian optimization-based technique for generating CFEs for regression problems. 
\citet{CF-GNN1} propose an RL-based approach for generating CFEs for graph neural networks, which are used to predict chemical molecule properties. 
\citet{cf-timeseries1} propose a case-based reasoning approach to generate CFEs for a time-series classifier. 
See \Cref{sec:otherdatamodality} and \Cref{sec:otherapplications} for a list of all the approaches.

\begin{challenge}\label{ch:interactive}
Counterfactual explanations as an interactive service to the applicants. 
\end{challenge}
Counterfactual explanations should be provided as an interactive interface, where an individual can come at regular intervals, inform the system of the modified state, and get updated instructions to achieve the counterfactual state. 
This can help when the individual could not precisely follow the earlier advice for various reasons. 

\progress
\citet{GAMUT} developed an interactive user-interface for providing explanations to data scientists. They found out that data scientists used interactivity as the primary mechanism for exploring, comparing, and explaining predictions.
\citet{glass-box-voice-assistant} propose to enhance ML explanations with a voice-assisted interactive service. 
\citet{akula-cfe-CX-ToM-images-interactivity} propose an approach that explains an ML model using an interactive sequence of CFEs. 
\citet{skyline-cfe-interactive} propose refining the CFEs for different feature change costs based on user interactions. 

\begin{challenge}\label{ch:incomplete-causal}
The ability of counterfactual explanations to work with incomplete---or missing---causal graphs. 
\end{challenge}
Incorporating causality in the counterfactual generation is essential for the CFEs to be grounded in reality. 
Complete causal graphs and structural equations are rarely available in the real world, and therefore the algorithm should be able to work with incomplete causal graphs. 

\progress
\citet{mahajan_preserving_2020}'s approach was the first to be compatible with incomplete causal graphs. Now other works like \citet{galhotra-sigmod-cfe}, \citet{verma2021amortized}, \citet{schleich2021geco}, \citet{gan_cfe_amortized} can also work with partial causal graphs.

\begin{challenge}\label{ch:missing-feature}
The ability of counterfactual explanations to work with missing feature values. 
\end{challenge}
Along the lines of an incomplete causal graph, counterfactual explanation algorithms should also be able to handle missing feature values, which often happens in the real world~\citep{missing-data}.

\begin{challenge}\label{ch:scale}
Scalability and throughput of counterfactual explanations generation. 
\end{challenge}
As we see in \cref{tab:main-table}, most approaches need to solve an optimization problem to generate one counterfactual explanation. 
Some papers generate multiple counterfactuals while optimizing once, but they still need to optimize separately for different input datapoints. However, for industrial deployment, the generation should be more scalable.  

\progress
\citet{mahajan_preserving_2020} learn a VAE which can generate multiple CFEs for any given input datapoint after training. Therefore, their approach is highly scalable and is termed as ``amortized inference''. 
\citet{verma2021amortized} proposed an RL-based technique, FastAR, that also generates amortized CFEs. \citet{conditional_gan_cfe_looveren}, \citet{rl_cfe_approach2_amortized}, \citep{gan_cfe_amortized}, \citet{hima-beyond-recourse-globalcfe}, and \citet{nemirovsky-hired-people-cfe-countergan} also propose approaches to this end.

\begin{challenge}\label{ch:bias-in-cfe}
Counterfactual explanations should account for bias in the classifier. 
\end{challenge}
Counterfactuals potentially capture and reflect the bias in the models. 
To underscore this as a possibility, \citet{Ustun19:Actionable} experimented on the difference in the difficulty of attaining a counterfactual state across genders, which clearly showed a significant difference. 
More work must be done to find how equally easy counterfactual explanations can be provided across different demographic groups, or how adjustments should be made to the prescribed changes to account for the bias. 

\progress
\citet{hima-beyond-recourse-globalcfe} generate recourse rules for a subgroup that they use to detect model biases. 
\citet{Gupta2019EqualRecourse} propose adding a regularizer while training a classifier that encourages the classifier to maintain a similar distance of the decision boundary from different demographic groups, thereby facilitating the opportunity of equal recourse across demographic groups (which is their definition of fairness). 
\citet{vonKgelgen2020EqualRecourse2} extend this fairness notion when the distance between the recourse is measured in a causal manner.  
\citet{galhotra-sigmod-cfe} propose LEWIS that uses CFEs to identify racial bias in COMPAS and gender in Adult datasets. 
\citet{Dash-evaluate-bias-images} propose using CFEs to detect bias in image classifiers and counterfactual regularizer to counteract that bias. 

\begin{challenge}\label{ch:robust}
Generate robust counterfactual explanations \citep{another-robustness-cfe-ferrario, mishra-cfe-robustness-survey}. 
\end{challenge}
Counterfactual explanation optimization problems force the modified datapoint to obtain the desired class label. However, the modified datapoint could be labeled either in a robust manner or due to the classifier's non-robustness, e.g., an overfitted classifier. \citet{issues_posthoc} term this as the \emph{stability} property of a counterfactual.  
There are three kinds of robustness needs: 1) robustness to model changes when models are retrained, for example, 2) robustness to the input datapoint (two individuals with a slight change in features should be given similar CFEs), and 3) robustness to small changes in the attained CFE (a CFE with minor changes to the originally suggested CFE should also be accepted). 
%This can generate counterfactuals that might be non-sensical and have the desired class label only because of the classifier's artifact. 

%This is specifically a challenge for approaches that solve an optimization problem each time they generate a counterfactual (see \RCref{ch:scale}.)  %We see potential overlap between this nascent literature and the certifiability literature from the adversarial machine learning community. 

\progress
\citet{slack2021manipulated} underscore this challenge by showing that small perturbations in the input datapoints can result in drastically different CFEs. 
\citet{rawal2021robust1} further emphasize this challenge by empirically demonstrating the invalidation of already prescribed recourses when the ML model gets retrained on datasets with temporal or geospatial distribution shifts. 
\citet{Artelt2021Evaluating-robustness} evaluate the robustness of closest CFEs when contrasted with CFEs generated with the data manifold constraint. 
\citet{stress-test-creditscore} propose the framework to measure the robustness of models by purposing generated CFEs as adversarial attack datasets. 
\citet{robustness-KC-score-is-good} empirically show that non-robust CFEs encounter a higher cost of change when adverse perturbations are applied to the datapoint, thus concluding that robustness in CFEs should be considered. 

% Techniques proposing a solution. 
\citet{upadhyay2021robust2} propose a technique named ROAR that uses adversarial training to generate recourses robust to changes in an ML model that is retrained on a distributionally shifted training dataset. 
\citet{dominiguez-olmedo-cfe-robustness} show that the CFEs that just cross the decision boundary are usually non-robust and formulate an optimization problem that generates robust recourse for linear models and neural networks. 
\citet{probabilistically-robust-cfe-hima-group} propose a technique named PROBE that generates robust CFEs while letting the users decide the trade-off between the CFE invalidation risk and its cost. 
\citet{black2022consistent-robust-cfe} argue that robust CFEs should have high confidence neighborhoods with small Lipschitz constants, and propose a Stable Neighbor Search algorithm to that end. 
\citet{nguyen2022distributionally-robustcfe} propose an algorithm to generate robust CFEs by considering a distribution over the parameters of the model if retrained. 
\citet{robustness-metric-paper-jpmorgan} propose counterfactual stability (the lower bound of the predicted class probability for the sampled datapoints in the neighborhood of a given CFE) as a metric for filtering robust CFEs. 
\citet{robust-CFE-GNN} propose a technique to generate robust CFEs for graph neural networks.

% \citet{uncertainty-CFEs-robustness} conduct a study on CFEs generation and showed CFEs that are generated from 4 of 5 datasets have high epistemic uncertainty (not robust).

% Non-stationarity: data drift of the feature space, or the classifier changes over time.
\begin{challenge}\label{ch:dynamics}
Counterfactual explanations should handle dynamics (data drift, classifier update, applicant's utility function changing, etc.)
\end{challenge}
All counterfactual explanation papers we review assume that the underlying black box is monotonic and does not change over time. However, this might not be true; credit card companies and banks update their models as frequently as 12-18 months~\citep{bank-models}. Therefore counterfactual explanation algorithms should take data drift, the dynamism and non-monotonicity of the classifier into account.

\begin{challenge}\label{ch:user-prefs}
Counterfactual explanations should capture the applicant's preferences. 
\end{challenge}
Along with the distinction between mutable and immutable features (finely classified into actionable, mutable, and immutable), counterfactual explanations should also capture preferences specific to an applicant. This is important because the ease of changing different features can differ across applicants. 

\progress
\citet{mahajan_preserving_2020} captures the applicant's preferences using an oracle, but that is expensive and is still a challenge. \citet{hima-beyond-recourse-globalcfe} use the Bradley-Terry model to learn the pairwise cost for each feature pair and hence the preference among them. 
\citet{user-specific-cost} argue that assuming each user's cost of changing different features is the same is unrealistic. They propose asking for the user's cost function or computing the expectation by sampling cost functions from a distribution. 

\begin{challenge}
Counterfactual explanations should also inform the applicants about what must not change
\end{challenge}
Suppose a CFE advises someone to increase their \emph{income} but does not tell that their \emph{length of last employment} should not decrease. To increase their income, the applicant who switches to a higher-paying job may find themselves in a worse position than earlier. Thus by failing to disclose what must not change, an explanation may lead the applicant to an unsuccessful state \citep{hidden_assumptions}. This corroborates \RCref{ch:interactive}, whereby an applicant might be able to interact with a platform to see the effect of a potential real-world action they are considering taking to achieve the counterfactual state. 

\begin{challenge}\label{ch:privacy}
Preserving model privacy. 
\end{challenge}
Privacy attacks on ML models can come in two major forms: member inference and model extraction. 
Both of these privacy attacks can be enhanced due to the provision of CFEs. \citet{modelprivacy-CFE} empirically demonstrate that adversaries can train a surrogate model with very high fidelity to the original model (i.e., model extraction attack) with as few as 1,000 queries to the model (which is required during CFE generation). The problem is further aggravated when diverse CFEs are provided. 
\citet{modelprivacy-generalxai} have demonstrated that gradient-based explanations methods leak a lot of information and make the models vulnerable to membership inference attacks. 
\citet{Miura2021MEGEX-privacy-attack-cfe} propose MEGEX, a data-free model extraction attack that learns a surrogate model without access to its training data by training a generative model. 
\citet{DualCF-model-extraction} propose using the CFE of a CFE to train a surrogate model and show that it is more efficient in model extraction when compared to \citep{modelprivacy-CFE}.

\begin{challenge}\label{ch:fairwashing}
Guarding against fairwashing. 
\end{challenge}
\citet{ulrich-fairwashingxai2019} and \citet{ulrich-fairwashingxai2021} have pointed out the risk of an adversary using model explanations to rationalize a model's decisions and obscure its bias. It remains to be seen if the fair recourse approaches can guard against fairwashing.

\begin{challenge}
CFE interpretability with engineered features \citep{schleich2021geco}. 
\end{challenge}
Most current CFE approaches assume that the features they change are directly input to the ML model. This might not be the case -- it is known that model developers use highly engineered features for training the ML models. In this light, approaches need to be developed that take feature engineering into account (potentially a non-differentiable step). Approaches that work with black-box access will naturally be able to work in this setting.

\begin{challenge}
Handling of categorical features in counterfactual explanations
\end{challenge}
Different papers have come up with various methods to handle categorical features, like converting them to one-hot encoding and then enforcing the sum of those columns to be 1 using regularization or a hard constraint, or clamping an optimization problem to a specific categorical value, or leaving them to be automatically handled by genetic approaches and SMT solvers. 
Measuring distance in categorical features is also not obvious. Some papers use an indicator function, which equates to 1 for unequal values and 0 if the same; other papers convert to one-hot encoding and use standard distance metrics like L1/L2 norm, or use the distance in Markov chains \citep{forster-capturing-2021}. 
Therefore none of the methods developed to handle categorical features are obvious; future research must consider this and develop appropriate methods. 

\begin{challenge}\label{ch:user-study}
Evaluate counterfactual explanations using a user study. 
\end{challenge}
The evaluation for counterfactual explanations must be done using a user study because evaluation proxies (see~\cref{sec:eval}) might not be able to precisely capture the psychological and other intricacies of human cognition on the ease of actionability of a counterfactual. \citet{if_only_better_CFE_keane} emphasize the importance of user studies in the context of CFEs. 
\progress
\citet{forster2020evaluating-xai-user-study} conduct a user study with 144 participants to understand the format of explanation they prefer. They conclude that users prefer concrete, consistent, relevant explanations, and lengthy explanations if they are concrete. 
\citet{forster-capturing-2021} conduct a user study with 46 participants who were asked to rate the realisticness of the CFEs generated by theirs and a baseline approach. Using statistical tests, they concluded that the CFEs generated by their approach were perceived to be more real and typical. 
\citet{hima-beyond-recourse-globalcfe} conduct a user study with 21 participants who were asked to detect a bias in the recourse summaries for demographic groups. 
\citet{kanamori-decision-tree-globalcfe} conduct a user study with 35 participants to compare their global CFE generating technique with that of \citet{hima-beyond-recourse-globalcfe}. 
\citet{sequential-CFE} conduct a user study with 54 participants and found that most users prefer specific directives over generic and non-directive explanations. 
\citet{user-study-cfe-causal-diff-keane} conduct a user study with 127 participants and found that counterfactual explanations elicited higher trust and satisfaction than causal explanations. 
\citet{CFE-user-study-judges} conduct a user study with 8 U.S. state court judges to understand their response to CFEs from pretrial risk assessment instruments (PRAI). They conclude that judges ignored the CFEs and focused on the factual features of the defendant. 
\citet{keep-your-friends-close-cfe} conduct a user study with 74 users in an interactive game setting and found that users benefit less from receiving computationally plausible CFEs than the closest CFEs (measured using feature distance). 
\citet{Zhang-CFE-user-study} conduct a user study with 200 users to check their understanding of global, local, and CF explanations. 
\citet{cfe-user-study-quickdraw} conduct a user study on 1070 participants to understand how users perceive explanations when provided examples from the desired class vs. when provided examples from all other classes.

\begin{challenge}\label{ch:visual}
Counterfactual explanations should be integrated with data visualization interfaces. 
\end{challenge}
Counterfactual explanations will directly interact with consumers with varying technical knowledge levels; therefore, counterfactual generation algorithms should be integrated with visualization interfaces. We already know that visualization can influence human behavior \citep{Correll19:Ethical}, and a collaboration between machine learning and HCI communities could help address this challenge. 

\progress
\citet{what-if-tool, dece-visualcfe, vice-visualcfe, advice-visualcfe, Leung-XAI-Customer-Churn-Interface} have developed interactive graphical user interfaces for displaying CFEs. 
DECE~\citep{dece-visualcfe} also summarizes CFEs for subgroups that can help detect model biases, if any. 
\citet{Rivelo-visualcfe} develop a visualization tool for CFEs for text classification models. 
\citet{GAMUT} also build a visual interactive user interface for providing model explanations.

\begin{challenge}\label{ch:multi-agent}
Generating optimal recourses when considering a multi-agent scenario. 
\end{challenge}
\citet{multi-agent-recourse} demonstrate the non-optimality of recourses generated when a single agent's interest is considered in a multi-agent scenario like the prisoner's dilemma. In the real world, an agent's actions affect other agents, hence generating recourses that consider the interests of multiple agents would be useful. 

\begin{challenge}
Incentivize users to improve features in non-manipulative ways. 
\end{challenge}
An approach that provides a recourse to users might want to prevent the ``gamification'' of the model (when users manipulate simple features like the purpose of a loan to get approved). This also protects the ML models from adversarial robustness attacks. 

\progress
\citet{Chen2020Strategic-recourse-linear} propose the optimization objective for linear classification models when the goal is to develop an accurate model that encourages actual feature improvement for users. They categorize features into three categories: improvement, manipulative, and immutable. Users should be encouraged to change the improvement features, not the manipulative ones when optimizing for recourse. 
\citet{meaningful-recourse-konig} suggest using causality to generate meaningful recourses and prevent gamification of the model.

\begin{challenge}\label{ch:regulatory-colab}
Strengthen the ties between machine learning and regulatory communities. 
\end{challenge}
A joint statement between the machine learning community and regulatory community (OCC, Federal Reserve, FTC, CFPB) acknowledging successes and limitations of where counterfactual explanations will be adequate for legal and consumer-facing needs and would improve the adoption and use of counterfactual explanations in critical software. 

\progress
\citet{non-asimov-ai-regulation-paper} talk about how regulation and policies need to adapt to how ML models can explain their decisions. 

\section{Conclusions}
\label{sec:conclusions}
In this paper, we collected and reviewed \papers papers which proposed various algorithmic solutions to finding counterfactual explanations for the decisions produced by automated systems, specifically automated by machine learning. 
Evaluating all the papers on the same rubric helps in quickly understanding the peculiarities of different approaches and the advantages, and disadvantages of each of them, which can also help organizations choose the algorithm best suited to their application constraints. 
This has also helped us readily identify the gaps, which will be beneficial to researchers scouring for open problems in this space and quickly sifting the large body of literature. 
We hope this paper can also be the starting point for people wanting to get an introduction to the broad area of counterfactual explanations and guide them to proper resources for things they might be interested in. \\

\noindent\textbf{Acknowledgments.}  
We thank Jason Wittenbach, \href{https://homes.cs.washington.edu/~kusupati/}{Aditya Kusupati}, \href{https://divyat09.github.io}{Divyat Mahajan}, \href{http://jessicadai.com}{Jessica Dai}, \href{https://www.linkedin.com/in/singhalsoumye/}{Soumye Singhal}, \href{https://harshv834.github.io}{Harsh Vardhan}, and 
\href{http://web.mit.edu/jmmichel/www/}{Jesse Michel} for helpful comments. 

%Research performed in consultation to \href{https://www.arthur.ai}{Arthur AI}, who owns the resultant intellectual property. 
% \jpd{We'll be taking input from the community, so good to keep adding folks to Acks as they send comments in.}
% \SV{We should also add a URL their website (have clickable names), because there are a thousand Jason.}

% I had a question about how back in the time should be go to collect papers. There must be papers even in 1999, which talked about inverse classification as such. This would also be related to metamorphic testing or killing a bug in software engineering, therefore it is not clear to me what distinction should we make. 

\bibliographystyle{ACM-Reference-Format}
\bibliography{refs}


\begin{thebibliography}{357}

%%% the Web address.

\ifx \showCODEN    \undefined \def \showCODEN     #1{\unskip}     \fi
\ifx \showDOI      \undefined \def \showDOI       #1{#1}\fi
\ifx \showISBNx    \undefined \def \showISBNx     #1{\unskip}     \fi
\ifx \showISBNxiii \undefined \def \showISBNxiii  #1{\unskip}     \fi
\ifx \showISSN     \undefined \def \showISSN      #1{\unskip}     \fi
\ifx \showLCCN     \undefined \def \showLCCN      #1{\unskip}     \fi
\ifx \shownote     \undefined \def \shownote      #1{#1}          \fi
\ifx \showarticletitle \undefined \def \showarticletitle #1{#1}   \fi
\ifx \showURL      \undefined \def \showURL       {\relax}        \fi
% invisible to TeX
\providecommand\bibfield[2]{#2}
\providecommand\bibinfo[2]{#2}
\providecommand\natexlab[1]{#1}
\providecommand\showeprint[2][]{arXiv:#2}

\bibitem[\protect\citeauthoryear{Abid, Yuksekgonul, and Zou}{Abid
  et~al\mbox{.}}{2022}]%
        {conceptual-cfe-abid}
\bibfield{author}{\bibinfo{person}{Abubakar Abid}, \bibinfo{person}{Mert
  Yuksekgonul}, {and} \bibinfo{person}{James Zou}.}
  \bibinfo{year}{2022}\natexlab{}.
\newblock \showarticletitle{Meaningfully debugging model mistakes using
  conceptual counterfactual explanations}. In
  \bibinfo{booktitle}{\emph{Proceedings of the 39th International Conference on
  Machine Learning}} \emph{(\bibinfo{series}{Proceedings of Machine Learning
  Research})}, Vol.~\bibinfo{volume}{162}. \bibinfo{publisher}{PMLR},
  \bibinfo{pages}{66--88}.
\newblock
\urldef\tempurl%
\url{https://proceedings.mlr.press/v162/abid22a.html}
\showURL{%
\tempurl}


\bibitem[\protect\citeauthoryear{Abrate and Bonchi}{Abrate and Bonchi}{2021}]%
        {abrate-cfe-graph-brain-networks}
\bibfield{author}{\bibinfo{person}{Carlo Abrate} {and}
  \bibinfo{person}{Francesco Bonchi}.} \bibinfo{year}{2021}\natexlab{}.
\newblock \showarticletitle{Counterfactual Graphs for Explainable
  Classification of Brain Networks}. In \bibinfo{booktitle}{\emph{Proceedings
  of the 27th ACM SIGKDD Conference on Knowledge Discovery \& Data Mining}}
  \emph{(\bibinfo{series}{KDD '21})}. \bibinfo{publisher}{Association for
  Computing Machinery}, \bibinfo{address}{New York, NY, USA}, 10.
\newblock
\urldef\tempurl%
\url{https://doi.org/10.1145/3447548.3467154}
\showDOI{\tempurl}


\bibitem[\protect\citeauthoryear{Adadi and Berrada}{Adadi and Berrada}{2018}]%
        {xai-survey2}
\bibfield{author}{\bibinfo{person}{Amina Adadi} {and} \bibinfo{person}{Mohammed
  Berrada}.} \bibinfo{year}{2018}\natexlab{}.
\newblock \showarticletitle{Peeking inside the black-box: A survey on
  Explainable Artificial Intelligence (XAI)}.
\newblock \bibinfo{journal}{\emph{IEEE Access}}  \bibinfo{volume}{PP}
  (\bibinfo{date}{09} \bibinfo{year}{2018}), \bibinfo{pages}{1--1}.
\newblock
\urldef\tempurl%
\url{https://doi.org/10.1109/ACCESS.2018.2870052}
\showDOI{\tempurl}


\bibitem[\protect\citeauthoryear{Aggarwal, Chen, and Han}{Aggarwal
  et~al\mbox{.}}{2010}]%
        {inverse-classification1}
\bibfield{author}{\bibinfo{person}{Charu~C. Aggarwal}, \bibinfo{person}{Chen
  Chen}, {and} \bibinfo{person}{Jiawei Han}.} \bibinfo{year}{2010}\natexlab{}.
\newblock \showarticletitle{The Inverse Classification Problem}.
\newblock \bibinfo{journal}{\emph{J. Comput. Sci. Technol.}}
  \bibinfo{volume}{25}, \bibinfo{number}{3} (\bibinfo{date}{May}
  \bibinfo{year}{2010}), \bibinfo{pages}{458–468}.
\newblock
\urldef\tempurl%
\url{https://doi.org/10.1007/s11390-010-9337-x}
\showDOI{\tempurl}


\bibitem[\protect\citeauthoryear{Aivodji, Arai, Fortineau, Gambs, Hara, and
  Tapp}{Aivodji et~al\mbox{.}}{2019}]%
        {ulrich-fairwashingxai2019}
\bibfield{author}{\bibinfo{person}{Ulrich Aivodji}, \bibinfo{person}{Hiromi
  Arai}, \bibinfo{person}{Olivier Fortineau}, \bibinfo{person}{S{\'e}bastien
  Gambs}, \bibinfo{person}{Satoshi Hara}, {and} \bibinfo{person}{Alain Tapp}.}
  \bibinfo{year}{2019}\natexlab{}.
\newblock \showarticletitle{Fairwashing: the risk of rationalization}. In
  \bibinfo{booktitle}{\emph{Proceedings of the 36th International Conference on
  Machine Learning}} \emph{(\bibinfo{series}{Proceedings of Machine Learning
  Research})}, Vol.~\bibinfo{volume}{97}. \bibinfo{publisher}{PMLR},
  \bibinfo{pages}{161--170}.
\newblock
\urldef\tempurl%
\url{https://proceedings.mlr.press/v97/aivodji19a.html}
\showURL{%
\tempurl}


\bibitem[\protect\citeauthoryear{A\"{\i}vodji, Arai, Gambs, and
  Hara}{A\"{\i}vodji et~al\mbox{.}}{2021}]%
        {ulrich-fairwashingxai2021}
\bibfield{author}{\bibinfo{person}{Ulrich A\"{\i}vodji},
  \bibinfo{person}{Hiromi Arai}, \bibinfo{person}{S\'{e}bastien Gambs}, {and}
  \bibinfo{person}{Satoshi Hara}.} \bibinfo{year}{2021}\natexlab{}.
\newblock \showarticletitle{Characterizing the risk of fairwashing}. In
  \bibinfo{booktitle}{\emph{Advances in Neural Information Processing
  Systems}}, Vol.~\bibinfo{volume}{34}. \bibinfo{publisher}{Curran Associates,
  Inc.}, \bibinfo{pages}{14822--14834}.
\newblock
\urldef\tempurl%
\url{https://proceedings.neurips.cc/paper/2021/file/7caf5e22ea3eb8175ab518429c8589a4-Paper.pdf}
\showURL{%
\tempurl}


\bibitem[\protect\citeauthoryear{A{\"\i}vodji, Bolot, and Gambs}{A{\"\i}vodji
  et~al\mbox{.}}{2020}]%
        {modelprivacy-CFE}
\bibfield{author}{\bibinfo{person}{Ulrich A{\"\i}vodji},
  \bibinfo{person}{Alexandre Bolot}, {and} \bibinfo{person}{S{\'e}bastien
  Gambs}.} \bibinfo{year}{2020}\natexlab{}.
\newblock \showarticletitle{Model extraction from counterfactual explanations}.
\newblock \bibinfo{journal}{\emph{arXiv preprint arXiv:2009.01884}}
  (\bibinfo{year}{2020}).
\newblock


\bibitem[\protect\citeauthoryear{Akula, Wang, and Zhu}{Akula
  et~al\mbox{.}}{2020}]%
        {Akula-CFE-CoCoX-images}
\bibfield{author}{\bibinfo{person}{Arjun Akula}, \bibinfo{person}{Shuai Wang},
  {and} \bibinfo{person}{Song-Chun Zhu}.} \bibinfo{year}{2020}\natexlab{}.
\newblock \showarticletitle{CoCoX: Generating Conceptual and Counterfactual
  Explanations via Fault-Lines}.
\newblock \bibinfo{journal}{\emph{Proceedings of the AAAI Conference on
  Artificial Intelligence}} \bibinfo{volume}{34}, \bibinfo{number}{03}
  (\bibinfo{date}{Apr.} \bibinfo{year}{2020}), \bibinfo{pages}{2594--2601}.
\newblock
\urldef\tempurl%
\url{https://doi.org/10.1609/aaai.v34i03.5643}
\showDOI{\tempurl}


\bibitem[\protect\citeauthoryear{Akula, Wang, Liu, Saba-Sadiya, Lu, Todorovic,
  Chai, and Zhu}{Akula et~al\mbox{.}}{2022}]%
        {akula-cfe-CX-ToM-images-interactivity}
\bibfield{author}{\bibinfo{person}{Arjun~R. Akula}, \bibinfo{person}{Keze
  Wang}, \bibinfo{person}{Changsong Liu}, \bibinfo{person}{Sari Saba-Sadiya},
  \bibinfo{person}{Hongjing Lu}, \bibinfo{person}{Sinisa Todorovic},
  \bibinfo{person}{Joyce Chai}, {and} \bibinfo{person}{Song-Chun Zhu}.}
  \bibinfo{year}{2022}\natexlab{}.
\newblock \showarticletitle{CX-ToM: Counterfactual explanations with
  theory-of-mind for enhancing human trust in image recognition models}.
\newblock \bibinfo{journal}{\emph{iScience}} \bibinfo{volume}{25},
  \bibinfo{number}{1} (\bibinfo{year}{2022}), \bibinfo{pages}{103581}.
\newblock
\urldef\tempurl%
\url{https://doi.org/10.1016/j.isci.2021.103581}
\showDOI{\tempurl}


\bibitem[\protect\citeauthoryear{Albini, Long, Dervovic, and Magazzeni}{Albini
  et~al\mbox{.}}{2022}]%
        {CF-Shap-paper}
\bibfield{author}{\bibinfo{person}{Emanuele Albini}, \bibinfo{person}{Jason
  Long}, \bibinfo{person}{Danial Dervovic}, {and} \bibinfo{person}{Daniele
  Magazzeni}.} \bibinfo{year}{2022}\natexlab{}.
\newblock \showarticletitle{Counterfactual Shapley Additive Explanations}. In
  \bibinfo{booktitle}{\emph{2022 ACM Conference on Fairness, Accountability,
  and Transparency}} \emph{(\bibinfo{series}{FAccT '22})}.
  \bibinfo{publisher}{Association for Computing Machinery},
  \bibinfo{address}{New York, NY, USA}, 17.
\newblock
\urldef\tempurl%
\url{https://doi.org/10.1145/3531146.3533168}
\showDOI{\tempurl}


\bibitem[\protect\citeauthoryear{Albini, Rago, Baroni, and Toni}{Albini
  et~al\mbox{.}}{2021}]%
        {bayesian-network-CFE}
\bibfield{author}{\bibinfo{person}{Emanuele Albini}, \bibinfo{person}{Antonio
  Rago}, \bibinfo{person}{Pietro Baroni}, {and} \bibinfo{person}{Francesca
  Toni}.} \bibinfo{year}{2021}\natexlab{}.
\newblock \showarticletitle{Influence-Driven Explanations for Bayesian Network
  Classifiers}. In \bibinfo{booktitle}{\emph{PRICAI 2021}}.
  \bibinfo{publisher}{Springer-Verlag}, \bibinfo{address}{Berlin, Heidelberg},
  13.
\newblock
\urldef\tempurl%
\url{https://doi.org/10.1007/978-3-030-89188-6_7}
\showDOI{\tempurl}


\bibitem[\protect\citeauthoryear{Ali, Al-Obeidat, Tubaishat, Zia, Ilyas, and
  Rocha}{Ali et~al\mbox{.}}{2021}]%
        {Gohar-images-cfe-bayesian-model}
\bibfield{author}{\bibinfo{person}{Gohar Ali}, \bibinfo{person}{Feras
  Al-Obeidat}, \bibinfo{person}{Abdallah Tubaishat}, \bibinfo{person}{Tehseen
  Zia}, \bibinfo{person}{Muhammad Ilyas}, {and} \bibinfo{person}{Alvaro
  Rocha}.} \bibinfo{year}{2021}\natexlab{}.
\newblock \showarticletitle{Counterfactual explanation of Bayesian model
  uncertainty}.
\newblock \bibinfo{journal}{\emph{Neural Computing and Applications}}
  (\bibinfo{date}{Sept.} \bibinfo{year}{2021}).
\newblock
\urldef\tempurl%
\url{https://doi.org/10.1007/s00521-021-06528-z}
\showDOI{\tempurl}


\bibitem[\protect\citeauthoryear{Alipour, Ray, Lin, Cogswell, Schulze, Yao, and
  Burachas}{Alipour et~al\mbox{.}}{2021}]%
        {Alipour-visual-question-answering-cfe}
\bibfield{author}{\bibinfo{person}{Kamran Alipour}, \bibinfo{person}{Arijit
  Ray}, \bibinfo{person}{Xiao Lin}, \bibinfo{person}{Michael Cogswell},
  \bibinfo{person}{Jurgen~P. Schulze}, \bibinfo{person}{Yi Yao}, {and}
  \bibinfo{person}{Giedrius~T. Burachas}.} \bibinfo{year}{2021}\natexlab{}.
\newblock \showarticletitle{Improving users' mental model with
  attention-directed counterfactual edits}.
\newblock \bibinfo{journal}{\emph{Applied AI Letters}} \bibinfo{volume}{2},
  \bibinfo{number}{4} (\bibinfo{year}{2021}).
\newblock
\urldef\tempurl%
\url{https://doi.org/10.1002/ail2.47}
\showDOI{\tempurl}


\bibitem[\protect\citeauthoryear{Andrews, Diederich, and Tickle}{Andrews
  et~al\mbox{.}}{1995}]%
        {Andrews_exp6}
\bibfield{author}{\bibinfo{person}{Robert Andrews}, \bibinfo{person}{Joachim
  Diederich}, {and} \bibinfo{person}{Alan~B. Tickle}.}
  \bibinfo{year}{1995}\natexlab{}.
\newblock \showarticletitle{Survey and Critique of Techniques for Extracting
  Rules from Trained Artificial Neural Networks}.
\newblock \bibinfo{journal}{\emph{Know.-Based Syst.}} \bibinfo{volume}{8},
  \bibinfo{number}{6} (\bibinfo{year}{1995}), 17.
\newblock
\urldef\tempurl%
\url{https://doi.org/10.1016/0950-7051(96)81920-4}
\showDOI{\tempurl}


\bibitem[\protect\citeauthoryear{Antoran, Bhatt, Adel, Weller, and
  Hern{\'a}ndez-Lobato}{Antoran et~al\mbox{.}}{2021}]%
        {antoran2021-cfe-for-uncertainty-estimation}
\bibfield{author}{\bibinfo{person}{Javier Antoran}, \bibinfo{person}{Umang
  Bhatt}, \bibinfo{person}{Tameem Adel}, \bibinfo{person}{Adrian Weller}, {and}
  \bibinfo{person}{Jos{\'e}~Miguel Hern{\'a}ndez-Lobato}.}
  \bibinfo{year}{2021}\natexlab{}.
\newblock \showarticletitle{Getting a {\{}CLUE{\}}: A Method for Explaining
  Uncertainty Estimates}. In \bibinfo{booktitle}{\emph{International Conference
  on Learning Representations}}.
\newblock
\urldef\tempurl%
\url{https://openreview.net/forum?id=XSLF1XFq5h}
\showURL{%
\tempurl}


\bibitem[\protect\citeauthoryear{Apley and Zhu}{Apley and Zhu}{2020}]%
        {intro_ALE}
\bibfield{author}{\bibinfo{person}{Daniel Apley} {and} \bibinfo{person}{Jingyu
  Zhu}.} \bibinfo{year}{2020}\natexlab{}.
\newblock \showarticletitle{Visualizing the effects of predictor variables in
  black box supervised learning models}.
\newblock \bibinfo{journal}{\emph{Journal of the Royal Statistical Society:
  Series B (Statistical Methodology)}}  \bibinfo{volume}{82(4)}
  (\bibinfo{date}{06} \bibinfo{year}{2020}), \bibinfo{pages}{1059--1086}.
\newblock
\urldef\tempurl%
\url{https://doi.org/10.1111/rssb.12377}
\showDOI{\tempurl}


\bibitem[\protect\citeauthoryear{Artelt}{Artelt}{2021}]%
        {artelt-ceml-toolbox}
\bibfield{author}{\bibinfo{person}{André Artelt}.} \bibinfo{year}{2019 -
  2021}\natexlab{}.
\newblock \bibinfo{title}{CEML: Counterfactuals for Explaining Machine Learning
  models - A Python toolbox}.
\newblock
  \bibinfo{howpublished}{\url{https://www.github.com/andreArtelt/ceml}}.
\newblock


\bibitem[\protect\citeauthoryear{Artelt and Hammer}{Artelt and Hammer}{2019}]%
        {artelt_computation_2019}
\bibfield{author}{\bibinfo{person}{Andr{\'e} Artelt} {and}
  \bibinfo{person}{Barbara Hammer}.} \bibinfo{year}{2019}\natexlab{}.
\newblock \bibinfo{title}{On the computation of counterfactual explanations --
  A survey}.
\newblock
\newblock
\urldef\tempurl%
\url{http://arxiv.org/abs/1911.07749}
\showURL{%
\tempurl}


\bibitem[\protect\citeauthoryear{Artelt and Hammer}{Artelt and Hammer}{2020}]%
        {efficient-contrastive}
\bibfield{author}{\bibinfo{person}{André Artelt} {and}
  \bibinfo{person}{Barbara Hammer}.} \bibinfo{year}{2020}\natexlab{}.
\newblock \bibinfo{title}{Efficient computation of contrastive explanations}.
\newblock
\newblock
\urldef\tempurl%
\url{https://doi.org/10.48550/ARXIV.2010.02647}
\showDOI{\tempurl}


\bibitem[\protect\citeauthoryear{Artelt and Hammer}{Artelt and Hammer}{2021}]%
        {convex_optimization_cfe}
\bibfield{author}{\bibinfo{person}{Andr{\'e} Artelt} {and}
  \bibinfo{person}{Barbara Hammer}.} \bibinfo{year}{2021}\natexlab{}.
\newblock \bibinfo{title}{Convex optimization for actionable \& plausible
  counterfactual explanations}.
\newblock
\newblock
\urldef\tempurl%
\url{https://doi.org/10.48550/ARXIV.2105.07630}
\showDOI{\tempurl}


\bibitem[\protect\citeauthoryear{Artelt, Hinder, Vaquet, Feldhans, and
  Hammer}{Artelt et~al\mbox{.}}{2021a}]%
        {artelt-cfe-model-adaptation}
\bibfield{author}{\bibinfo{person}{Andr{\'e} Artelt}, \bibinfo{person}{Fabian
  Hinder}, \bibinfo{person}{Valerie Vaquet}, \bibinfo{person}{Robert Feldhans},
  {and} \bibinfo{person}{Barbara Hammer}.} \bibinfo{year}{2021}\natexlab{a}.
\newblock \showarticletitle{Contrastive Explanations for Explaining Model
  Adaptations}. In \bibinfo{booktitle}{\emph{Advances in Computational
  Intelligence}}. \bibinfo{publisher}{Springer International Publishing},
  \bibinfo{address}{Cham}, \bibinfo{pages}{101--112}.
\newblock


\bibitem[\protect\citeauthoryear{Artelt, Vaquet, Velioglu, Hinder, Brinkrolf,
  Schilling, and Hammer}{Artelt et~al\mbox{.}}{2021b}]%
        {Artelt2021Evaluating-robustness}
\bibfield{author}{\bibinfo{person}{Andr{\'e} Artelt}, \bibinfo{person}{Valerie
  Vaquet}, \bibinfo{person}{Riza Velioglu}, \bibinfo{person}{Fabian Hinder},
  \bibinfo{person}{Johannes Brinkrolf}, \bibinfo{person}{Malte Schilling},
  {and} \bibinfo{person}{Barbara Hammer}.} \bibinfo{year}{2021}\natexlab{b}.
\newblock \showarticletitle{Evaluating Robustness of Counterfactual
  Explanations}.
\newblock \bibinfo{journal}{\emph{2021 IEEE Symposium Series on Computational
  Intelligence (SSCI)}} (\bibinfo{year}{2021}), \bibinfo{pages}{01--09}.
\newblock


\bibitem[\protect\citeauthoryear{Asher, De~Lara, Paul, and Russell}{Asher
  et~al\mbox{.}}{2022}]%
        {cfe-fair-adequate-Asher}
\bibfield{author}{\bibinfo{person}{Nicholas Asher}, \bibinfo{person}{Lucas
  De~Lara}, \bibinfo{person}{Soumya Paul}, {and} \bibinfo{person}{Chris
  Russell}.} \bibinfo{year}{2022}\natexlab{}.
\newblock \showarticletitle{Counterfactual Models for Fair and Adequate
  Explanations}.
\newblock \bibinfo{journal}{\emph{Machine Learning and Knowledge Extraction}}
  \bibinfo{volume}{4}, \bibinfo{number}{2} (\bibinfo{year}{2022}),
  \bibinfo{pages}{316--349}.
\newblock
\urldef\tempurl%
\url{https://doi.org/10.3390/make4020014}
\showDOI{\tempurl}


\bibitem[\protect\citeauthoryear{Ates, Aksar, Leung, and Coskun}{Ates
  et~al\mbox{.}}{2021}]%
        {cfe_multivariate_time_series}
\bibfield{author}{\bibinfo{person}{Emre Ates}, \bibinfo{person}{Burak Aksar},
  \bibinfo{person}{Vitus~J. Leung}, {and} \bibinfo{person}{Ayse~K. Coskun}.}
  \bibinfo{year}{2021}\natexlab{}.
\newblock \showarticletitle{Counterfactual Explanations for Multivariate Time
  Series}. In \bibinfo{booktitle}{\emph{2021 International Conference on
  Applied Artificial Intelligence (ICAPAI)}}. \bibinfo{pages}{1--8}.
\newblock
\urldef\tempurl%
\url{https://doi.org/10.1109/ICAPAI49758.2021.9462056}
\showDOI{\tempurl}


\bibitem[\protect\citeauthoryear{Bacciu and Numeroso}{Bacciu and
  Numeroso}{2022}]%
        {input_perturbation_GNN4}
\bibfield{author}{\bibinfo{person}{Davide Bacciu} {and} \bibinfo{person}{Danilo
  Numeroso}.} \bibinfo{year}{2022}\natexlab{}.
\newblock \showarticletitle{Explaining Deep Graph Networks via Input
  Perturbation}.
\newblock \bibinfo{journal}{\emph{IEEE Transactions on Neural Networks and
  Learning Systems}} (\bibinfo{year}{2022}).
\newblock
\urldef\tempurl%
\url{https://doi.org/10.1109/TNNLS.2022.3165618}
\showDOI{\tempurl}


\bibitem[\protect\citeauthoryear{Bajaj, Chu, Xue, Pei, Wang, Lam, and
  Zhang}{Bajaj et~al\mbox{.}}{2021}]%
        {robust-CFE-GNN}
\bibfield{author}{\bibinfo{person}{Mohit Bajaj}, \bibinfo{person}{Lingyang
  Chu}, \bibinfo{person}{Zi~Yu Xue}, \bibinfo{person}{Jian Pei},
  \bibinfo{person}{Lanjun Wang}, \bibinfo{person}{Peter Cho-Ho Lam}, {and}
  \bibinfo{person}{Yong Zhang}.} \bibinfo{year}{2021}\natexlab{}.
\newblock \bibinfo{title}{Robust Counterfactual Explanations on Graph Neural
  Networks}.
\newblock
\newblock
\urldef\tempurl%
\url{https://doi.org/10.48550/ARXIV.2107.04086}
\showDOI{\tempurl}


\bibitem[\protect\citeauthoryear{Balasubramanian, Sharpe, Barr, Wittenbach, and
  Bruss}{Balasubramanian et~al\mbox{.}}{2020}]%
        {latent-cf_images}
\bibfield{author}{\bibinfo{person}{Rachana Balasubramanian},
  \bibinfo{person}{Samuel Sharpe}, \bibinfo{person}{Brian Barr},
  \bibinfo{person}{Jason Wittenbach}, {and} \bibinfo{person}{C.~Bayan Bruss}.}
  \bibinfo{year}{2020}\natexlab{}.
\newblock \bibinfo{title}{Latent-CF: A Simple Baseline for Reverse
  Counterfactual Explanations}.
\newblock
\newblock
\urldef\tempurl%
\url{https://doi.org/10.48550/ARXIV.2012.09301}
\showDOI{\tempurl}


\bibitem[\protect\citeauthoryear{Barocas, Selbst, and Raghavan}{Barocas
  et~al\mbox{.}}{2020}]%
        {hidden_assumptions}
\bibfield{author}{\bibinfo{person}{Solon Barocas}, \bibinfo{person}{Andrew~D.
  Selbst}, {and} \bibinfo{person}{Manish Raghavan}.}
  \bibinfo{year}{2020}\natexlab{}.
\newblock \showarticletitle{The Hidden Assumptions behind Counterfactual
  Explanations and Principal Reasons}. In \bibinfo{booktitle}{\emph{Proceedings
  of the Conference on Fairness, Accountability, and Transparency (FAccT)}}
  \emph{(\bibinfo{series}{FAT* '20})}. \bibinfo{publisher}{Association for
  Computing Machinery}, \bibinfo{address}{New York, NY, USA}, 10.
\newblock
\urldef\tempurl%
\url{https://doi.org/10.1145/3351095.3372830}
\showDOI{\tempurl}


\bibitem[\protect\citeauthoryear{Barr, Harrington, Sharpe, and Bruss}{Barr
  et~al\mbox{.}}{2021}]%
        {mahajan-work-extension-linear-interpolation}
\bibfield{author}{\bibinfo{person}{Brian Barr}, \bibinfo{person}{Matthew~R.
  Harrington}, \bibinfo{person}{Samuel Sharpe}, {and} \bibinfo{person}{C.~Bayan
  Bruss}.} \bibinfo{year}{2021}\natexlab{}.
\newblock \bibinfo{title}{Counterfactual Explanations via Latent Space
  Projection and Interpolation}.
\newblock
\newblock
\urldef\tempurl%
\url{https://doi.org/10.48550/ARXIV.2112.00890}
\showDOI{\tempurl}


\bibitem[\protect\citeauthoryear{Bas}{Bas}{1980}]%
        {VanFraassenBas1980:phil5}
\bibfield{author}{\bibinfo{person}{C.~Van~Fraassen Bas}.}
  \bibinfo{year}{1980}\natexlab{}.
\newblock \bibinfo{booktitle}{\emph{The Scientific Image}}.
\newblock \bibinfo{publisher}{Oxford University Press}.
\newblock


\bibitem[\protect\citeauthoryear{Beckers}{Beckers}{2022}]%
        {beckers-causal-xai}
\bibfield{author}{\bibinfo{person}{Sander Beckers}.}
  \bibinfo{year}{2022}\natexlab{}.
\newblock \bibinfo{title}{Causal Explanations and XAI}.
\newblock
\newblock
\urldef\tempurl%
\url{https://doi.org/10.48550/ARXIV.2201.13169}
\showDOI{\tempurl}


\bibitem[\protect\citeauthoryear{Bertossi}{Bertossi}{2020}]%
        {declarative-CFE}
\bibfield{author}{\bibinfo{person}{Leopoldo~E. Bertossi}.}
  \bibinfo{year}{2020}\natexlab{}.
\newblock \bibinfo{title}{Declarative Approaches to Counterfactual Explanations
  for Classification}.
\newblock
\newblock


\bibitem[\protect\citeauthoryear{Binns, Van~Kleek, Veale, Lyngs, Zhao, and
  Shadbolt}{Binns et~al\mbox{.}}{2018}]%
        {Binns:2018}
\bibfield{author}{\bibinfo{person}{Reuben Binns}, \bibinfo{person}{Max
  Van~Kleek}, \bibinfo{person}{Michael Veale}, \bibinfo{person}{Ulrik Lyngs},
  \bibinfo{person}{Jun Zhao}, {and} \bibinfo{person}{Nigel Shadbolt}.}
  \bibinfo{year}{2018}\natexlab{}.
\newblock \showarticletitle{'It's Reducing a Human Being to a Percentage':
  Perceptions of Justice in Algorithmic Decisions}. In
  \bibinfo{booktitle}{\emph{Proceedings of the 2018 CHI Conference on Human
  Factors in Computing Systems}} \emph{(\bibinfo{series}{CHI '18})}.
  \bibinfo{publisher}{Association for Computing Machinery},
  \bibinfo{address}{New York, NY, USA}, 14.
\newblock
\urldef\tempurl%
\url{https://doi.org/10.1145/3173574.3173951}
\showDOI{\tempurl}


\bibitem[\protect\citeauthoryear{Black, Wang, and Fredrikson}{Black
  et~al\mbox{.}}{2022}]%
        {black2022consistent-robust-cfe}
\bibfield{author}{\bibinfo{person}{Emily Black}, \bibinfo{person}{Zifan Wang},
  {and} \bibinfo{person}{Matt Fredrikson}.} \bibinfo{year}{2022}\natexlab{}.
\newblock \showarticletitle{Consistent Counterfactuals for Deep Models}. In
  \bibinfo{booktitle}{\emph{International Conference on Learning
  Representations}}.
\newblock
\urldef\tempurl%
\url{https://arxiv.org/abs/2110.03109}
\showURL{%
\tempurl}


\bibitem[\protect\citeauthoryear{Blanchart}{Blanchart}{2021}]%
        {Pierre-tree-ensemble-pure-region}
\bibfield{author}{\bibinfo{person}{Pierre Blanchart}.}
  \bibinfo{year}{2021}\natexlab{}.
\newblock \bibinfo{title}{An exact counterfactual-example-based approach to
  tree-ensemble models interpretability}.
\newblock
\newblock
\urldef\tempurl%
\url{https://doi.org/10.48550/ARXIV.2105.14820}
\showDOI{\tempurl}


\bibitem[\protect\citeauthoryear{Boch and Lieberman}{Boch and
  Lieberman}{1970}]%
        {lsat-data}
\bibfield{author}{\bibinfo{person}{R.~D. Boch} {and} \bibinfo{person}{M.
  Lieberman}.} \bibinfo{year}{1970}\natexlab{}.
\newblock \showarticletitle{Fitting a response model for n dichotomously scored
  items}.
\newblock \bibinfo{journal}{\emph{Psychometrika}}  \bibinfo{volume}{35}
  (\bibinfo{year}{1970}), \bibinfo{pages}{179--97}.
\newblock


\bibitem[\protect\citeauthoryear{Bordt, Finck, Raidl, and von Luxburg}{Bordt
  et~al\mbox{.}}{2022}]%
        {post-hoc-CFE-not-good}
\bibfield{author}{\bibinfo{person}{Sebastian Bordt},
  \bibinfo{person}{Mich{\`{e}}le Finck}, \bibinfo{person}{Eric Raidl}, {and}
  \bibinfo{person}{Ulrike von Luxburg}.} \bibinfo{year}{2022}\natexlab{}.
\newblock \bibinfo{title}{Post-Hoc Explanations Fail to Achieve their Purpose
  in Adversarial Contexts}.
\newblock
\newblock
\urldef\tempurl%
\url{https://arxiv.org/abs/2201.10295}
\showURL{%
\tempurl}


\bibitem[\protect\citeauthoryear{Boukhers, Hartmann, and Jürjens}{Boukhers
  et~al\mbox{.}}{2022}]%
        {CFE-COIN-VQA}
\bibfield{author}{\bibinfo{person}{Zeyd Boukhers}, \bibinfo{person}{Timo
  Hartmann}, {and} \bibinfo{person}{Jan Jürjens}.}
  \bibinfo{year}{2022}\natexlab{}.
\newblock \bibinfo{title}{COIN: Counterfactual Image Generation for VQA
  Interpretation}.
\newblock
\newblock
\urldef\tempurl%
\url{https://doi.org/10.48550/ARXIV.2201.03342}
\showDOI{\tempurl}


\bibitem[\protect\citeauthoryear{Brandão, Canal, Krivić, Luff, and
  Coles}{Brandão et~al\mbox{.}}{2021}]%
        {brandao-cfe-agent-motion-planning}
\bibfield{author}{\bibinfo{person}{Martim Brandão}, \bibinfo{person}{Gerard
  Canal}, \bibinfo{person}{Senka Krivić}, \bibinfo{person}{Paul Luff}, {and}
  \bibinfo{person}{Amanda Coles}.} \bibinfo{year}{2021}\natexlab{}.
\newblock \showarticletitle{How experts explain motion planner output: a
  preliminary user-study to inform the design of explainable planners}. In
  \bibinfo{booktitle}{\emph{2021 30th IEEE International Conference on Robot \&
  Human Interactive Communication (RO-MAN)}}. \bibinfo{pages}{299--306}.
\newblock
\urldef\tempurl%
\url{https://doi.org/10.1109/RO-MAN50785.2021.9515407}
\showDOI{\tempurl}


\bibitem[\protect\citeauthoryear{Brown, Talbert, and Talbert}{Brown
  et~al\mbox{.}}{2021}]%
        {Brown_Talbert_cfe-anomalous}
\bibfield{author}{\bibinfo{person}{Katherine~Elizabeth Brown},
  \bibinfo{person}{Doug Talbert}, {and} \bibinfo{person}{Steve Talbert}.}
  \bibinfo{year}{2021}\natexlab{}.
\newblock \showarticletitle{The Uncertainty of Counterfactuals in Deep
  Learning}.
\newblock \bibinfo{journal}{\emph{The International FLAIRS Conference
  Proceedings}}  \bibinfo{volume}{34} (\bibinfo{year}{2021}).
\newblock
\urldef\tempurl%
\url{https://doi.org/10.32473/flairs.v34i1.128795}
\showDOI{\tempurl}


\bibitem[\protect\citeauthoryear{Browne and Swift}{Browne and Swift}{2020}]%
        {semantic-explanation_adversarial_cfe_diff}
\bibfield{author}{\bibinfo{person}{Kieran Browne} {and} \bibinfo{person}{Ben
  Swift}.} \bibinfo{year}{2020}\natexlab{}.
\newblock \bibinfo{title}{Semantics and explanation: why counterfactual
  explanations produce adversarial examples in deep neural networks}.
\newblock
\newblock
\urldef\tempurl%
\url{https://doi.org/10.48550/ARXIV.2012.10076}
\showDOI{\tempurl}


\bibitem[\protect\citeauthoryear{Brughmans and Martens}{Brughmans and
  Martens}{2021}]%
        {nice_cfe}
\bibfield{author}{\bibinfo{person}{Dieter Brughmans} {and}
  \bibinfo{person}{David Martens}.} \bibinfo{year}{2021}\natexlab{}.
\newblock \bibinfo{title}{NICE: An Algorithm for Nearest Instance
  Counterfactual Explanations}.
\newblock
\newblock
\urldef\tempurl%
\url{https://doi.org/10.48550/ARXIV.2104.07411}
\showDOI{\tempurl}


\bibitem[\protect\citeauthoryear{Bueff, Cytryński, Calabrese, Jones, Roberts,
  Moore, and Brown}{Bueff et~al\mbox{.}}{2022}]%
        {stress-test-creditscore}
\bibfield{author}{\bibinfo{person}{Andreas~C. Bueff}, \bibinfo{person}{Mateusz
  Cytryński}, \bibinfo{person}{Raffaella Calabrese}, \bibinfo{person}{Matthew
  Jones}, \bibinfo{person}{John Roberts}, \bibinfo{person}{Jonathon Moore},
  {and} \bibinfo{person}{Iain Brown}.} \bibinfo{year}{2022}\natexlab{}.
\newblock \showarticletitle{Machine learning interpretability for a stress
  scenario generation in credit scoring based on counterfactuals}.
\newblock \bibinfo{journal}{\emph{Expert Systems with Applications}}
  \bibinfo{volume}{202} (\bibinfo{year}{2022}).
\newblock
\showISSN{0957-4174}
\urldef\tempurl%
\url{https://doi.org/10.1016/j.eswa.2022.117271}
\showDOI{\tempurl}


\bibitem[\protect\citeauthoryear{Bui, Nguyen, and Nguyen}{Bui
  et~al\mbox{.}}{2022}]%
        {nguyen2022distributionally-robustcfe}
\bibfield{author}{\bibinfo{person}{Ngoc Bui}, \bibinfo{person}{Duy Nguyen},
  {and} \bibinfo{person}{Viet~Anh Nguyen}.} \bibinfo{year}{2022}\natexlab{}.
\newblock \bibinfo{title}{Counterfactual Plans under Distributional Ambiguity}.
\newblock
\newblock
\urldef\tempurl%
\url{https://doi.org/10.48550/ARXIV.2201.12487}
\showDOI{\tempurl}


\bibitem[\protect\citeauthoryear{Byrne}{Byrne}{2008}]%
        {Byrne:psycho1}
\bibfield{author}{\bibinfo{person}{Ruth Byrne}.}
  \bibinfo{year}{2008}\natexlab{}.
\newblock \showarticletitle{The Rational Imagination: How People Create
  Alternatives to Reality}.
\newblock \bibinfo{journal}{\emph{The Behavioral and brain sciences}}
  \bibinfo{volume}{30} (\bibinfo{date}{12} \bibinfo{year}{2008}),
  \bibinfo{pages}{439--53; discussion 453}.
\newblock
\urldef\tempurl%
\url{https://doi.org/10.1017/S0140525X07002579}
\showDOI{\tempurl}


\bibitem[\protect\citeauthoryear{Byrne}{Byrne}{2019}]%
        {Byrne2019:psycho2}
\bibfield{author}{\bibinfo{person}{Ruth M.~J. Byrne}.}
  \bibinfo{year}{2019}\natexlab{}.
\newblock \showarticletitle{Counterfactuals in Explainable Artificial
  Intelligence (XAI): Evidence from Human Reasoning}. In
  \bibinfo{booktitle}{\emph{Proceedings of the Twenty-Eighth International
  Joint Conference on Artificial Intelligence, {IJCAI-19}}}.
  \bibinfo{publisher}{International Joint Conferences on Artificial
  Intelligence Organization}, \bibinfo{address}{California, USA},
  \bibinfo{pages}{6276--6282}.
\newblock
\urldef\tempurl%
\url{https://doi.org/10.24963/ijcai.2019/876}
\showURL{%
\tempurl}


\bibitem[\protect\citeauthoryear{Cai, Jongejan, and Holbrook}{Cai
  et~al\mbox{.}}{2019}]%
        {cfe-user-study-quickdraw}
\bibfield{author}{\bibinfo{person}{Carrie~J. Cai}, \bibinfo{person}{Jonas
  Jongejan}, {and} \bibinfo{person}{Jess Holbrook}.}
  \bibinfo{year}{2019}\natexlab{}.
\newblock \showarticletitle{The Effects of Example-Based Explanations in a
  Machine Learning Interface}. In \bibinfo{booktitle}{\emph{Proceedings of the
  24th International Conference on Intelligent User Interfaces}}
  \emph{(\bibinfo{series}{IUI '19})}. \bibinfo{publisher}{Association for
  Computing Machinery}, \bibinfo{address}{New York, NY, USA},
  \bibinfo{pages}{258–262}.
\newblock
\urldef\tempurl%
\url{https://doi.org/10.1145/3301275.3302289}
\showDOI{\tempurl}


\bibitem[\protect\citeauthoryear{Carreira-Perpiñán and
  Hada}{Carreira-Perpiñán and Hada}{2021}]%
        {oblique-tree-cfe}
\bibfield{author}{\bibinfo{person}{Miguel~Á. Carreira-Perpiñán} {and}
  \bibinfo{person}{Suryabhan~Singh Hada}.} \bibinfo{year}{2021}\natexlab{}.
\newblock \showarticletitle{Counterfactual Explanations for Oblique Decision
  Trees: Exact, Efficient Algorithms}.
\newblock \bibinfo{journal}{\emph{Proceedings of the AAAI Conference on
  Artificial Intelligence}}  \bibinfo{volume}{35} (\bibinfo{date}{May}
  \bibinfo{year}{2021}), \bibinfo{pages}{6903--6911}.
\newblock
\urldef\tempurl%
\url{https://doi.org/10.1609/aaai.v35i8.16851}
\showDOI{\tempurl}


\bibitem[\protect\citeauthoryear{Carrizosa, Ramirez-Ayerbe, and
  Romero~Morales}{Carrizosa et~al\mbox{.}}{2021}]%
        {carrizosa-global-cfe-tree-models}
\bibfield{author}{\bibinfo{person}{Emilio Carrizosa}, \bibinfo{person}{Jasone
  Ramirez-Ayerbe}, {and} \bibinfo{person}{Dolores Romero~Morales}.}
  \bibinfo{year}{2021}\natexlab{}.
\newblock \bibinfo{title}{Generating Collective Counterfactual Explanations in
  Score-Based Classification via Mathematical Optimization}.
\newblock
\newblock
\urldef\tempurl%
\url{https://doi.org/10.13140/RG.2.2.22996.12168/1}
\showDOI{\tempurl}


\bibitem[\protect\citeauthoryear{Carrizosa, Ramírez-Ayerbe, and
  Romero~Morales}{Carrizosa et~al\mbox{.}}{2022}]%
        {carrizosa-cfe-functional}
\bibfield{author}{\bibinfo{person}{Emilio Carrizosa}, \bibinfo{person}{Jasone
  Ramírez-Ayerbe}, {and} \bibinfo{person}{Dolores Romero~Morales}.}
  \bibinfo{year}{2022}\natexlab{}.
\newblock \bibinfo{title}{Counterfactual Explanations for Functional Data: A
  Mathematical Optimization Approach}.
\newblock
\newblock
\urldef\tempurl%
\url{https://doi.org/10.13140/RG.2.2.25682.68801}
\showDOI{\tempurl}


\bibitem[\protect\citeauthoryear{Carvalho, Pereira, and Cardoso}{Carvalho
  et~al\mbox{.}}{2019}]%
        {carvalho2019:survey3}
\bibfield{author}{\bibinfo{person}{Diogo~V Carvalho},
  \bibinfo{person}{Eduardo~M Pereira}, {and} \bibinfo{person}{Jaime~S
  Cardoso}.} \bibinfo{year}{2019}\natexlab{}.
\newblock \showarticletitle{Machine learning interpretability: A survey on
  methods and metrics}.
\newblock \bibinfo{journal}{\emph{Electronics}} \bibinfo{volume}{8},
  \bibinfo{number}{8} (\bibinfo{year}{2019}), \bibinfo{pages}{832}.
\newblock


\bibitem[\protect\citeauthoryear{CFPB}{CFPB}{[n. d.]a}]%
        {ECOA2}
\bibfield{author}{\bibinfo{person}{CFPB}.} \bibinfo{year}{[n. d.]}\natexlab{a}.
\newblock \bibinfo{title}{Adverse Action Notice Requirements Under the ECOA and
  the FCRA}.
\newblock
  \bibinfo{howpublished}{\url{https://consumercomplianceoutlook.org/2013/second-quarter/adverse-action-notice-requirements-under-ecoa-fcra/}}.
\newblock
\newblock
\shownote{Accessed: 2020-10-15.}


\bibitem[\protect\citeauthoryear{CFPB}{CFPB}{[n. d.]b}]%
        {ECOA1}
\bibfield{author}{\bibinfo{person}{CFPB}.} \bibinfo{year}{[n. d.]}\natexlab{b}.
\newblock \bibinfo{title}{Notification of action taken, ECOA notice, and
  statement of specific reasons}.
\newblock
  \bibinfo{howpublished}{\url{https://www.consumerfinance.gov/policy-compliance/rulemaking/regulations/1002/9/}}.
\newblock
\newblock
\shownote{Accessed: 2020-10-15.}


\bibitem[\protect\citeauthoryear{Chen, Ji, Zeng, Li, Zhang, Chen, and
  Zhang}{Chen et~al\mbox{.}}{2021a}]%
        {chen-2021-kace-cfe-text}
\bibfield{author}{\bibinfo{person}{Qianglong Chen}, \bibinfo{person}{Feng Ji},
  \bibinfo{person}{Xiangji Zeng}, \bibinfo{person}{Feng-Lin Li},
  \bibinfo{person}{Ji Zhang}, \bibinfo{person}{Haiqing Chen}, {and}
  \bibinfo{person}{Yin Zhang}.} \bibinfo{year}{2021}\natexlab{a}.
\newblock \showarticletitle{{KACE}: Generating Knowledge Aware Contrastive
  Explanations for Natural Language Inference}. In
  \bibinfo{booktitle}{\emph{Proceedings of the 59th Annual Meeting of the
  Association for Computational Linguistics and the 11th International Joint
  Conference on Natural Language Processing (Volume 1: Long Papers)}}.
  \bibinfo{publisher}{Association for Computational Linguistics},
  \bibinfo{address}{Online}, \bibinfo{pages}{2516--2527}.
\newblock
\urldef\tempurl%
\url{https://doi.org/10.18653/v1/2021.acl-long.196}
\showDOI{\tempurl}


\bibitem[\protect\citeauthoryear{Chen, Kuo, Liu, Poon, Towey, Tse, and
  Zhou}{Chen et~al\mbox{.}}{2018}]%
        {metamorphic-testing}
\bibfield{author}{\bibinfo{person}{Tsong~Yueh Chen}, \bibinfo{person}{Fei-Ching
  Kuo}, \bibinfo{person}{Huai Liu}, \bibinfo{person}{Pak-Lok Poon},
  \bibinfo{person}{Dave Towey}, \bibinfo{person}{T.~H. Tse}, {and}
  \bibinfo{person}{Zhi~Quan Zhou}.} \bibinfo{year}{2018}\natexlab{}.
\newblock \showarticletitle{Metamorphic Testing: A Review of Challenges and
  Opportunities}.
\newblock \bibinfo{journal}{\emph{ACM Comput. Surv.}} \bibinfo{volume}{51},
  \bibinfo{number}{1} (\bibinfo{year}{2018}), 27.
\newblock
\urldef\tempurl%
\url{https://doi.org/10.1145/3143561}
\showDOI{\tempurl}


\bibitem[\protect\citeauthoryear{Chen, Wang, and Liu}{Chen
  et~al\mbox{.}}{2020}]%
        {Chen2020Strategic-recourse-linear}
\bibfield{author}{\bibinfo{person}{Yatong Chen}, \bibinfo{person}{Jialu Wang},
  {and} \bibinfo{person}{Yang Liu}.} \bibinfo{year}{2020}\natexlab{}.
\newblock \bibinfo{title}{Strategic Recourse in Linear Classification}.
\newblock
\newblock


\bibitem[\protect\citeauthoryear{Chen, Silvestri, Wang, Zhu, Ahn, and
  Tolomei}{Chen et~al\mbox{.}}{2021b}]%
        {RL-for-CFE-relace-paper}
\bibfield{author}{\bibinfo{person}{Ziheng Chen}, \bibinfo{person}{Fabrizio
  Silvestri}, \bibinfo{person}{Jia Wang}, \bibinfo{person}{He Zhu},
  \bibinfo{person}{Hongshik Ahn}, {and} \bibinfo{person}{Gabriele Tolomei}.}
  \bibinfo{year}{2021}\natexlab{b}.
\newblock \bibinfo{title}{ReLAX: Reinforcement Learning Agent eXplainer for
  Arbitrary Predictive Models}.
\newblock
\newblock
\urldef\tempurl%
\url{https://doi.org/10.48550/ARXIV.2110.11960}
\showDOI{\tempurl}


\bibitem[\protect\citeauthoryear{Cheng, Ming, and Qu}{Cheng
  et~al\mbox{.}}{2020}]%
        {dece-visualcfe}
\bibfield{author}{\bibinfo{person}{Furui Cheng}, \bibinfo{person}{Yao Ming},
  {and} \bibinfo{person}{Huamin Qu}.} \bibinfo{year}{2020}\natexlab{}.
\newblock \bibinfo{title}{DECE: Decision Explorer with Counterfactual
  Explanations for Machine Learning Models}.
\newblock
\newblock
\showeprint[arxiv]{cs.LG/2008.08353}


\bibitem[\protect\citeauthoryear{Codella, Rotemberg, Tschandl, Celebi, Dusza,
  Gutman, Helba, Kalloo, Liopyris, Marchetti, Kittler, and Halpern}{Codella
  et~al\mbox{.}}{2019}]%
        {isic-skin-data}
\bibfield{author}{\bibinfo{person}{Noel Codella}, \bibinfo{person}{Veronica
  Rotemberg}, \bibinfo{person}{Philipp Tschandl}, \bibinfo{person}{M.~Emre
  Celebi}, \bibinfo{person}{Stephen Dusza}, \bibinfo{person}{David Gutman},
  \bibinfo{person}{Brian Helba}, \bibinfo{person}{Aadi Kalloo},
  \bibinfo{person}{Konstantinos Liopyris}, \bibinfo{person}{Michael Marchetti},
  \bibinfo{person}{Harald Kittler}, {and} \bibinfo{person}{Allan Halpern}.}
  \bibinfo{year}{2019}\natexlab{}.
\newblock \bibinfo{title}{Skin Lesion Analysis Toward Melanoma Detection 2018:
  A Challenge Hosted by the International Skin Imaging Collaboration (ISIC)}.
\newblock
\newblock
\urldef\tempurl%
\url{https://doi.org/10.48550/ARXIV.1902.03368}
\showDOI{\tempurl}


\bibitem[\protect\citeauthoryear{Cohen, Afshar, Tapson, and van Schaik}{Cohen
  et~al\mbox{.}}{2017}]%
        {EMNIST-data}
\bibfield{author}{\bibinfo{person}{Gregory Cohen}, \bibinfo{person}{Saeed
  Afshar}, \bibinfo{person}{Jonathan~C. Tapson}, {and}
  \bibinfo{person}{Andr{\'e} van Schaik}.} \bibinfo{year}{2017}\natexlab{}.
\newblock \showarticletitle{EMNIST: Extending MNIST to handwritten letters}.
\newblock \bibinfo{journal}{\emph{2017 International Joint Conference on Neural
  Networks (IJCNN)}} (\bibinfo{year}{2017}), \bibinfo{pages}{2921--2926}.
\newblock


\bibitem[\protect\citeauthoryear{Commission}{Commission}{[n. d.]a}]%
        {EU-XAIfund1}
\bibfield{author}{\bibinfo{person}{European Commission}.} \bibinfo{year}{[n.
  d.]}\natexlab{a}.
\newblock \bibinfo{title}{Artificial Intelligence}.
\newblock
  \bibinfo{howpublished}{\url{https://ec.europa.eu/info/funding-tenders/opportunities/portal/screen/opportunities/topic-details/ict-26-2018-2020}}.
\newblock
\newblock
\shownote{Accessed: 2020-10-15.}


\bibitem[\protect\citeauthoryear{Commission}{Commission}{[n. d.]b}]%
        {GDPR}
\bibfield{author}{\bibinfo{person}{European Commission}.} \bibinfo{year}{[n.
  d.]}\natexlab{b}.
\newblock \bibinfo{title}{REGULATION (EU) 2016/679 OF THE EUROPEAN PARLIAMENT
  AND OF THE COUNCIL of 27 April 2016 on the protection of natural persons with
  regard to the processing of personal data and on the free movement of such
  data, and repealing Directive 95/46/EC (General Data Protection Regulation)}.
\newblock
  \bibinfo{howpublished}{\url{https://eur-lex.europa.eu/eli/reg/2016/679/oj}}.
\newblock
\newblock
\shownote{Accessed: 2020-10-15.}


\bibitem[\protect\citeauthoryear{Cong, Chu, Yang, and Pei}{Cong
  et~al\mbox{.}}{2021}]%
        {Cong-CFE-KS-test}
\bibfield{author}{\bibinfo{person}{Zicun Cong}, \bibinfo{person}{Lingyang Chu},
  \bibinfo{person}{Yu Yang}, {and} \bibinfo{person}{Jian Pei}.}
  \bibinfo{year}{2021}\natexlab{}.
\newblock \showarticletitle{Comprehensible Counterfactual Explanation on
  Kolmogorov-Smirnov Test}.
\newblock \bibinfo{journal}{\emph{Proc. VLDB Endow.}} \bibinfo{volume}{14},
  \bibinfo{number}{9} (\bibinfo{year}{2021}), \bibinfo{pages}{1583–1596}.
\newblock
\showISSN{2150-8097}
\urldef\tempurl%
\url{https://doi.org/10.14778/3461535.3461546}
\showDOI{\tempurl}


\bibitem[\protect\citeauthoryear{Correll}{Correll}{2019}]%
        {Correll19:Ethical}
\bibfield{author}{\bibinfo{person}{Michael Correll}.}
  \bibinfo{year}{2019}\natexlab{}.
\newblock \showarticletitle{Ethical Dimensions of Visualization Research}. In
  \bibinfo{booktitle}{\emph{Proceedings of the 2019 CHI Conference on Human
  Factors in Computing Systems}} \emph{(\bibinfo{series}{CHI '19})}.
  \bibinfo{publisher}{Association for Computing Machinery},
  \bibinfo{address}{New York, NY, USA}, \bibinfo{pages}{1–13}.
\newblock
\showISBNx{9781450359702}
\urldef\tempurl%
\url{https://doi.org/10.1145/3290605.3300418}
\showDOI{\tempurl}


\bibitem[\protect\citeauthoryear{Craven and Shavlik}{Craven and
  Shavlik}{1995}]%
        {craven_exp1}
\bibfield{author}{\bibinfo{person}{Mark~W. Craven} {and}
  \bibinfo{person}{Jude~W. Shavlik}.} \bibinfo{year}{1995}\natexlab{}.
\newblock \showarticletitle{Extracting Tree-Structured Representations of
  Trained Networks}. In \bibinfo{booktitle}{\emph{Conference on Neural
  Information Processing Systems (NeurIPS)}}
  \emph{(\bibinfo{series}{NIPS'95})}. \bibinfo{publisher}{MIT Press},
  \bibinfo{address}{Cambridge, MA, USA}, \bibinfo{pages}{24–30}.
\newblock


\bibitem[\protect\citeauthoryear{Crupi., {San Miguel González}., Castelnovo.,
  and Regoli.}{Crupi. et~al\mbox{.}}{2022}]%
        {cf-causal-latent-ontop}
\bibfield{author}{\bibinfo{person}{Riccardo Crupi.}, \bibinfo{person}{Beatriz
  {San Miguel González}.}, \bibinfo{person}{Alessandro Castelnovo.}, {and}
  \bibinfo{person}{Daniele Regoli.}} \bibinfo{year}{2022}\natexlab{}.
\newblock \showarticletitle{Leveraging Causal Relations to Provide
  Counterfactual Explanations and Feasible Recommendations to End Users}. In
  \bibinfo{booktitle}{\emph{Proceedings of the 14th International Conference on
  Agents and Artificial Intelligence - Volume 2: ICAART,}}.
  \bibinfo{publisher}{SciTePress}, \bibinfo{pages}{24--32}.
\newblock
\urldef\tempurl%
\url{https://doi.org/10.5220/0010761500003116}
\showDOI{\tempurl}


\bibitem[\protect\citeauthoryear{Dandl, Molnar, Binder, and Bischl}{Dandl
  et~al\mbox{.}}{2020}]%
        {dandl_multi-objective_2020}
\bibfield{author}{\bibinfo{person}{Susanne Dandl}, \bibinfo{person}{Christoph
  Molnar}, \bibinfo{person}{Martin Binder}, {and} \bibinfo{person}{Bernd
  Bischl}.} \bibinfo{year}{2020}\natexlab{}.
\newblock \showarticletitle{Multi-Objective Counterfactual Explanations}. In
  \bibinfo{booktitle}{\emph{Parallel Problem Solving from Nature -- PPSN XVI}}.
  \bibinfo{publisher}{Springer International Publishing},
  \bibinfo{address}{Cham}, \bibinfo{pages}{448--469}.
\newblock


\bibitem[\protect\citeauthoryear{DARPA}{DARPA}{[n. d.]}]%
        {DARPA1}
\bibfield{author}{\bibinfo{person}{DARPA}.} \bibinfo{year}{[n. d.]}\natexlab{}.
\newblock \bibinfo{title}{Broad Agency Announcement: Explainable Artificial
  Intelligence (XAI)}.
\newblock
  \bibinfo{howpublished}{\url{https://www.darpa.mil/attachments/DARPA-BAA-16-53.pdf}}.
\newblock
\newblock
\shownote{Accessed: 2020-10-15.}


\bibitem[\protect\citeauthoryear{Dash, Balasubramanian, and Sharma}{Dash
  et~al\mbox{.}}{2022}]%
        {Dash-evaluate-bias-images}
\bibfield{author}{\bibinfo{person}{Saloni Dash}, \bibinfo{person}{Vineeth~N
  Balasubramanian}, {and} \bibinfo{person}{Amit Sharma}.}
  \bibinfo{year}{2022}\natexlab{}.
\newblock \showarticletitle{Evaluating and Mitigating Bias in Image
  Classifiers: A Causal Perspective Using Counterfactuals}. In
  \bibinfo{booktitle}{\emph{Proceedings of the IEEE/CVF Winter Conference on
  Applications of Computer Vision (WACV)}}. \bibinfo{pages}{915--924}.
\newblock


\bibitem[\protect\citeauthoryear{{Datta}, {Sen}, and {Zick}}{{Datta}
  et~al\mbox{.}}{2016}]%
        {QII_SA1}
\bibfield{author}{\bibinfo{person}{A. {Datta}}, \bibinfo{person}{S. {Sen}},
  {and} \bibinfo{person}{Y. {Zick}}.} \bibinfo{year}{2016}\natexlab{}.
\newblock \showarticletitle{Algorithmic Transparency via Quantitative Input
  Influence: Theory and Experiments with Learning Systems}. In
  \bibinfo{booktitle}{\emph{2016 IEEE Symposium on Security and Privacy (SP)}}.
  \bibinfo{publisher}{IEEE}, \bibinfo{address}{New York, USA},
  \bibinfo{pages}{598--617}.
\newblock


\bibitem[\protect\citeauthoryear{de~Lara, González-Sanz, Asher, and
  Loubes}{de~Lara et~al\mbox{.}}{2021}]%
        {causality-from-transport-cfe}
\bibfield{author}{\bibinfo{person}{Lucas de Lara}, \bibinfo{person}{Alberto
  González-Sanz}, \bibinfo{person}{Nicholas Asher}, {and}
  \bibinfo{person}{Jean-Michel Loubes}.} \bibinfo{year}{2021}\natexlab{}.
\newblock \bibinfo{title}{Transport-based Counterfactual Models}.
\newblock
\newblock
\urldef\tempurl%
\url{https://doi.org/10.48550/ARXIV.2108.13025}
\showDOI{\tempurl}


\bibitem[\protect\citeauthoryear{De~Toni, Lepri, and Passerini}{De~Toni
  et~al\mbox{.}}{2022}]%
        {RL-for-CFE-mcts-paper}
\bibfield{author}{\bibinfo{person}{Giovanni De~Toni}, \bibinfo{person}{Bruno
  Lepri}, {and} \bibinfo{person}{Andrea Passerini}.}
  \bibinfo{year}{2022}\natexlab{}.
\newblock \bibinfo{title}{Synthesizing explainable counterfactual policies for
  algorithmic recourse with program synthesis}.
\newblock
\newblock
\urldef\tempurl%
\url{https://doi.org/10.48550/ARXIV.2201.07135}
\showDOI{\tempurl}


\bibitem[\protect\citeauthoryear{Dean, Rich, and Recht}{Dean
  et~al\mbox{.}}{2020}]%
        {sarah_rec}
\bibfield{author}{\bibinfo{person}{Sarah Dean}, \bibinfo{person}{Sarah Rich},
  {and} \bibinfo{person}{Benjamin Recht}.} \bibinfo{year}{2020}\natexlab{}.
\newblock \showarticletitle{Recommendations and User Agency: The Reachability
  of Collaboratively-Filtered Information}. In
  \bibinfo{booktitle}{\emph{Proceedings of the 2020 Conference on Fairness,
  Accountability, and Transparency}} \emph{(\bibinfo{series}{FAT* '20})}.
  \bibinfo{publisher}{Association for Computing Machinery},
  \bibinfo{address}{New York, NY, USA}, 10.
\newblock
\urldef\tempurl%
\url{https://doi.org/10.1145/3351095.3372866}
\showDOI{\tempurl}


\bibitem[\protect\citeauthoryear{Delaney, Greene, and Keane}{Delaney
  et~al\mbox{.}}{2021a}]%
        {cf-timeseries1}
\bibfield{author}{\bibinfo{person}{Eoin Delaney}, \bibinfo{person}{Derek
  Greene}, {and} \bibinfo{person}{Mark~T Keane}.}
  \bibinfo{year}{2021}\natexlab{a}.
\newblock \showarticletitle{Instance-based counterfactual explanations for time
  series classification}. In \bibinfo{booktitle}{\emph{International Conference
  on Case-Based Reasoning}}. Springer, \bibinfo{pages}{32--47}.
\newblock


\bibitem[\protect\citeauthoryear{Delaney, Greene, and Keane}{Delaney
  et~al\mbox{.}}{2021b}]%
        {Delaney2021Uncertainty-CFE}
\bibfield{author}{\bibinfo{person}{Eoin Delaney}, \bibinfo{person}{Derek
  Greene}, {and} \bibinfo{person}{Mark~T. Keane}.}
  \bibinfo{year}{2021}\natexlab{b}.
\newblock \bibinfo{title}{Uncertainty Estimation and Out-of-Distribution
  Detection for Counterfactual Explanations: Pitfalls and Solutions}.
\newblock
\newblock


\bibitem[\protect\citeauthoryear{Deng}{Deng}{2014}]%
        {Deng_exp5}
\bibfield{author}{\bibinfo{person}{Houtao Deng}.}
  \bibinfo{year}{2014}\natexlab{}.
\newblock \showarticletitle{Interpreting Tree Ensembles with inTrees}.
\newblock \bibinfo{journal}{\emph{arXiv:1408.5456}} (\bibinfo{date}{08}
  \bibinfo{year}{2014}).
\newblock
\urldef\tempurl%
\url{https://doi.org/10.1007/s41060-018-0144-8}
\showDOI{\tempurl}


\bibitem[\protect\citeauthoryear{Deng, Dong, Socher, Li, Li, and Fei-Fei}{Deng
  et~al\mbox{.}}{2009}]%
        {imagenet-data}
\bibfield{author}{\bibinfo{person}{Jia Deng}, \bibinfo{person}{Wei Dong},
  \bibinfo{person}{Richard Socher}, \bibinfo{person}{Li-Jia Li},
  \bibinfo{person}{Kai Li}, {and} \bibinfo{person}{Li Fei-Fei}.}
  \bibinfo{year}{2009}\natexlab{}.
\newblock \showarticletitle{ImageNet: A large-scale hierarchical image
  database}. In \bibinfo{booktitle}{\emph{2009 IEEE Conference on Computer
  Vision and Pattern Recognition}}. \bibinfo{pages}{248--255}.
\newblock


\bibitem[\protect\citeauthoryear{Dhurandhar, Chen, Luss, Tu, Ting, Shanmugam,
  and Das}{Dhurandhar et~al\mbox{.}}{2018}]%
        {dhurandhar_explanations_2018}
\bibfield{author}{\bibinfo{person}{Amit Dhurandhar}, \bibinfo{person}{Pin-Yu
  Chen}, \bibinfo{person}{Ronny Luss}, \bibinfo{person}{Chun-Chen Tu},
  \bibinfo{person}{Paishun Ting}, \bibinfo{person}{Karthikeyan Shanmugam},
  {and} \bibinfo{person}{Payel Das}.} \bibinfo{year}{2018}\natexlab{}.
\newblock \showarticletitle{Explanations Based on the Missing: Towards
  Contrastive Explanations with Pertinent Negatives}. In
  \bibinfo{booktitle}{\emph{Proceedings of the 32nd International Conference on
  Neural Information Processing Systems}} \emph{(\bibinfo{series}{NIPS'18})}.
  \bibinfo{publisher}{Curran Associates Inc.}, \bibinfo{address}{Red Hook, NY,
  USA}, \bibinfo{pages}{590–601}.
\newblock


\bibitem[\protect\citeauthoryear{Dhurandhar, Pedapati, Balakrishnan, Chen,
  Shanmugam, and Puri}{Dhurandhar et~al\mbox{.}}{2019}]%
        {dhurandhar_model_2019}
\bibfield{author}{\bibinfo{person}{Amit Dhurandhar}, \bibinfo{person}{Tejaswini
  Pedapati}, \bibinfo{person}{Avinash Balakrishnan}, \bibinfo{person}{Pin-Yu
  Chen}, \bibinfo{person}{Karthikeyan Shanmugam}, {and} \bibinfo{person}{Ruchir
  Puri}.} \bibinfo{year}{2019}\natexlab{}.
\newblock \bibinfo{title}{Model {Agnostic} {Contrastive} {Explanations} for
  {Structured} {Data}}.
\newblock
\newblock
\urldef\tempurl%
\url{http://arxiv.org/abs/1906.00117}
\showURL{%
\tempurl}


\bibitem[\protect\citeauthoryear{Dijkstra}{Dijkstra}{1959}]%
        {dijkstra1959}
\bibfield{author}{\bibinfo{person}{Edsger~W Dijkstra}.}
  \bibinfo{year}{1959}\natexlab{}.
\newblock \showarticletitle{A note on two problems in connexion with graphs}.
\newblock \bibinfo{journal}{\emph{Numerische mathematik}} \bibinfo{volume}{1},
  \bibinfo{number}{1} (\bibinfo{year}{1959}), \bibinfo{pages}{269--271}.
\newblock


\bibitem[\protect\citeauthoryear{Dodge, Liao, Zhang, Bellamy, and Dugan}{Dodge
  et~al\mbox{.}}{2019}]%
        {Dodge-explaining:2019}
\bibfield{author}{\bibinfo{person}{Jonathan Dodge}, \bibinfo{person}{Q.~Vera
  Liao}, \bibinfo{person}{Yunfeng Zhang}, \bibinfo{person}{Rachel K.~E.
  Bellamy}, {and} \bibinfo{person}{Casey Dugan}.}
  \bibinfo{year}{2019}\natexlab{}.
\newblock \showarticletitle{Explaining Models: An Empirical Study of How
  Explanations Impact Fairness Judgment}. In
  \bibinfo{booktitle}{\emph{Proceedings of the 24th International Conference on
  Intelligent User Interfaces}} \emph{(\bibinfo{series}{IUI '19})}.
  \bibinfo{publisher}{Association for Computing Machinery},
  \bibinfo{address}{New York, NY, USA}, 11.
\newblock
\showISBNx{9781450362726}
\urldef\tempurl%
\url{https://doi.org/10.1145/3301275.3302310}
\showDOI{\tempurl}


\bibitem[\protect\citeauthoryear{Doersch}{Doersch}{2016}]%
        {doersch-autoencoder}
\bibfield{author}{\bibinfo{person}{Carl Doersch}.}
  \bibinfo{year}{2016}\natexlab{}.
\newblock \bibinfo{title}{Tutorial on Variational Autoencoders}.
\newblock
\newblock
\showeprint[arxiv]{stat.ML/1606.05908}


\bibitem[\protect\citeauthoryear{Domingos}{Domingos}{1998}]%
        {Pedro_exp4}
\bibfield{author}{\bibinfo{person}{Pedro Domingos}.}
  \bibinfo{year}{1998}\natexlab{}.
\newblock \showarticletitle{Knowledge Discovery Via Multiple Models}.
\newblock \bibinfo{journal}{\emph{Intell. Data Anal.}} \bibinfo{volume}{2},
  \bibinfo{number}{3} (\bibinfo{date}{May} \bibinfo{year}{1998}),
  \bibinfo{pages}{187–202}.
\newblock
\showISSN{1088-467X}


\bibitem[\protect\citeauthoryear{Dominguez-Olmedo, Karimi, and
  Sch{\"o}lkopf}{Dominguez-Olmedo et~al\mbox{.}}{2022}]%
        {dominiguez-olmedo-cfe-robustness}
\bibfield{author}{\bibinfo{person}{Ricardo Dominguez-Olmedo},
  \bibinfo{person}{Amir~H Karimi}, {and} \bibinfo{person}{Bernhard
  Sch{\"o}lkopf}.} \bibinfo{year}{2022}\natexlab{}.
\newblock \showarticletitle{On the Adversarial Robustness of Causal Algorithmic
  Recourse}. In \bibinfo{booktitle}{\emph{Proceedings of the 39th International
  Conference on Machine Learning}} \emph{(\bibinfo{series}{Proceedings of
  Machine Learning Research})}. \bibinfo{publisher}{PMLR},
  \bibinfo{pages}{5324--5342}.
\newblock
\urldef\tempurl%
\url{https://proceedings.mlr.press/v162/dominguez-olmedo22a.html}
\showURL{%
\tempurl}


\bibitem[\protect\citeauthoryear{Doshi-Velez, Kortz, Budish, Bavitz, Gershman,
  O'Brien, Schieber, Waldo, Weinberger, and Wood}{Doshi-Velez
  et~al\mbox{.}}{2017}]%
        {AI-Accountability:2017}
\bibfield{author}{\bibinfo{person}{Finale Doshi-Velez}, \bibinfo{person}{Mason
  Kortz}, \bibinfo{person}{Ryan Budish}, \bibinfo{person}{Chris Bavitz},
  \bibinfo{person}{Sam Gershman}, \bibinfo{person}{D. O'Brien},
  \bibinfo{person}{Stuart Schieber}, \bibinfo{person}{J. Waldo},
  \bibinfo{person}{D. Weinberger}, {and} \bibinfo{person}{Alexandra Wood}.}
  \bibinfo{year}{2017}\natexlab{}.
\newblock \bibinfo{title}{Accountability of AI Under the Law: The Role of
  Explanation}.
\newblock
\newblock


\bibitem[\protect\citeauthoryear{Downs, Chu, Yacoby, Doshi-Velez, and
  Pan}{Downs et~al\mbox{.}}{2020}]%
        {cruds_cf}
\bibfield{author}{\bibinfo{person}{Michael Downs}, \bibinfo{person}{Jonathan
  Chu}, \bibinfo{person}{Yaniv Yacoby}, \bibinfo{person}{Finale Doshi-Velez},
  {and} \bibinfo{person}{Weiwei. Pan}.} \bibinfo{year}{2020}\natexlab{}.
\newblock \showarticletitle{CRUDS: Counterfactual Recourse Using Disentangled
  Subspaces}. In \bibinfo{booktitle}{\emph{Workshop on Human Interpretability
  in Machine Learning (WHI)}}.
\newblock
\urldef\tempurl%
\url{https://finale.seas.harvard.edu/files/finale/files/cruds-_counterfactual_recourse_using_disentangled_subspaces.pdf}
\showURL{%
\tempurl}


\bibitem[\protect\citeauthoryear{Dua and Graff}{Dua and Graff}{2017}]%
        {UCI-repo}
\bibfield{author}{\bibinfo{person}{Dheeru Dua} {and} \bibinfo{person}{Casey
  Graff}.} \bibinfo{year}{2017}\natexlab{}.
\newblock \bibinfo{title}{{UCI} Machine Learning Repository - Adult Income}.
\newblock
\newblock
\urldef\tempurl%
\url{http://archive.ics.uci.edu/ml/datasets/Adult}
\showURL{%
\tempurl}


\bibitem[\protect\citeauthoryear{Dunkelau and Leuschel}{Dunkelau and
  Leuschel}{2019}]%
        {dunkelau_fairness-aware}
\bibfield{author}{\bibinfo{person}{Jannik Dunkelau} {and}
  \bibinfo{person}{Michael Leuschel}.} \bibinfo{year}{2019}\natexlab{}.
\newblock \bibinfo{title}{Fairness-{Aware} {Machine} {Learning}}.
\newblock , \bibinfo{numpages}{60}~pages.
\newblock
\urldef\tempurl%
\url{https://www.phil-fak.uni-duesseldorf.de/fileadmin/Redaktion/Institute/Sozialwissenschaften/Kommunikations-_und_Medienwissenschaft/KMW_I/Working_Paper/Dunkelau___Leuschel__2019__Fairness-Aware_Machine_Learning.pdf}
\showURL{%
\tempurl}


\bibitem[\protect\citeauthoryear{Duong, Li, and Xu}{Duong
  et~al\mbox{.}}{2021}]%
        {prototype_based_cfe}
\bibfield{author}{\bibinfo{person}{Tri~Dung Duong}, \bibinfo{person}{Qian Li},
  {and} \bibinfo{person}{Guandong Xu}.} \bibinfo{year}{2021}\natexlab{}.
\newblock \bibinfo{title}{Prototype-based Counterfactual Explanation for Causal
  Classification}.
\newblock
\newblock
\urldef\tempurl%
\url{https://doi.org/10.48550/ARXIV.2105.00703}
\showDOI{\tempurl}


\bibitem[\protect\citeauthoryear{Dutta, Long, Mishra, Tilli, and
  Magazzeni}{Dutta et~al\mbox{.}}{2022}]%
        {robustness-metric-paper-jpmorgan}
\bibfield{author}{\bibinfo{person}{Sanghamitra Dutta}, \bibinfo{person}{Jason
  Long}, \bibinfo{person}{Saumitra Mishra}, \bibinfo{person}{Cecilia Tilli},
  {and} \bibinfo{person}{Daniele Magazzeni}.} \bibinfo{year}{2022}\natexlab{}.
\newblock \showarticletitle{Robust Counterfactual Explanations for Tree-Based
  Ensembles}. In \bibinfo{booktitle}{\emph{Proceedings of the 39th
  International Conference on Machine Learning}}
  \emph{(\bibinfo{series}{Proceedings of Machine Learning Research})},
  Vol.~\bibinfo{volume}{162}. \bibinfo{publisher}{PMLR},
  \bibinfo{pages}{5742--5756}.
\newblock
\urldef\tempurl%
\url{https://proceedings.mlr.press/v162/dutta22a.html}
\showURL{%
\tempurl}


\bibitem[\protect\citeauthoryear{Elliott, Law, and Russell}{Elliott
  et~al\mbox{.}}{2021}]%
        {Elliott2021_adversarial_cfe_images}
\bibfield{author}{\bibinfo{person}{Andrew Elliott}, \bibinfo{person}{Stephen
  Law}, {and} \bibinfo{person}{Chris Russell}.}
  \bibinfo{year}{2021}\natexlab{}.
\newblock \showarticletitle{Explaining Classifiers using Adversarial
  Perturbations on the Perceptual Ball}. In
  \bibinfo{booktitle}{\emph{Conference on Computer Vision and Pattern
  Recognition (CVPR)}}.
\newblock
\urldef\tempurl%
\url{https://doi.org/10.48550/ARXIV.1912.09405}
\showDOI{\tempurl}


\bibitem[\protect\citeauthoryear{Faber, Moghaddam, and Wattenhofer}{Faber
  et~al\mbox{.}}{2020}]%
        {contrastive-graph-GNN3}
\bibfield{author}{\bibinfo{person}{Lukas Faber}, \bibinfo{person}{Amin~K.
  Moghaddam}, {and} \bibinfo{person}{Roger Wattenhofer}.}
  \bibinfo{year}{2020}\natexlab{}.
\newblock \bibinfo{title}{Contrastive Graph Neural Network Explanation}.
\newblock
\newblock
\urldef\tempurl%
\url{https://doi.org/10.48550/ARXIV.2010.13663}
\showDOI{\tempurl}


\bibitem[\protect\citeauthoryear{Faggella}{Faggella}{2020}]%
        {medical-treatment-ml}
\bibfield{author}{\bibinfo{person}{Daniel Faggella}.}
  \bibinfo{year}{2020}\natexlab{}.
\newblock \bibinfo{title}{Machine Learning for Medical Diagnostics – 4
  Current Applications}.
\newblock
  \bibinfo{howpublished}{\url{https://emerj.com/ai-sector-overviews/machine-learning-medical-diagnostics-4-current-applications/}}.
\newblock
\newblock
\shownote{Accessed: 2020-10-15.}


\bibitem[\protect\citeauthoryear{Fawkes, Evans, and Sejdinovic}{Fawkes
  et~al\mbox{.}}{2022}]%
        {cfe-model-bias2}
\bibfield{author}{\bibinfo{person}{Jake Fawkes}, \bibinfo{person}{Robin Evans},
  {and} \bibinfo{person}{Dino Sejdinovic}.} \bibinfo{year}{2022}\natexlab{}.
\newblock \bibinfo{title}{Selection, Ignorability and Challenges With Causal
  Fairness}.
\newblock
\newblock
\urldef\tempurl%
\url{https://doi.org/10.48550/ARXIV.2202.13774}
\showDOI{\tempurl}


\bibitem[\protect\citeauthoryear{Fdez-Sánchez, Pascual-Triana, Fernández, and
  Herrera}{Fdez-Sánchez et~al\mbox{.}}{2021}]%
        {cfe-biology-multiclass}
\bibfield{author}{\bibinfo{person}{J.A. Fdez-Sánchez}, \bibinfo{person}{J.D.
  Pascual-Triana}, \bibinfo{person}{A. Fernández}, {and} \bibinfo{person}{F.
  Herrera}.} \bibinfo{year}{2021}\natexlab{}.
\newblock \showarticletitle{Learning interpretable multi-class models by means
  of hierarchical decomposition: Threshold Control for Nested Dichotomies}.
\newblock \bibinfo{journal}{\emph{Neurocomputing}}  \bibinfo{volume}{463}
  (\bibinfo{year}{2021}), \bibinfo{pages}{514--524}.
\newblock
\urldef\tempurl%
\url{https://doi.org/10.1016/j.neucom.2021.07.097}
\showDOI{\tempurl}


\bibitem[\protect\citeauthoryear{Feghahati, Shelton, Pazzani, and
  Tang}{Feghahati et~al\mbox{.}}{2020}]%
        {cdeepex_images}
\bibfield{author}{\bibinfo{person}{Amir~H. Feghahati},
  \bibinfo{person}{Christian~R. Shelton}, \bibinfo{person}{Michael~J. Pazzani},
  {and} \bibinfo{person}{Kevin Tang}.} \bibinfo{year}{2020}\natexlab{}.
\newblock \showarticletitle{CDeepEx: Contrastive Deep Explanations}. In
  \bibinfo{booktitle}{\emph{ECAI}}.
\newblock


\bibitem[\protect\citeauthoryear{Fernández, Martín~de Diego, Aceña,
  Fernández-Isabel, and Moguerza}{Fernández et~al\mbox{.}}{2020}]%
        {fernandez-random:2020}
\bibfield{author}{\bibinfo{person}{Rubén~R. Fernández},
  \bibinfo{person}{Isaac Martín~de Diego}, \bibinfo{person}{Víctor Aceña},
  \bibinfo{person}{Alberto Fernández-Isabel}, {and} \bibinfo{person}{Javier~M.
  Moguerza}.} \bibinfo{year}{2020}\natexlab{}.
\newblock \showarticletitle{Random forest explainability using counterfactual
  sets}.
\newblock \bibinfo{journal}{\emph{Information Fusion}}  \bibinfo{volume}{63}
  (\bibinfo{year}{2020}), \bibinfo{pages}{196--207}.
\newblock
\urldef\tempurl%
\url{https://doi.org/10.1016/j.inffus.2020.07.001}
\showDOI{\tempurl}


\bibitem[\protect\citeauthoryear{Fernández-Loría, Provost, and
  Han}{Fernández-Loría et~al\mbox{.}}{2020}]%
        {fernandez-loria_explaining_2020}
\bibfield{author}{\bibinfo{person}{Carlos Fernández-Loría},
  \bibinfo{person}{Foster Provost}, {and} \bibinfo{person}{Xintian Han}.}
  \bibinfo{year}{2020}\natexlab{}.
\newblock \bibinfo{title}{Explaining Data-Driven Decisions made by AI Systems:
  The Counterfactual Approach}.
\newblock
\newblock
\urldef\tempurl%
\url{http://arxiv.org/abs/2001.07417}
\showURL{%
\tempurl}


\bibitem[\protect\citeauthoryear{Ferrario and Loi}{Ferrario and Loi}{2020}]%
        {another-robustness-cfe-ferrario}
\bibfield{author}{\bibinfo{person}{Andrea Ferrario} {and}
  \bibinfo{person}{Michele Loi}.} \bibinfo{year}{2020}\natexlab{}.
\newblock \bibinfo{title}{A Series of Unfortunate Counterfactual Events: the
  Role of Time in Counterfactual Explanations}.
\newblock
\newblock
\urldef\tempurl%
\url{https://doi.org/10.48550/ARXIV.2010.04687}
\showDOI{\tempurl}


\bibitem[\protect\citeauthoryear{FICO}{FICO}{2018}]%
        {fico-data}
\bibfield{author}{\bibinfo{person}{FICO}.} \bibinfo{year}{2018}\natexlab{}.
\newblock \bibinfo{title}{FICO (HELOC) dataset}.
\newblock
\newblock
\urldef\tempurl%
\url{https://community.fico.com/s/explainable-machine-learning-challenge?tabset-3158a=2}
\showURL{%
\tempurl}


\bibitem[\protect\citeauthoryear{Filandrianos, Thomas, Dervakos, and
  Stamou}{Filandrianos et~al\mbox{.}}{2022}]%
        {Filandrianos-CFE-Images}
\bibfield{author}{\bibinfo{person}{Giorgos Filandrianos},
  \bibinfo{person}{Konstantinos Thomas}, \bibinfo{person}{Edmund Dervakos},
  {and} \bibinfo{person}{Giorgos Stamou}.} \bibinfo{year}{2022}\natexlab{}.
\newblock \showarticletitle{Conceptual Edits as Counterfactual Explanations}.
  In \bibinfo{booktitle}{\emph{Proceedings of the {AAAI} 2022 Spring Symposium
  on Machine Learning and Knowledge Engineering for Hybrid Intelligence
  {(AAAI-MAKE} 2022), Stanford University, Palo Alto, California, USA, March
  21-23, 2022}} \emph{(\bibinfo{series}{{CEUR} Workshop Proceedings})},
  Vol.~\bibinfo{volume}{3121}. \bibinfo{publisher}{CEUR-WS.org}.
\newblock
\urldef\tempurl%
\url{http://ceur-ws.org/Vol-3121/paper6.pdf}
\showURL{%
\tempurl}


\bibitem[\protect\citeauthoryear{F{\"o}rster, Hühn, Klier, and
  Kluge}{F{\"o}rster et~al\mbox{.}}{2021}]%
        {forster-capturing-2021}
\bibfield{author}{\bibinfo{person}{Maximilian F{\"o}rster},
  \bibinfo{person}{Philipp Hühn}, \bibinfo{person}{Mathias Klier}, {and}
  \bibinfo{person}{Kilian Kluge}.} \bibinfo{year}{2021}\natexlab{}.
\newblock \showarticletitle{Capturing Users' Reality: A Novel Approach to
  Generate Coherent Counterfactual Explanations}.
\newblock
\urldef\tempurl%
\url{https://doi.org/10.24251/HICSS.2021.155}
\showDOI{\tempurl}


\bibitem[\protect\citeauthoryear{F{\"o}rster, Klier, Kluge, and
  Sigler}{F{\"o}rster et~al\mbox{.}}{2020}]%
        {forster2020evaluating-xai-user-study}
\bibfield{author}{\bibinfo{person}{Maximilian F{\"o}rster},
  \bibinfo{person}{Mathias Klier}, \bibinfo{person}{Kilian Kluge}, {and}
  \bibinfo{person}{Irina Sigler}.} \bibinfo{year}{2020}\natexlab{}.
\newblock \showarticletitle{Evaluating explainable Artifical intelligence--What
  users really appreciate}.
\newblock  (\bibinfo{year}{2020}).
\newblock
\urldef\tempurl%
\url{https://aisel.aisnet.org/ecis2020_rp/195}
\showURL{%
\tempurl}


\bibitem[\protect\citeauthoryear{{Fraunhofer IOSB}, Becker, Burkart, Birnstill,
  and Beyerer}{{Fraunhofer IOSB} et~al\mbox{.}}{2021}]%
        {Fraunhofer-pure-region-sampling}
\bibfield{author}{\bibinfo{person}{{Fraunhofer IOSB}},
  \bibinfo{person}{Maximilian Becker}, \bibinfo{person}{Nadia Burkart},
  \bibinfo{person}{Pascal Birnstill}, {and} \bibinfo{person}{J{\"{u}}rgen
  Beyerer}.} \bibinfo{year}{2021}\natexlab{}.
\newblock \showarticletitle{A Step Towards Global Counterfactual Explanations:
  Approximating the Feature Space Through Hierarchical Division and Graph
  Search}.
\newblock \bibinfo{journal}{\emph{Advances in Artificial Intelligence and
  Machine Learning}} (\bibinfo{year}{2021}), \bibinfo{pages}{90--110}.
\newblock
\urldef\tempurl%
\url{https://doi.org/10.54364/aaiml.2021.1107}
\showDOI{\tempurl}


\bibitem[\protect\citeauthoryear{Freiesleben}{Freiesleben}{2022}]%
        {freiesleben2020CFE_AE1}
\bibfield{author}{\bibinfo{person}{Timo Freiesleben}.}
  \bibinfo{year}{2022}\natexlab{}.
\newblock \showarticletitle{The intriguing relation between counterfactual
  explanations and adversarial examples}.
\newblock \bibinfo{journal}{\emph{Minds Mach. (Dordr.)}} \bibinfo{volume}{32},
  \bibinfo{number}{1} (\bibinfo{year}{2022}), \bibinfo{pages}{77--109}.
\newblock


\bibitem[\protect\citeauthoryear{Friedman}{Friedman}{2001}]%
        {intro_PDP}
\bibfield{author}{\bibinfo{person}{Jerome~H. Friedman}.}
  \bibinfo{year}{2001}\natexlab{}.
\newblock \showarticletitle{Greedy Function Approximation: A Gradient Boosting
  Machine}.
\newblock \bibinfo{journal}{\emph{The Annals of Statistics}}
  \bibinfo{volume}{29}, \bibinfo{number}{5} (\bibinfo{year}{2001}),
  \bibinfo{pages}{1189--1232}.
\newblock
\showISSN{00905364}
\urldef\tempurl%
\url{http://www.jstor.org/stable/2699986}
\showURL{%
\tempurl}


\bibitem[\protect\citeauthoryear{Frohberg and Binder}{Frohberg and
  Binder}{2022}]%
        {CRASS-dataset-NLP}
\bibfield{author}{\bibinfo{person}{Jörg Frohberg} {and} \bibinfo{person}{Frank
  Binder}.} \bibinfo{year}{2022}\natexlab{}.
\newblock \showarticletitle{CRASS: A Novel Data Set and Benchmark to Test
  Counterfactual Reasoning of Large Language Models}. In
  \bibinfo{booktitle}{\emph{Proceedings of the Language Resources and
  Evaluation Conference}}. \bibinfo{publisher}{European Language Resources
  Association}, \bibinfo{address}{Marseille, France},
  \bibinfo{pages}{2126--2140}.
\newblock
\urldef\tempurl%
\url{https://aclanthology.org/2022.lrec-1.229}
\showURL{%
\tempurl}


\bibitem[\protect\citeauthoryear{Fujiwara, Wei, Zhao, and Ma}{Fujiwara
  et~al\mbox{.}}{2022}]%
        {dimension_reduction_cfe}
\bibfield{author}{\bibinfo{person}{Takanori Fujiwara}, \bibinfo{person}{Xinhai
  Wei}, \bibinfo{person}{Jian Zhao}, {and} \bibinfo{person}{Kwan-Liu Ma}.}
  \bibinfo{year}{2022}\natexlab{}.
\newblock \showarticletitle{Interactive Dimensionality Reduction for
  Comparative Analysis}.
\newblock \bibinfo{journal}{\emph{{IEEE} Transactions on Visualization and
  Computer Graphics}} (\bibinfo{year}{2022}), \bibinfo{pages}{758--768}.
\newblock
\urldef\tempurl%
\url{https://doi.org/10.1109/tvcg.2021.3114807}
\showDOI{\tempurl}


\bibitem[\protect\citeauthoryear{Förster, Hühn, Klier, and Kluge}{Förster
  et~al\mbox{.}}{2021}]%
        {coherent_CFEs}
\bibfield{author}{\bibinfo{person}{Maximilian Förster},
  \bibinfo{person}{Philipp Hühn}, \bibinfo{person}{Mathias Klier}, {and}
  \bibinfo{person}{Kilian Kluge}.} \bibinfo{year}{2021}\natexlab{}.
\newblock \showarticletitle{Capturing Users' Reality: A Novel Approach to
  Generate Coherent Counterfactual Explanations}.
\newblock
\urldef\tempurl%
\url{https://doi.org/10.24251/HICSS.2021.155}
\showDOI{\tempurl}


\bibitem[\protect\citeauthoryear{Galhotra, Pradhan, and Salimi}{Galhotra
  et~al\mbox{.}}{2021}]%
        {galhotra-sigmod-cfe}
\bibfield{author}{\bibinfo{person}{Sainyam Galhotra}, \bibinfo{person}{Romila
  Pradhan}, {and} \bibinfo{person}{Babak Salimi}.}
  \bibinfo{year}{2021}\natexlab{}.
\newblock \showarticletitle{Explaining Black-Box Algorithms Using Probabilistic
  Contrastive Counterfactuals}. In \bibinfo{booktitle}{\emph{SIGMOD '21:
  International Conference on Management of Data, Virtual Event, China, June
  20-25, 2021}}. \bibinfo{publisher}{{ACM}}.
\newblock
\urldef\tempurl%
\url{https://doi.org/10.1145/3448016.3458455}
\showDOI{\tempurl}


\bibitem[\protect\citeauthoryear{Gan, Zhang, Zhang, and Li}{Gan
  et~al\mbox{.}}{2021}]%
        {cfe-financial-risk-debugging}
\bibfield{author}{\bibinfo{person}{Jingwei Gan}, \bibinfo{person}{Shinan
  Zhang}, \bibinfo{person}{Chi Zhang}, {and} \bibinfo{person}{Andy Li}.}
  \bibinfo{year}{2021}\natexlab{}.
\newblock \showarticletitle{Automated Counterfactual Generation in Financial
  Model Risk Management}. In \bibinfo{booktitle}{\emph{2021 IEEE International
  Conference on Big Data (Big Data)}}. \bibinfo{pages}{4064--4068}.
\newblock
\urldef\tempurl%
\url{https://doi.org/10.1109/BigData52589.2021.9671561}
\showDOI{\tempurl}


\bibitem[\protect\citeauthoryear{Garc{\'i}a-Laencina, Sancho-G{\'o}mez, and
  Figueiras-Vidal}{Garc{\'i}a-Laencina et~al\mbox{.}}{2009}]%
        {missing-data}
\bibfield{author}{\bibinfo{person}{P.~J. Garc{\'i}a-Laencina},
  \bibinfo{person}{J. Sancho-G{\'o}mez}, {and} \bibinfo{person}{A.~R.
  Figueiras-Vidal}.} \bibinfo{year}{2009}\natexlab{}.
\newblock \showarticletitle{Pattern classification with missing data: a
  review}.
\newblock \bibinfo{journal}{\emph{Neural Computing and Applications}}
  \bibinfo{volume}{19} (\bibinfo{year}{2009}), \bibinfo{pages}{263--282}.
\newblock


\bibitem[\protect\citeauthoryear{Garisch}{Garisch}{[n. d.]}]%
        {bank-models}
\bibfield{author}{\bibinfo{person}{Gordon Garisch}.} \bibinfo{year}{[n.
  d.]}\natexlab{}.
\newblock \bibinfo{title}{MODEL LIFECYCLE TRANSFORMATION: HOW BANKS ARE
  UNLOCKING EFFICIENCIES}.
\newblock
  \bibinfo{howpublished}{\url{https://financialservicesblog.accenture.com/model-lifecycle-transformation-how-banks-are-unlocking-efficiencies}}.
\newblock
\newblock
\shownote{Accessed: 2022-10-15.}


\bibitem[\protect\citeauthoryear{Ge, Liu, Li, Xu, Geng, Li, Tan, Sun, and
  Zhang}{Ge et~al\mbox{.}}{2021}]%
        {use-cfe-instead-of-erasure}
\bibfield{author}{\bibinfo{person}{Yingqiang Ge}, \bibinfo{person}{Shuchang
  Liu}, \bibinfo{person}{Zelong Li}, \bibinfo{person}{Shuyuan Xu},
  \bibinfo{person}{Shijie Geng}, \bibinfo{person}{Yunqi Li},
  \bibinfo{person}{Juntao Tan}, \bibinfo{person}{Fei Sun}, {and}
  \bibinfo{person}{Yongfeng Zhang}.} \bibinfo{year}{2021}\natexlab{}.
\newblock \bibinfo{title}{Counterfactual Evaluation for Explainable AI}.
\newblock
\newblock
\urldef\tempurl%
\url{https://doi.org/10.48550/ARXIV.2109.01962}
\showDOI{\tempurl}


\bibitem[\protect\citeauthoryear{Ghandeharioun, Kim, Li, Jou, Eoff, and
  Picard}{Ghandeharioun et~al\mbox{.}}{2022}]%
        {ghandeharioun-DISSECT-images}
\bibfield{author}{\bibinfo{person}{Asma Ghandeharioun}, \bibinfo{person}{Been
  Kim}, \bibinfo{person}{Chun-Liang Li}, \bibinfo{person}{Brendan Jou},
  \bibinfo{person}{Brian Eoff}, {and} \bibinfo{person}{Rosalind Picard}.}
  \bibinfo{year}{2022}\natexlab{}.
\newblock \showarticletitle{{DISSECT}: Disentangled Simultaneous Explanations
  via Concept Traversals}. In \bibinfo{booktitle}{\emph{International
  Conference on Learning Representations}}.
\newblock
\urldef\tempurl%
\url{https://openreview.net/forum?id=qY79G8jGsep}
\showURL{%
\tempurl}


\bibitem[\protect\citeauthoryear{Ghazimatin, Balalau, Saha~Roy, and
  Weikum}{Ghazimatin et~al\mbox{.}}{2020}]%
        {cfe-reco-approach1}
\bibfield{author}{\bibinfo{person}{Azin Ghazimatin}, \bibinfo{person}{Oana
  Balalau}, \bibinfo{person}{Rishiraj Saha~Roy}, {and} \bibinfo{person}{Gerhard
  Weikum}.} \bibinfo{year}{2020}\natexlab{}.
\newblock \showarticletitle{PRINCE: Provider-Side Interpretability with
  Counterfactual Explanations in Recommender Systems}. In
  \bibinfo{booktitle}{\emph{Proceedings of the 13th International Conference on
  Web Search and Data Mining}} \emph{(\bibinfo{series}{WSDM '20})}.
  \bibinfo{publisher}{Association for Computing Machinery},
  \bibinfo{address}{New York, NY, USA}, 9.
\newblock
\urldef\tempurl%
\url{https://doi.org/10.1145/3336191.3371824}
\showDOI{\tempurl}


\bibitem[\protect\citeauthoryear{Giannopoulos, Papastefanatos, Sacharidis, and
  Stefanidis}{Giannopoulos et~al\mbox{.}}{2021}]%
        {cfe-reco-approach7}
\bibfield{author}{\bibinfo{person}{Giorgos Giannopoulos},
  \bibinfo{person}{George Papastefanatos}, \bibinfo{person}{Dimitris
  Sacharidis}, {and} \bibinfo{person}{Kostas Stefanidis}.}
  \bibinfo{year}{2021}\natexlab{}.
\newblock \bibinfo{booktitle}{\emph{Interactivity, Fairness and Explanations in
  Recommendations}}.
\newblock \bibinfo{publisher}{Association for Computing Machinery},
  \bibinfo{address}{New York, NY, USA}.
\newblock
\urldef\tempurl%
\url{https://doi.org/10.1145/3450614.3462238}
\showURL{%
\tempurl}


\bibitem[\protect\citeauthoryear{Goldstein, Kapelner, Bleich, and
  Pitkin}{Goldstein et~al\mbox{.}}{2013}]%
        {ICE_PDP1}
\bibfield{author}{\bibinfo{person}{Alex Goldstein}, \bibinfo{person}{Adam
  Kapelner}, \bibinfo{person}{Justin Bleich}, {and} \bibinfo{person}{Emil
  Pitkin}.} \bibinfo{year}{2013}\natexlab{}.
\newblock \showarticletitle{Peeking Inside the Black Box: Visualizing
  Statistical Learning With Plots of Individual Conditional Expectation}.
\newblock \bibinfo{journal}{\emph{Journal of Computational and Graphical
  Statistics}}  \bibinfo{volume}{24} (\bibinfo{date}{09} \bibinfo{year}{2013}).
\newblock
\urldef\tempurl%
\url{https://doi.org/10.1080/10618600.2014.907095}
\showDOI{\tempurl}


\bibitem[\protect\citeauthoryear{Gomez, Holter, Yuan, and Bertini}{Gomez
  et~al\mbox{.}}{2020}]%
        {vice-visualcfe}
\bibfield{author}{\bibinfo{person}{Oscar Gomez}, \bibinfo{person}{Steffen
  Holter}, \bibinfo{person}{Jun Yuan}, {and} \bibinfo{person}{Enrico Bertini}.}
  \bibinfo{year}{2020}\natexlab{}.
\newblock \showarticletitle{ViCE: Visual Counterfactual Explanations for
  Machine Learning Models}. In \bibinfo{booktitle}{\emph{Proceedings of the
  25th International Conference on Intelligent User Interfaces}}
  \emph{(\bibinfo{series}{IUI '20})}. 5.
\newblock
\urldef\tempurl%
\url{https://doi.org/10.1145/3377325.3377536}
\showURL{%
\tempurl}


\bibitem[\protect\citeauthoryear{Gomez, Holter, Yuan, and Bertini}{Gomez
  et~al\mbox{.}}{2021}]%
        {advice-visualcfe}
\bibfield{author}{\bibinfo{person}{Oscar Gomez}, \bibinfo{person}{Steffen
  Holter}, \bibinfo{person}{Jun Yuan}, {and} \bibinfo{person}{Enrico Bertini}.}
  \bibinfo{year}{2021}\natexlab{}.
\newblock \bibinfo{title}{AdViCE: Aggregated Visual Counterfactual Explanations
  for Machine Learning Model Validation}.
\newblock
\newblock
\urldef\tempurl%
\url{https://doi.org/10.48550/ARXIV.2109.05629}
\showDOI{\tempurl}


\bibitem[\protect\citeauthoryear{Goodman and Flaxman}{Goodman and
  Flaxman}{2016}]%
        {Euro-GDPR2}
\bibfield{author}{\bibinfo{person}{Bryce Goodman} {and} \bibinfo{person}{S.
  Flaxman}.} \bibinfo{year}{2016}\natexlab{}.
\newblock \showarticletitle{EU regulations on algorithmic decision-making and a
  "right to explanation"}.
\newblock \bibinfo{journal}{\emph{ArXiv}}  \bibinfo{volume}{abs/1606.08813}
  (\bibinfo{year}{2016}).
\newblock


\bibitem[\protect\citeauthoryear{Goyal, Wu, Ernst, Batra, Parikh, and
  Lee}{Goyal et~al\mbox{.}}{2019}]%
        {cfe-visual-explanation-goyal2019}
\bibfield{author}{\bibinfo{person}{Yash Goyal}, \bibinfo{person}{Ziyan Wu},
  \bibinfo{person}{Jan Ernst}, \bibinfo{person}{Dhruv Batra},
  \bibinfo{person}{Devi Parikh}, {and} \bibinfo{person}{Stefan Lee}.}
  \bibinfo{year}{2019}\natexlab{}.
\newblock \showarticletitle{Counterfactual Visual Explanations}. In
  \bibinfo{booktitle}{\emph{Proceedings of the 36th International Conference on
  Machine Learning}} \emph{(\bibinfo{series}{Proceedings of Machine Learning
  Research})}, Vol.~\bibinfo{volume}{97}. \bibinfo{publisher}{PMLR},
  \bibinfo{pages}{2376--2384}.
\newblock
\urldef\tempurl%
\url{https://proceedings.mlr.press/v97/goyal19a.html}
\showURL{%
\tempurl}


\bibitem[\protect\citeauthoryear{Gralla}{Gralla}{2016}]%
        {prime-racist}
\bibfield{author}{\bibinfo{person}{Preston Gralla}.}
  \bibinfo{year}{2016}\natexlab{}.
\newblock \bibinfo{title}{Amazon Prime and the racist algorithms}.
\newblock
\newblock
\urldef\tempurl%
\url{https://www.computerworld.com/article/3068622/amazon-prime-and-the-racist-algorithms.html}
\showURL{%
\tempurl}


\bibitem[\protect\citeauthoryear{Grath, Costabello, Van, Sweeney, Kamiab, Shen,
  and Lecue}{Grath et~al\mbox{.}}{2018}]%
        {grath_interpretable_2018}
\bibfield{author}{\bibinfo{person}{Rory~Mc Grath}, \bibinfo{person}{Luca
  Costabello}, \bibinfo{person}{Chan~Le Van}, \bibinfo{person}{Paul Sweeney},
  \bibinfo{person}{Farbod Kamiab}, \bibinfo{person}{Zhao Shen}, {and}
  \bibinfo{person}{Freddy Lecue}.} \bibinfo{year}{2018}\natexlab{}.
\newblock \bibinfo{title}{Interpretable {Credit} {Application} {Predictions}
  {With} {Counterfactual} {Explanations}}.
\newblock
\newblock
\urldef\tempurl%
\url{http://arxiv.org/abs/1811.05245}
\showURL{%
\tempurl}


\bibitem[\protect\citeauthoryear{Group}{Group}{2018}]%
        {home-credit-data}
\bibfield{author}{\bibinfo{person}{Home~Credit Group}.}
  \bibinfo{year}{2018}\natexlab{}.
\newblock \bibinfo{title}{Home Credit Default Risk.}
\newblock
\newblock
\urldef\tempurl%
\url{https://www.kaggle.com/c/home-credit-default-risk/data}
\showURL{%
\tempurl}


\bibitem[\protect\citeauthoryear{Guidotti, Monreale, Ruggieri, Pedreschi,
  Turini, and Giannotti}{Guidotti et~al\mbox{.}}{2018a}]%
        {guidotti_local_2018}
\bibfield{author}{\bibinfo{person}{Riccardo Guidotti}, \bibinfo{person}{Anna
  Monreale}, \bibinfo{person}{Salvatore Ruggieri}, \bibinfo{person}{Dino
  Pedreschi}, \bibinfo{person}{Franco Turini}, {and} \bibinfo{person}{Fosca
  Giannotti}.} \bibinfo{year}{2018}\natexlab{a}.
\newblock \bibinfo{title}{Local {Rule}-{Based} {Explanations} of {Black} {Box}
  {Decision} {Systems}}.
\newblock
\newblock
\urldef\tempurl%
\url{http://arxiv.org/abs/1805.10820}
\showURL{%
\tempurl}


\bibitem[\protect\citeauthoryear{Guidotti, Monreale, Ruggieri, Turini,
  Giannotti, and Pedreschi}{Guidotti et~al\mbox{.}}{2018b}]%
        {xai-survey4}
\bibfield{author}{\bibinfo{person}{Riccardo Guidotti}, \bibinfo{person}{Anna
  Monreale}, \bibinfo{person}{Salvatore Ruggieri}, \bibinfo{person}{Franco
  Turini}, \bibinfo{person}{Fosca Giannotti}, {and} \bibinfo{person}{Dino
  Pedreschi}.} \bibinfo{year}{2018}\natexlab{b}.
\newblock \showarticletitle{A Survey of Methods for Explaining Black Box
  Models}.
\newblock \bibinfo{journal}{\emph{ACM Comput. Surv.}} \bibinfo{volume}{51},
  \bibinfo{number}{5}, Article \bibinfo{articleno}{93} (\bibinfo{date}{Aug.}
  \bibinfo{year}{2018}), \bibinfo{numpages}{42}~pages.
\newblock
\urldef\tempurl%
\url{https://doi.org/10.1145/3236009}
\showDOI{\tempurl}


\bibitem[\protect\citeauthoryear{Guidotti and Ruggieri}{Guidotti and
  Ruggieri}{2021}]%
        {guidotti-cfe-ensemble-of-explainers}
\bibfield{author}{\bibinfo{person}{Riccardo Guidotti} {and}
  \bibinfo{person}{Salvatore Ruggieri}.} \bibinfo{year}{2021}\natexlab{}.
\newblock \showarticletitle{Ensemble of Counterfactual Explainers}.
  \bibinfo{publisher}{Springer-Verlag}, \bibinfo{address}{Berlin, Heidelberg},
  11.
\newblock
\urldef\tempurl%
\url{https://doi.org/10.1007/978-3-030-88942-5_28}
\showDOI{\tempurl}


\bibitem[\protect\citeauthoryear{Gulshad and Smeulders}{Gulshad and
  Smeulders}{2021}]%
        {attribute_based_images}
\bibfield{author}{\bibinfo{person}{Sadaf Gulshad} {and} \bibinfo{person}{Arnold
  Smeulders}.} \bibinfo{year}{2021}\natexlab{}.
\newblock \showarticletitle{Counterfactual attribute-based visual explanations
  for classification}.
\newblock \bibinfo{journal}{\emph{International Journal of Multimedia
  Information Retrieval}} (\bibinfo{year}{2021}), \bibinfo{pages}{127--140}.
\newblock
\urldef\tempurl%
\url{https://doi.org/10.1007/s13735-021-00208-3}
\showDOI{\tempurl}


\bibitem[\protect\citeauthoryear{Guo, Nguyen, and Yadav}{Guo
  et~al\mbox{.}}{2021}]%
        {guo-cfe-counternet}
\bibfield{author}{\bibinfo{person}{Hangzhi Guo}, \bibinfo{person}{Thanh~Hong
  Nguyen}, {and} \bibinfo{person}{Amulya Yadav}.}
  \bibinfo{year}{2021}\natexlab{}.
\newblock \bibinfo{title}{CounterNet: End-to-End Training of Counterfactual
  Aware Predictions}.
\newblock
\newblock
\urldef\tempurl%
\url{https://doi.org/10.48550/ARXIV.2109.07557}
\showDOI{\tempurl}


\bibitem[\protect\citeauthoryear{Gupta, Genc, and O'Sullivan}{Gupta
  et~al\mbox{.}}{2022}]%
        {gupta-cfe-constraint-satisfaction-problem}
\bibfield{author}{\bibinfo{person}{Sharmi~Dev Gupta}, \bibinfo{person}{Begum
  Genc}, {and} \bibinfo{person}{Barry O'Sullivan}.}
  \bibinfo{year}{2022}\natexlab{}.
\newblock \bibinfo{title}{Finding Counterfactual Explanations through
  Constraint Relaxations}.
\newblock
\newblock
\urldef\tempurl%
\url{https://doi.org/10.48550/ARXIV.2204.03429}
\showDOI{\tempurl}


\bibitem[\protect\citeauthoryear{Gupta, Nokhiz, Roy, and
  Venkatasubramanian}{Gupta et~al\mbox{.}}{2019}]%
        {Gupta2019EqualRecourse}
\bibfield{author}{\bibinfo{person}{Vivek Gupta}, \bibinfo{person}{Pegah
  Nokhiz}, \bibinfo{person}{Chitradeep~Dutta Roy}, {and}
  \bibinfo{person}{Suresh Venkatasubramanian}.}
  \bibinfo{year}{2019}\natexlab{}.
\newblock \bibinfo{title}{Equalizing Recourse across Groups}.
\newblock
\newblock


\bibitem[\protect\citeauthoryear{Guyomard, Fessant, Bouadi, and Guyet}{Guyomard
  et~al\mbox{.}}{2021}]%
        {encoder_cfe_image}
\bibfield{author}{\bibinfo{person}{Victor Guyomard},
  \bibinfo{person}{Françoise Fessant}, \bibinfo{person}{Tassadit Bouadi},
  {and} \bibinfo{person}{Thomas Guyet}.} \bibinfo{year}{2021}\natexlab{}.
\newblock \showarticletitle{Post-hoc counterfactual generation with supervised
  autoencoder}.
\newblock


\bibitem[\protect\citeauthoryear{Hada and Carreira-Perpi{\~{n}}{\'a}n}{Hada and
  Carreira-Perpi{\~{n}}{\'a}n}{2021}]%
        {oblique-tree-cfe-copy}
\bibfield{author}{\bibinfo{person}{Suryabhan~Singh Hada} {and}
  \bibinfo{person}{Miguel~{\'A}. Carreira-Perpi{\~{n}}{\'a}n}.}
  \bibinfo{year}{2021}\natexlab{}.
\newblock \showarticletitle{Exploring Counterfactual Explanations for
  Classification and Regression Trees}. In \bibinfo{booktitle}{\emph{Machine
  Learning and Principles and Practice of Knowledge Discovery in Databases}}.
  \bibinfo{publisher}{Springer International Publishing},
  \bibinfo{address}{Cham}, \bibinfo{pages}{489--504}.
\newblock
\showISBNx{978-3-030-93736-2}


\bibitem[\protect\citeauthoryear{Haldar, John, and Saha}{Haldar
  et~al\mbox{.}}{2021}]%
        {cfe_anomaly_detection}
\bibfield{author}{\bibinfo{person}{Swastik Haldar},
  \bibinfo{person}{Philips~George John}, {and} \bibinfo{person}{Diptikalyan
  Saha}.} \bibinfo{year}{2021}\natexlab{}.
\newblock \showarticletitle{Reliable Counterfactual Explanations for
  Autoencoder Based Anomalies}. In \bibinfo{booktitle}{\emph{8th ACM IKDD CODS
  and 26th COMAD}}. \bibinfo{publisher}{Association for Computing Machinery},
  \bibinfo{address}{New York, NY, USA}, \bibinfo{pages}{83–91}.
\newblock
\urldef\tempurl%
\url{https://doi.org/10.1145/3430984.3431015}
\showDOI{\tempurl}


\bibitem[\protect\citeauthoryear{Han and Ghosh}{Han and Ghosh}{2021}]%
        {debugging-data-cfe3-han}
\bibfield{author}{\bibinfo{person}{Xing Han} {and} \bibinfo{person}{Joydeep
  Ghosh}.} \bibinfo{year}{2021}\natexlab{}.
\newblock \showarticletitle{Model-Agnostic Explanations using Minimal Forcing
  Subsets}. In \bibinfo{booktitle}{\emph{2021 International Joint Conference on
  Neural Networks (IJCNN)}}. \bibinfo{pages}{1--8}.
\newblock
\urldef\tempurl%
\url{https://doi.org/10.1109/IJCNN52387.2021.9533992}
\showDOI{\tempurl}


\bibitem[\protect\citeauthoryear{Hashemi and Fathi}{Hashemi and Fathi}{2020}]%
        {CFE_genetic_creditscorecards}
\bibfield{author}{\bibinfo{person}{Masoud Hashemi} {and} \bibinfo{person}{Ali
  Fathi}.} \bibinfo{year}{2020}\natexlab{}.
\newblock \bibinfo{title}{PermuteAttack: Counterfactual Explanation of Machine
  Learning Credit Scorecards}.
\newblock
\newblock
\urldef\tempurl%
\url{https://doi.org/10.48550/ARXIV.2008.10138}
\showDOI{\tempurl}


\bibitem[\protect\citeauthoryear{Hendricks, Hu, Darrell, and Akata}{Hendricks
  et~al\mbox{.}}{2018}]%
        {Hendricks-image-CFE}
\bibfield{author}{\bibinfo{person}{Lisa~Anne Hendricks},
  \bibinfo{person}{Ronghang Hu}, \bibinfo{person}{Trevor Darrell}, {and}
  \bibinfo{person}{Zeynep Akata}.} \bibinfo{year}{2018}\natexlab{}.
\newblock \bibinfo{title}{Generating Counterfactual Explanations with Natural
  Language}.
\newblock
\newblock
\urldef\tempurl%
\url{https://doi.org/10.48550/ARXIV.1806.09809}
\showDOI{\tempurl}


\bibitem[\protect\citeauthoryear{Henelius, Puolam\"{a}ki, Bostr\"{o}m, Asker,
  and Papapetrou}{Henelius et~al\mbox{.}}{2014}]%
        {Andreas_exp7}
\bibfield{author}{\bibinfo{person}{Andreas Henelius}, \bibinfo{person}{Kai
  Puolam\"{a}ki}, \bibinfo{person}{Henrik Bostr\"{o}m}, \bibinfo{person}{Lars
  Asker}, {and} \bibinfo{person}{Panagiotis Papapetrou}.}
  \bibinfo{year}{2014}\natexlab{}.
\newblock \showarticletitle{A Peek into the Black Box: Exploring Classifiers by
  Randomization}.
\newblock \bibinfo{journal}{\emph{Data Min. Knowl. Discov.}}
  \bibinfo{volume}{28}, \bibinfo{number}{5-6} (\bibinfo{year}{2014}), 27.
\newblock
\urldef\tempurl%
\url{https://doi.org/10.1007/s10618-014-0368-8}
\showDOI{\tempurl}


\bibitem[\protect\citeauthoryear{Hinder and Hammer}{Hinder and Hammer}{2020}]%
        {hinder-cfe-drift}
\bibfield{author}{\bibinfo{person}{Fabian Hinder} {and}
  \bibinfo{person}{Barbara Hammer}.} \bibinfo{year}{2020}\natexlab{}.
\newblock \bibinfo{title}{Counterfactual Explanations of Concept Drift}.
\newblock
\newblock
\urldef\tempurl%
\url{https://doi.org/10.48550/ARXIV.2006.12822}
\showDOI{\tempurl}


\bibitem[\protect\citeauthoryear{Hohman, Head, Caruana, DeLine, and
  Drucker}{Hohman et~al\mbox{.}}{2019}]%
        {GAMUT}
\bibfield{author}{\bibinfo{person}{Fred Hohman}, \bibinfo{person}{Andrew Head},
  \bibinfo{person}{Rich Caruana}, \bibinfo{person}{Robert DeLine}, {and}
  \bibinfo{person}{Steven~Mark Drucker}.} \bibinfo{year}{2019}\natexlab{}.
\newblock \showarticletitle{Gamut: A Design Probe to Understand How Data
  Scientists Understand Machine Learning Models}.
\newblock \bibinfo{journal}{\emph{Proceedings of the 2019 CHI Conference on
  Human Factors in Computing Systems}} (\bibinfo{year}{2019}).
\newblock


\bibitem[\protect\citeauthoryear{Hong, Haimovich, and Taylor}{Hong
  et~al\mbox{.}}{2018}]%
        {hospitaltriage-data}
\bibfield{author}{\bibinfo{person}{Woo~Suk Hong},
  \bibinfo{person}{Adrian~Daniel Haimovich}, {and} \bibinfo{person}{R.~Andrew
  Taylor}.} \bibinfo{year}{2018}\natexlab{}.
\newblock \showarticletitle{Predicting hospital admission at emergency
  department triage using machine learning}.
\newblock \bibinfo{journal}{\emph{PLOS ONE}} \bibinfo{volume}{13},
  \bibinfo{number}{7} (\bibinfo{year}{2018}).
\newblock
\urldef\tempurl%
\url{https://doi.org/10.1371/journal.pone.0201016}
\showDOI{\tempurl}


\bibitem[\protect\citeauthoryear{House}{House}{2022}]%
        {blueprints-bill-of-rights}
\bibfield{author}{\bibinfo{person}{The US~White House}.}
  \bibinfo{year}{2022}\natexlab{}.
\newblock \bibinfo{title}{Blueprint for an AI bill of rights}.
\newblock
\newblock
\urldef\tempurl%
\url{https://www.whitehouse.gov/ostp/ai-bill-of-rights/#discrimination}
\showURL{%
\tempurl}


\bibitem[\protect\citeauthoryear{Hsieh, Moreira, and Ouyang}{Hsieh
  et~al\mbox{.}}{2021}]%
        {Dice-time-series-ext}
\bibfield{author}{\bibinfo{person}{Chihcheng Hsieh}, \bibinfo{person}{Catarina
  Moreira}, {and} \bibinfo{person}{Chun Ouyang}.}
  \bibinfo{year}{2021}\natexlab{}.
\newblock \showarticletitle{DiCE4EL: Interpreting Process Predictions using a
  Milestone-Aware Counterfactual Approach}. In \bibinfo{booktitle}{\emph{2021
  3rd International Conference on Process Mining (ICPM)}}.
  \bibinfo{pages}{88--95}.
\newblock
\urldef\tempurl%
\url{https://doi.org/10.1109/ICPM53251.2021.9576881}
\showDOI{\tempurl}


\bibitem[\protect\citeauthoryear{Huang, Metzger, and Pohl}{Huang
  et~al\mbox{.}}{2022}]%
        {Loreley-huang-2022}
\bibfield{author}{\bibinfo{person}{Tsung-Hao Huang}, \bibinfo{person}{Andreas
  Metzger}, {and} \bibinfo{person}{Klaus Pohl}.}
  \bibinfo{year}{2022}\natexlab{}.
\newblock \showarticletitle{Counterfactual Explanations for Predictive Business
  Process Monitoring}. \bibinfo{publisher}{Springer International Publishing},
  \bibinfo{address}{Cham}, \bibinfo{pages}{399--413}.
\newblock


\bibitem[\protect\citeauthoryear{Hvilshøj, Iosifidis, and Assent}{Hvilshøj
  et~al\mbox{.}}{2021a}]%
        {Hvilshoj-ECINN-images}
\bibfield{author}{\bibinfo{person}{Frederik Hvilshøj},
  \bibinfo{person}{Alexandros Iosifidis}, {and} \bibinfo{person}{Ira Assent}.}
  \bibinfo{year}{2021}\natexlab{a}.
\newblock \bibinfo{title}{ECINN: Efficient Counterfactuals from Invertible
  Neural Networks}.
\newblock
\newblock
\urldef\tempurl%
\url{https://doi.org/10.48550/ARXIV.2103.13701}
\showDOI{\tempurl}


\bibitem[\protect\citeauthoryear{Hvilshøj, Iosifidis, and Assent}{Hvilshøj
  et~al\mbox{.}}{2021b}]%
        {hvilshoj-cfe-new-metrics}
\bibfield{author}{\bibinfo{person}{Frederik Hvilshøj},
  \bibinfo{person}{Alexandros Iosifidis}, {and} \bibinfo{person}{Ira Assent}.}
  \bibinfo{year}{2021}\natexlab{b}.
\newblock \bibinfo{title}{On Quantitative Evaluations of Counterfactuals}.
\newblock
\newblock
\urldef\tempurl%
\url{https://doi.org/10.48550/ARXIV.2111.00177}
\showDOI{\tempurl}


\bibitem[\protect\citeauthoryear{Höltgen, Schut, Brauner, and Gal}{Höltgen
  et~al\mbox{.}}{2021}]%
        {holtgen-DeDUCE-images}
\bibfield{author}{\bibinfo{person}{Benedikt Höltgen}, \bibinfo{person}{Lisa
  Schut}, \bibinfo{person}{Jan~M. Brauner}, {and} \bibinfo{person}{Yarin Gal}.}
  \bibinfo{year}{2021}\natexlab{}.
\newblock \bibinfo{title}{DeDUCE: Generating Counterfactual Explanations
  Efficiently}.
\newblock
\newblock
\urldef\tempurl%
\url{https://doi.org/10.48550/ARXIV.2111.15639}
\showDOI{\tempurl}


\bibitem[\protect\citeauthoryear{in~Data Science Conference The Global Open
  Source Severity~of Illness Score~Consortium.}{in~Data Science Conference The
  Global Open Source Severity~of Illness Score~Consortium.}{2020}]%
        {woman-in-cs-data}
\bibfield{author}{\bibinfo{person}{Global~Women in~Data Science Conference The
  Global Open Source Severity~of Illness Score~Consortium.}}
  \bibinfo{year}{2020}\natexlab{}.
\newblock \bibinfo{title}{WiDS Datathon 2020}.
\newblock
\newblock
\urldef\tempurl%
\url{https://www.kaggle.com/c/widsdatathon2020}
\showURL{%
\tempurl}


\bibitem[\protect\citeauthoryear{Insurance}{Insurance}{2011}]%
        {allstate-data}
\bibfield{author}{\bibinfo{person}{Allstate Insurance}.}
  \bibinfo{year}{2011}\natexlab{}.
\newblock \bibinfo{title}{Allstate Claim Prediction Challenge}.
\newblock
\newblock
\urldef\tempurl%
\url{https://www.kaggle.com/c/ClaimPredictionChallenge}
\showURL{%
\tempurl}


\bibitem[\protect\citeauthoryear{intelligence artificielle}{intelligence
  artificielle}{[n. d.]}]%
        {Fr-XAI2}
\bibfield{author}{\bibinfo{person}{France intelligence artificielle}.}
  \bibinfo{year}{[n. d.]}\natexlab{}.
\newblock \bibinfo{title}{RAPPORT DE SYNTHESE FRANCE INTELLIGENCE
  ARTIFICIELLE}.
\newblock
  \bibinfo{howpublished}{\url{https://www.economie.gouv.fr/files/files/PDF/2017/Rapport_synthese_France_IA_.pdf}}.
\newblock
\newblock
\shownote{Accessed: 2020-10-15.}


\bibitem[\protect\citeauthoryear{Irvin, Rajpurkar, Ko, Yu, Ciurea-Ilcus, Chute,
  Marklund, Haghgoo, Ball, Shpanskaya, Seekins, Mong, Halabi, Sandberg, Jones,
  Larson, Langlotz, Patel, Lungren, and Ng}{Irvin et~al\mbox{.}}{2019}]%
        {chexpert-data}
\bibfield{author}{\bibinfo{person}{Jeremy Irvin}, \bibinfo{person}{Pranav
  Rajpurkar}, \bibinfo{person}{Michael Ko}, \bibinfo{person}{Yifan Yu},
  \bibinfo{person}{Silviana Ciurea-Ilcus}, \bibinfo{person}{Chris Chute},
  \bibinfo{person}{Henrik Marklund}, \bibinfo{person}{Behzad Haghgoo},
  \bibinfo{person}{Robyn Ball}, \bibinfo{person}{Katie Shpanskaya},
  \bibinfo{person}{Jayne Seekins}, \bibinfo{person}{David~A. Mong},
  \bibinfo{person}{Safwan~S. Halabi}, \bibinfo{person}{Jesse~K. Sandberg},
  \bibinfo{person}{Ricky Jones}, \bibinfo{person}{David~B. Larson},
  \bibinfo{person}{Curtis~P. Langlotz}, \bibinfo{person}{Bhavik~N. Patel},
  \bibinfo{person}{Matthew~P. Lungren}, {and} \bibinfo{person}{Andrew~Y. Ng}.}
  \bibinfo{year}{2019}\natexlab{}.
\newblock \bibinfo{title}{CheXpert: A Large Chest Radiograph Dataset with
  Uncertainty Labels and Expert Comparison}.
\newblock
\newblock
\urldef\tempurl%
\url{https://doi.org/10.48550/ARXIV.1901.07031}
\showDOI{\tempurl}


\bibitem[\protect\citeauthoryear{Jacob, Zablocki, Ben-Younes, Chen, Pérez, and
  Cord}{Jacob et~al\mbox{.}}{[n. d.]}]%
        {cfe-images-STEEX}
\bibfield{author}{\bibinfo{person}{Paul Jacob}, \bibinfo{person}{Éloi
  Zablocki}, \bibinfo{person}{Hédi Ben-Younes}, \bibinfo{person}{Mickaël
  Chen}, \bibinfo{person}{Patrick Pérez}, {and} \bibinfo{person}{Matthieu
  Cord}.} \bibinfo{year}{[n. d.]}\natexlab{}.
\newblock \bibinfo{title}{STEEX: Steering Counterfactual Explanations with
  Semantics}.
\newblock
\newblock
\urldef\tempurl%
\url{https://doi.org/10.48550/ARXIV.2111.09094}
\showDOI{\tempurl}


\bibitem[\protect\citeauthoryear{Jeanneret, Simon, and Jurie}{Jeanneret
  et~al\mbox{.}}{2022}]%
        {jeanneret-diffusion-cfe-images}
\bibfield{author}{\bibinfo{person}{Guillaume Jeanneret}, \bibinfo{person}{Loïc
  Simon}, {and} \bibinfo{person}{Frédéric Jurie}.}
  \bibinfo{year}{2022}\natexlab{}.
\newblock \bibinfo{title}{Diffusion Models for Counterfactual Explanations}.
\newblock
\newblock
\urldef\tempurl%
\url{https://doi.org/10.48550/ARXIV.2203.15636}
\showDOI{\tempurl}


\bibitem[\protect\citeauthoryear{Jeff~Larson and Angwin}{Jeff~Larson and
  Angwin}{2016}]%
        {compas-data}
\bibfield{author}{\bibinfo{person}{Lauren~Kirchner Jeff~Larson, Surya~Mattu}
  {and} \bibinfo{person}{Julia Angwin}.} \bibinfo{year}{2016}\natexlab{}.
\newblock \bibinfo{title}{UCI Machine Learning Repository}.
\newblock
\newblock
\urldef\tempurl%
\url{https://github.com/propublica/compas-analysis/}
\showURL{%
\tempurl}


\bibitem[\protect\citeauthoryear{Jia, McDermid, and Habli}{Jia
  et~al\mbox{.}}{2021}]%
        {unifying-cfe2-deeplift}
\bibfield{author}{\bibinfo{person}{Yan Jia}, \bibinfo{person}{John McDermid},
  {and} \bibinfo{person}{Ibrahim Habli}.} \bibinfo{year}{2021}\natexlab{}.
\newblock \showarticletitle{Enhancing the Value of Counterfactual Explanations
  for Deep Learning}. In \bibinfo{booktitle}{\emph{Artificial Intelligence in
  Medicine}}. \bibinfo{publisher}{Springer International Publishing},
  \bibinfo{address}{Cham}, \bibinfo{pages}{389--394}.
\newblock


\bibitem[\protect\citeauthoryear{Johnson, Bulgarelli, Pollard, Horng, Celi, and
  Mark}{Johnson et~al\mbox{.}}{2021}]%
        {MMIC-IV}
\bibfield{author}{\bibinfo{person}{Alistair Johnson}, \bibinfo{person}{Lucas
  Bulgarelli}, \bibinfo{person}{Tom Pollard}, \bibinfo{person}{Steven Horng},
  \bibinfo{person}{Leo~Anthony Celi}, {and} \bibinfo{person}{Roger Mark}.}
  \bibinfo{year}{2021}\natexlab{}.
\newblock \bibinfo{title}{MIMIC-IV}.
\newblock
\newblock
\urldef\tempurl%
\url{https://doi.org/10.13026/S6N6-XD98}
\showDOI{\tempurl}


\bibitem[\protect\citeauthoryear{Jordan and Freiburger}{Jordan and
  Freiburger}{2015}]%
        {bail-data}
\bibfield{author}{\bibinfo{person}{Kareem~L. Jordan} {and}
  \bibinfo{person}{Tina~L. Freiburger}.} \bibinfo{year}{2015}\natexlab{}.
\newblock \showarticletitle{The Effect of Race/Ethnicity on Sentencing:
  Examining Sentence Type, Jail Length, and Prison Length}.
\newblock \bibinfo{journal}{\emph{Journal of Ethnicity in Criminal Justice}}
  \bibinfo{volume}{13}, \bibinfo{number}{3} (\bibinfo{year}{2015}).
\newblock
\urldef\tempurl%
\url{https://doi.org/10.1080/15377938.2014.984045}
\showDOI{\tempurl}


\bibitem[\protect\citeauthoryear{Joshi, Koyejo, Vijitbenjaronk, Kim, and
  Ghosh}{Joshi et~al\mbox{.}}{2019}]%
        {joshi_towards_2019}
\bibfield{author}{\bibinfo{person}{Shalmali Joshi}, \bibinfo{person}{Oluwasanmi
  Koyejo}, \bibinfo{person}{Warut Vijitbenjaronk}, \bibinfo{person}{Been Kim},
  {and} \bibinfo{person}{Joydeep Ghosh}.} \bibinfo{year}{2019}\natexlab{}.
\newblock \bibinfo{title}{Towards {Realistic} {Individual} {Recourse} and
  {Actionable} {Explanations} in {Black}-{Box} {Decision} {Making} {Systems}}.
\newblock
\newblock
\urldef\tempurl%
\url{http://arxiv.org/abs/1907.09615}
\showURL{%
\tempurl}


\bibitem[\protect\citeauthoryear{Jung, Kang, Kim, Won, and Lee}{Jung
  et~al\mbox{.}}{2020}]%
        {Gradual_Construction}
\bibfield{author}{\bibinfo{person}{Hong-Gyu Jung}, \bibinfo{person}{Sin-Han
  Kang}, \bibinfo{person}{Hee-Dong Kim}, \bibinfo{person}{Dong-Ok Won}, {and}
  \bibinfo{person}{Seong-Whan Lee}.} \bibinfo{year}{2020}\natexlab{}.
\newblock \bibinfo{title}{Counterfactual Explanation Based on Gradual
  Construction for Deep Networks}.
\newblock
\newblock
\urldef\tempurl%
\url{https://doi.org/10.48550/ARXIV.2008.01897}
\showDOI{\tempurl}


\bibitem[\protect\citeauthoryear{Kaffes, Sacharidis, and Giannopoulos}{Kaffes
  et~al\mbox{.}}{2021}]%
        {cfe-reco-approach3}
\bibfield{author}{\bibinfo{person}{Vassilis Kaffes}, \bibinfo{person}{Dimitris
  Sacharidis}, {and} \bibinfo{person}{Giorgos Giannopoulos}.}
  \bibinfo{year}{2021}\natexlab{}.
\newblock \showarticletitle{Model-Agnostic Counterfactual Explanations of
  Recommendations}. In \bibinfo{booktitle}{\emph{Proceedings of the 29th ACM
  Conference on User Modeling, Adaptation and Personalization}}
  \emph{(\bibinfo{series}{UMAP '21})}. \bibinfo{publisher}{Association for
  Computing Machinery}, \bibinfo{address}{New York, NY, USA}, 6.
\newblock
\urldef\tempurl%
\url{https://doi.org/10.1145/3450613.3456846}
\showDOI{\tempurl}


\bibitem[\protect\citeauthoryear{Kaggle}{Kaggle}{2012}]%
        {givemesomecredit-data}
\bibfield{author}{\bibinfo{person}{Kaggle}.} \bibinfo{year}{2012}\natexlab{}.
\newblock \bibinfo{title}{Give Me Some Credit}.
\newblock
\newblock
\urldef\tempurl%
\url{https://www.kaggle.com/c/GiveMeSomeCredit}
\showURL{%
\tempurl}


\bibitem[\protect\citeauthoryear{Kahneman and Miller}{Kahneman and
  Miller}{1986}]%
        {Kahneman1986:psycho3}
\bibfield{author}{\bibinfo{person}{D. Kahneman} {and} \bibinfo{person}{D.
  Miller}.} \bibinfo{year}{1986}\natexlab{}.
\newblock \showarticletitle{Norm Theory: Comparing Reality to Its
  Alternatives.}
\newblock \bibinfo{journal}{\emph{Psychological Review}}  \bibinfo{volume}{93}
  (\bibinfo{year}{1986}), \bibinfo{pages}{136--153}.
\newblock


\bibitem[\protect\citeauthoryear{Kanamori, Takagi, Kobayashi, and
  Arimura}{Kanamori et~al\mbox{.}}{2020}]%
        {Kanamori2020:DACE}
\bibfield{author}{\bibinfo{person}{Kentaro Kanamori}, \bibinfo{person}{Takuya
  Takagi}, \bibinfo{person}{Ken Kobayashi}, {and} \bibinfo{person}{Hiroki
  Arimura}.} \bibinfo{year}{2020}\natexlab{}.
\newblock \showarticletitle{DACE: Distribution-Aware Counterfactual Explanation
  by Mixed-Integer Linear Optimization}. In
  \bibinfo{booktitle}{\emph{International Joint Conference on Artificial
  Intelligence (IJCAI)}}. \bibinfo{address}{California, USA}.
\newblock
\urldef\tempurl%
\url{https://doi.org/10.24963/ijcai.2020/395}
\showDOI{\tempurl}


\bibitem[\protect\citeauthoryear{Kanamori, Takagi, Kobayashi, and Ike}{Kanamori
  et~al\mbox{.}}{2022}]%
        {kanamori-decision-tree-globalcfe}
\bibfield{author}{\bibinfo{person}{Kentaro Kanamori}, \bibinfo{person}{Takuya
  Takagi}, \bibinfo{person}{Ken Kobayashi}, {and} \bibinfo{person}{Yuichi
  Ike}.} \bibinfo{year}{2022}\natexlab{}.
\newblock \showarticletitle{Counterfactual Explanation Trees: Transparent and
  Consistent Actionable Recourse with Decision Trees}. In
  \bibinfo{booktitle}{\emph{Proceedings of The 25th International Conference on
  Artificial Intelligence and Statistics}} \emph{(\bibinfo{series}{Proceedings
  of Machine Learning Research})}. \bibinfo{publisher}{PMLR},
  \bibinfo{pages}{1846--1870}.
\newblock


\bibitem[\protect\citeauthoryear{Kanamori, Takagi, Kobayashi, Ike, Uemura, and
  Arimura}{Kanamori et~al\mbox{.}}{2021}]%
        {Ordered-CFE-Kanamori}
\bibfield{author}{\bibinfo{person}{Kentaro Kanamori}, \bibinfo{person}{Takuya
  Takagi}, \bibinfo{person}{Ken Kobayashi}, \bibinfo{person}{Yuichi Ike},
  \bibinfo{person}{Kento Uemura}, {and} \bibinfo{person}{Hiroki Arimura}.}
  \bibinfo{year}{2021}\natexlab{}.
\newblock \showarticletitle{Ordered Counterfactual Explanation by Mixed-Integer
  Linear Optimization}.
\newblock \bibinfo{journal}{\emph{Proceedings of the AAAI Conference on
  Artificial Intelligence}} \bibinfo{volume}{35}, \bibinfo{number}{13}
  (\bibinfo{year}{2021}), 11.
\newblock
\urldef\tempurl%
\url{https://doi.org/10.1609/aaai.v35i13.17376}
\showDOI{\tempurl}


\bibitem[\protect\citeauthoryear{Karimi, Barthe, Balle, and Valera}{Karimi
  et~al\mbox{.}}{2020a}]%
        {karimi_model-agnostic_2020}
\bibfield{author}{\bibinfo{person}{A.-H. Karimi}, \bibinfo{person}{G. Barthe},
  \bibinfo{person}{B. Balle}, {and} \bibinfo{person}{I. Valera}.}
  \bibinfo{year}{2020}\natexlab{a}.
\newblock \bibinfo{title}{Model-Agnostic Counterfactual Explanations for
  Consequential Decisions}.
\newblock
\newblock
\urldef\tempurl%
\url{http://arxiv.org/abs/1905.11190}
\showURL{%
\tempurl}


\bibitem[\protect\citeauthoryear{Karimi, Sch\"{o}lkopf, and Valera}{Karimi
  et~al\mbox{.}}{2021}]%
        {karimi_algorithmic_2020}
\bibfield{author}{\bibinfo{person}{Amir-Hossein Karimi},
  \bibinfo{person}{Bernhard Sch\"{o}lkopf}, {and} \bibinfo{person}{Isabel
  Valera}.} \bibinfo{year}{2021}\natexlab{}.
\newblock \showarticletitle{Algorithmic Recourse: From Counterfactual
  Explanations to Interventions}. In \bibinfo{booktitle}{\emph{Proceedings of
  the 2021 ACM Conference on Fairness, Accountability, and Transparency}}
  \emph{(\bibinfo{series}{FAccT '21})}. \bibinfo{publisher}{Association for
  Computing Machinery}, \bibinfo{address}{New York, NY, USA}, 10.
\newblock
\urldef\tempurl%
\url{https://doi.org/10.1145/3442188.3445899}
\showDOI{\tempurl}


\bibitem[\protect\citeauthoryear{Karimi, von Kügelgen, Schölkopf, and
  Valera}{Karimi et~al\mbox{.}}{2020b}]%
        {karimi-imperfect:2020}
\bibfield{author}{\bibinfo{person}{Amir-Hossein Karimi},
  \bibinfo{person}{Julius von Kügelgen}, \bibinfo{person}{Bernhard
  Schölkopf}, {and} \bibinfo{person}{Isabel Valera}.}
  \bibinfo{year}{2020}\natexlab{b}.
\newblock \bibinfo{title}{Algorithmic recourse under imperfect causal
  knowledge: a probabilistic approach}.
\newblock
\newblock
\urldef\tempurl%
\url{http://arxiv.org/abs/2006.06831}
\showURL{%
\tempurl}


\bibitem[\protect\citeauthoryear{Karlsson, Rebane, Papapetrou, and
  Gionis}{Karlsson et~al\mbox{.}}{2020}]%
        {karlsson-cfe-time-series}
\bibfield{author}{\bibinfo{person}{Isak Karlsson}, \bibinfo{person}{Jonathan
  Rebane}, \bibinfo{person}{Panagiotis Papapetrou}, {and}
  \bibinfo{person}{Aristides Gionis}.} \bibinfo{year}{2020}\natexlab{}.
\newblock \showarticletitle{Locally and Globally Explainable Time Series
  Tweaking}.
\newblock \bibinfo{journal}{\emph{Knowl. Inf. Syst.}} (\bibinfo{year}{2020}),
  30.
\newblock
\urldef\tempurl%
\url{https://doi.org/10.1007/s10115-019-01389-4}
\showDOI{\tempurl}


\bibitem[\protect\citeauthoryear{Kasirzadeh and Smart}{Kasirzadeh and
  Smart}{2021}]%
        {Atoosa-philosophy-cfe}
\bibfield{author}{\bibinfo{person}{Atoosa Kasirzadeh} {and}
  \bibinfo{person}{Andrew Smart}.} \bibinfo{year}{2021}\natexlab{}.
\newblock \showarticletitle{The Use and Misuse of Counterfactuals in Ethical
  Machine Learning}. In \bibinfo{booktitle}{\emph{Proceedings of the 2021 ACM
  Conference on Fairness, Accountability, and Transparency}}.
  \bibinfo{publisher}{Association for Computing Machinery},
  \bibinfo{address}{New York, NY, USA}, 9.
\newblock
\urldef\tempurl%
\url{https://doi.org/10.1145/3442188.3445886}
\showDOI{\tempurl}


\bibitem[\protect\citeauthoryear{Keane, Kenny, Delaney, and Smyth}{Keane
  et~al\mbox{.}}{2021}]%
        {if_only_better_CFE_keane}
\bibfield{author}{\bibinfo{person}{Mark~T. Keane}, \bibinfo{person}{Eoin~M.
  Kenny}, \bibinfo{person}{Eoin Delaney}, {and} \bibinfo{person}{Barry Smyth}.}
  \bibinfo{year}{2021}\natexlab{}.
\newblock \showarticletitle{If Only We Had Better Counterfactual Explanations:
  Five Key Deficits to Rectify in the Evaluation of Counterfactual {XAI}
  Techniques}.
\newblock \bibinfo{journal}{\emph{CoRR}} (\bibinfo{year}{2021}).
\newblock
\urldef\tempurl%
\url{https://arxiv.org/abs/2103.01035}
\showURL{%
\tempurl}


\bibitem[\protect\citeauthoryear{Keane and Smyth}{Keane and Smyth}{2020}]%
        {keane2020good}
\bibfield{author}{\bibinfo{person}{Mark~T. Keane} {and} \bibinfo{person}{Barry
  Smyth}.} \bibinfo{year}{2020}\natexlab{}.
\newblock \bibinfo{title}{Good Counterfactuals and Where to Find Them: A
  Case-Based Technique for Generating Counterfactuals for Explainable AI
  (XAI)}.
\newblock
\newblock
\showeprint[arxiv]{cs.AI/2005.13997}


\bibitem[\protect\citeauthoryear{Kenny and Keane}{Kenny and Keane}{2021}]%
        {Kenny-cfe-images-semifactual}
\bibfield{author}{\bibinfo{person}{Eoin~M. Kenny} {and} \bibinfo{person}{Mark~T
  Keane}.} \bibinfo{year}{2021}\natexlab{}.
\newblock \showarticletitle{On Generating Plausible Counterfactual and
  Semi-Factual Explanations for Deep Learning}.
\newblock \bibinfo{journal}{\emph{Proceedings of the AAAI Conference on
  Artificial Intelligence}}  \bibinfo{volume}{35} (\bibinfo{date}{May}
  \bibinfo{year}{2021}), 11.
\newblock
\urldef\tempurl%
\url{https://ojs.aaai.org/index.php/AAAI/article/view/17377}
\showURL{%
\tempurl}


\bibitem[\protect\citeauthoryear{Khorram and Fuxin}{Khorram and Fuxin}{2022}]%
        {khorram-cycle-cfe-images}
\bibfield{author}{\bibinfo{person}{Saeed Khorram} {and} \bibinfo{person}{Li
  Fuxin}.} \bibinfo{year}{2022}\natexlab{}.
\newblock \showarticletitle{Cycle-Consistent Counterfactuals by Latent
  Transformations}. In \bibinfo{booktitle}{\emph{Proceedings of the IEEE/CVF
  Conference on Computer Vision and Pattern Recognition (CVPR)}}. 10.
\newblock


\bibitem[\protect\citeauthoryear{Kment}{Kment}{2006}]%
        {Kment:phil1}
\bibfield{author}{\bibinfo{person}{Boris Kment}.}
  \bibinfo{year}{2006}\natexlab{}.
\newblock \showarticletitle{Counterfactuals and Explanation}.
\newblock \bibinfo{journal}{\emph{Mind}}  \bibinfo{volume}{115}
  (\bibinfo{date}{04} \bibinfo{year}{2006}).
\newblock
\urldef\tempurl%
\url{https://doi.org/10.1093/mind/fzl261}
\showDOI{\tempurl}


\bibitem[\protect\citeauthoryear{Knight}{Knight}{2019}]%
        {apple-sexist}
\bibfield{author}{\bibinfo{person}{Will Knight}.}
  \bibinfo{year}{2019}\natexlab{}.
\newblock \bibinfo{title}{The Apple Card Didn't 'See' Gender—and That's the
  Problem}.
\newblock
\newblock
\urldef\tempurl%
\url{https://www.wired.com/story/the-apple-card-didnt-see-genderand-thats-the-problem/}
\showURL{%
\tempurl}


\bibitem[\protect\citeauthoryear{Kommiya~Mothilal, Mahajan, Tan, and
  Sharma}{Kommiya~Mothilal et~al\mbox{.}}{2021}]%
        {unifying-cfe1-divyat}
\bibfield{author}{\bibinfo{person}{Ramaravind Kommiya~Mothilal},
  \bibinfo{person}{Divyat Mahajan}, \bibinfo{person}{Chenhao Tan}, {and}
  \bibinfo{person}{Amit Sharma}.} \bibinfo{year}{2021}\natexlab{}.
\newblock \bibinfo{booktitle}{\emph{Towards Unifying Feature Attribution and
  Counterfactual Explanations: Different Means to the Same End}}.
\newblock \bibinfo{publisher}{Association for Computing Machinery},
  \bibinfo{address}{New York, NY, USA}.
\newblock


\bibitem[\protect\citeauthoryear{Koo, Klabjan, and Utke}{Koo
  et~al\mbox{.}}{2020}]%
        {inverse-classification-multiple-algos}
\bibfield{author}{\bibinfo{person}{Jaehoon Koo}, \bibinfo{person}{Diego
  Klabjan}, {and} \bibinfo{person}{Jean Utke}.}
  \bibinfo{year}{2020}\natexlab{}.
\newblock \bibinfo{title}{Inverse Classification with Limited Budget and
  Maximum Number of Perturbed Samples}.
\newblock
\newblock
\urldef\tempurl%
\url{https://doi.org/10.48550/ARXIV.2009.14111}
\showDOI{\tempurl}


\bibitem[\protect\citeauthoryear{Koopman and Renooij}{Koopman and
  Renooij}{2021}]%
        {Koopman2021PersuasiveCE}
\bibfield{author}{\bibinfo{person}{Tara Koopman} {and} \bibinfo{person}{Silja
  Renooij}.} \bibinfo{year}{2021}\natexlab{}.
\newblock \showarticletitle{Persuasive Contrastive Explanations for Bayesian
  Networks}. In \bibinfo{booktitle}{\emph{Symbolic and Quantitative Approaches
  to Reasoning with Uncertainty}}. \bibinfo{publisher}{Springer International
  Publishing}, \bibinfo{address}{Cham}, \bibinfo{pages}{229--242}.
\newblock


\bibitem[\protect\citeauthoryear{Korikov and Beck}{Korikov and Beck}{2021}]%
        {korikov-cfe-inverse-optimization2}
\bibfield{author}{\bibinfo{person}{Anton Korikov} {and}
  \bibinfo{person}{J.~Christopher Beck}.} \bibinfo{year}{2021}\natexlab{}.
\newblock \showarticletitle{{Counterfactual Explanations via Inverse Constraint
  Programming}}. In \bibinfo{booktitle}{\emph{27th International Conference on
  Principles and Practice of Constraint Programming (CP 2021)}}
  \emph{(\bibinfo{series}{Leibniz International Proceedings in Informatics
  (LIPIcs)})}, Vol.~\bibinfo{volume}{210}. \bibinfo{publisher}{Schloss Dagstuhl
  -- Leibniz-Zentrum f{\"u}r Informatik}, \bibinfo{address}{Dagstuhl, Germany},
  \bibinfo{pages}{35:1--35:16}.
\newblock
\showISBNx{978-3-95977-211-2}
\showISSN{1868-8969}
\urldef\tempurl%
\url{https://doi.org/10.4230/LIPIcs.CP.2021.35}
\showDOI{\tempurl}


\bibitem[\protect\citeauthoryear{Korikov, Shleyfman, and Beck}{Korikov
  et~al\mbox{.}}{2021}]%
        {korikov-cfe-inverse-optimization1}
\bibfield{author}{\bibinfo{person}{Anton Korikov}, \bibinfo{person}{Alexander
  Shleyfman}, {and} \bibinfo{person}{J.~Christopher Beck}.}
  \bibinfo{year}{2021}\natexlab{}.
\newblock \showarticletitle{Counterfactual Explanations for Optimization-Based
  Decisions in the Context of the GDPR}. In
  \bibinfo{booktitle}{\emph{Proceedings of the Thirtieth International Joint
  Conference on Artificial Intelligence, {IJCAI-21}}}.
  \bibinfo{pages}{4097--4103}.
\newblock
\urldef\tempurl%
\url{https://doi.org/10.24963/ijcai.2021/564}
\showDOI{\tempurl}


\bibitem[\protect\citeauthoryear{Kovalev, Utkin, Coolen, and
  Konstantinov}{Kovalev et~al\mbox{.}}{2021}]%
        {kovalev-CFE-functional}
\bibfield{author}{\bibinfo{person}{Maxim Kovalev}, \bibinfo{person}{Lev Utkin},
  \bibinfo{person}{Frank Coolen}, {and} \bibinfo{person}{Andrei Konstantinov}.}
  \bibinfo{year}{2021}\natexlab{}.
\newblock \showarticletitle{Counterfactual Explanation of Machine Learning
  Survival Models}.
\newblock \bibinfo{journal}{\emph{Informatica}} \bibinfo{volume}{32},
  \bibinfo{number}{4} (\bibinfo{date}{jan} \bibinfo{year}{2021}),
  \bibinfo{pages}{817–847}.
\newblock
\showISSN{0868-4952}
\urldef\tempurl%
\url{https://doi.org/10.15388/21-INFOR468}
\showDOI{\tempurl}


\bibitem[\protect\citeauthoryear{Krishnan, Sivakumar, and
  Bhattacharya}{Krishnan et~al\mbox{.}}{1999}]%
        {KRISHNAN_exp2}
\bibfield{author}{\bibinfo{person}{R. Krishnan}, \bibinfo{person}{G.
  Sivakumar}, {and} \bibinfo{person}{P. Bhattacharya}.}
  \bibinfo{year}{1999}\natexlab{}.
\newblock \showarticletitle{Extracting decision trees from trained neural
  networks}.
\newblock \bibinfo{journal}{\emph{Pattern Recognition}} \bibinfo{volume}{32},
  \bibinfo{number}{12} (\bibinfo{year}{1999}), \bibinfo{pages}{1999 -- 2009}.
\newblock
\showISSN{0031-3203}
\urldef\tempurl%
\url{https://doi.org/10.1016/S0031-3203(98)00181-2}
\showDOI{\tempurl}


\bibitem[\protect\citeauthoryear{Krishnan and Wu}{Krishnan and Wu}{2017}]%
        {Krishnan_exp8}
\bibfield{author}{\bibinfo{person}{Sanjay Krishnan} {and}
  \bibinfo{person}{Eugene Wu}.} \bibinfo{year}{2017}\natexlab{}.
\newblock \showarticletitle{PALM: Machine Learning Explanations For Iterative
  Debugging}. In \bibinfo{booktitle}{\emph{Proceedings of the 2nd Workshop on
  Human-In-the-Loop Data Analytics}} \emph{(\bibinfo{series}{HILDA'17})}.
  \bibinfo{publisher}{Association for Computing Machinery},
  \bibinfo{address}{New York, NY, USA}, Article \bibinfo{articleno}{4},
  \bibinfo{numpages}{6}~pages.
\newblock
\showISBNx{9781450350297}
\urldef\tempurl%
\url{https://doi.org/10.1145/3077257.3077271}
\showDOI{\tempurl}


\bibitem[\protect\citeauthoryear{Kuhl, Artelt, and Hammer}{Kuhl
  et~al\mbox{.}}{2022}]%
        {keep-your-friends-close-cfe}
\bibfield{author}{\bibinfo{person}{Ulrike Kuhl}, \bibinfo{person}{Andr{\'e}
  Artelt}, {and} \bibinfo{person}{Barbara Hammer}.}
  \bibinfo{year}{2022}\natexlab{}.
\newblock \showarticletitle{Keep Your Friends Close and Your Counterfactuals
  Closer: Improved Learning From Closest Rather Than Plausible Counterfactual
  Explanations in an Abstract Setting}.
\newblock \bibinfo{journal}{\emph{ArXiv}}  \bibinfo{volume}{abs/2205.05515}
  (\bibinfo{year}{2022}).
\newblock


\bibitem[\protect\citeauthoryear{König, Freiesleben, and
  Grosse-Wentrup}{König et~al\mbox{.}}{2021}]%
        {meaningful-recourse-konig}
\bibfield{author}{\bibinfo{person}{Gunnar König}, \bibinfo{person}{Timo
  Freiesleben}, {and} \bibinfo{person}{Moritz Grosse-Wentrup}.}
  \bibinfo{year}{2021}\natexlab{}.
\newblock \bibinfo{title}{A Causal Perspective on Meaningful and Robust
  Algorithmic Recourse}.
\newblock
\newblock
\urldef\tempurl%
\url{https://doi.org/10.48550/ARXIV.2107.07853}
\showDOI{\tempurl}


\bibitem[\protect\citeauthoryear{Labaien, Zugasti, and De~Carlos}{Labaien
  et~al\mbox{.}}{2021}]%
        {da-dgcex_images}
\bibfield{author}{\bibinfo{person}{Jokin Labaien}, \bibinfo{person}{Ekhi
  Zugasti}, {and} \bibinfo{person}{Xabier De~Carlos}.}
  \bibinfo{year}{2021}\natexlab{}.
\newblock \bibinfo{title}{DA-DGCEx: Ensuring Validity of Deep Guided
  Counterfactual Explanations With Distribution-Aware Autoencoder Loss}.
\newblock
\newblock
\urldef\tempurl%
\url{https://doi.org/10.48550/ARXIV.2104.09062}
\showDOI{\tempurl}


\bibitem[\protect\citeauthoryear{Lash, Lin, Street, Robinson, and Ohlmann}{Lash
  et~al\mbox{.}}{2017}]%
        {inverse-classification2}
\bibfield{author}{\bibinfo{person}{Michael~T. Lash}, \bibinfo{person}{Qihang
  Lin}, \bibinfo{person}{William~Nick Street}, \bibinfo{person}{Jennifer~G.
  Robinson}, {and} \bibinfo{person}{Jeffrey~W. Ohlmann}.}
  \bibinfo{year}{2017}\natexlab{}.
\newblock \showarticletitle{Generalized Inverse Classification}. In
  \bibinfo{booktitle}{\emph{SDM}}. \bibinfo{publisher}{Society for Industrial
  and Applied Mathematics}, \bibinfo{address}{Philadelphia, PA, USA},
  \bibinfo{pages}{162--170}.
\newblock


\bibitem[\protect\citeauthoryear{Laugel, Lesot, Marsala, and Detyniecki}{Laugel
  et~al\mbox{.}}{2019a}]%
        {issues_posthoc}
\bibfield{author}{\bibinfo{person}{Thibault Laugel},
  \bibinfo{person}{Marie-Jeanne Lesot}, \bibinfo{person}{Christophe Marsala},
  {and} \bibinfo{person}{Marcin Detyniecki}.} \bibinfo{year}{2019}\natexlab{a}.
\newblock \bibinfo{title}{Issues with post-hoc counterfactual explanations: a
  discussion}.
\newblock
\newblock
\showeprint[arxiv]{1906.04774}


\bibitem[\protect\citeauthoryear{Laugel, Lesot, Marsala, Renard, and
  Detyniecki}{Laugel et~al\mbox{.}}{2018}]%
        {medina_comparison-based_2018}
\bibfield{author}{\bibinfo{person}{Thibault Laugel},
  \bibinfo{person}{Marie-Jeanne Lesot}, \bibinfo{person}{Christophe Marsala},
  \bibinfo{person}{Xavier Renard}, {and} \bibinfo{person}{Marcin Detyniecki}.}
  \bibinfo{year}{2018}\natexlab{}.
\newblock \showarticletitle{Comparison-{Based} {Inverse} {Classification} for
  {Interpretability} in {Machine} {Learning}}.
\newblock In \bibinfo{booktitle}{\emph{Information Processing and Management of
  Uncertainty in Knowledge-Based Systems, Theory and Foundations (IPMU)}}.
  \bibinfo{publisher}{Springer International Publishing}.
\newblock
\urldef\tempurl%
\url{https://doi.org/10.1007/978-3-319-91473-2_9}
\showDOI{\tempurl}


\bibitem[\protect\citeauthoryear{Laugel, Lesot, Marsala, Renard, and
  Detyniecki}{Laugel et~al\mbox{.}}{2019b}]%
        {laugel_dangers_2019}
\bibfield{author}{\bibinfo{person}{Thibault Laugel},
  \bibinfo{person}{Marie-Jeanne Lesot}, \bibinfo{person}{Christophe Marsala},
  \bibinfo{person}{Xavier Renard}, {and} \bibinfo{person}{Marcin Detyniecki}.}
  \bibinfo{year}{2019}\natexlab{b}.
\newblock \bibinfo{title}{The {Dangers} of {Post}-hoc {Interpretability}:
  {Unjustified} {Counterfactual} {Explanations}}.
\newblock
\newblock
\urldef\tempurl%
\url{http://arxiv.org/abs/1907.09294}
\showURL{%
\tempurl}


\bibitem[\protect\citeauthoryear{Le, Wang, and Lee}{Le et~al\mbox{.}}{2019}]%
        {Grace:2019}
\bibfield{author}{\bibinfo{person}{Thai Le}, \bibinfo{person}{Suhang Wang},
  {and} \bibinfo{person}{Dongwon Lee}.} \bibinfo{year}{2019}\natexlab{}.
\newblock \bibinfo{title}{GRACE: Generating Concise and Informative Contrastive
  Sample to Explain Neural Network Model's Prediction}.
\newblock
\newblock
\showeprint[arxiv]{cs.LG/1911.02042}


\bibitem[\protect\citeauthoryear{LeCun and Cortes}{LeCun and Cortes}{2010}]%
        {lecun-mnisthandwrittendigit-2010}
\bibfield{author}{\bibinfo{person}{Yann LeCun} {and} \bibinfo{person}{Corinna
  Cortes}.} \bibinfo{year}{2010}\natexlab{}.
\newblock \showarticletitle{{MNIST} handwritten digit database}.
\newblock \bibinfo{howpublished}{http://yann.lecun.com/exdb/mnist/}.
\newblock  (\bibinfo{year}{2010}).
\newblock
\urldef\tempurl%
\url{http://yann.lecun.com/exdb/mnist/}
\showURL{%
\tempurl}


\bibitem[\protect\citeauthoryear{Leung, Pazdor, and Souza}{Leung
  et~al\mbox{.}}{2021}]%
        {Leung-XAI-Customer-Churn-Interface}
\bibfield{author}{\bibinfo{person}{Carson~K. Leung}, \bibinfo{person}{Adam~G.M.
  Pazdor}, {and} \bibinfo{person}{Joglas Souza}.}
  \bibinfo{year}{2021}\natexlab{}.
\newblock \showarticletitle{Explainable Artificial Intelligence for Data
  Science on Customer Churn}. In \bibinfo{booktitle}{\emph{2021 IEEE 8th
  International Conference on Data Science and Advanced Analytics (DSAA)}}.
  \bibinfo{pages}{1--10}.
\newblock
\urldef\tempurl%
\url{https://doi.org/10.1109/DSAA53316.2021.9564166}
\showDOI{\tempurl}


\bibitem[\protect\citeauthoryear{Lewis}{Lewis}{1973}]%
        {Lewis1973:phil2}
\bibfield{author}{\bibinfo{person}{David Lewis}.}
  \bibinfo{year}{1973}\natexlab{}.
\newblock \bibinfo{booktitle}{\emph{Counterfactuals}}.
\newblock \bibinfo{publisher}{Blackwell Publishers}, \bibinfo{address}{Oxford}.
\newblock


\bibitem[\protect\citeauthoryear{Ley, Mishra, and Magazzeni}{Ley
  et~al\mbox{.}}{2022}]%
        {ley-global-cfe-ares-improve}
\bibfield{author}{\bibinfo{person}{Dan Ley}, \bibinfo{person}{Saumitra Mishra},
  {and} \bibinfo{person}{Daniele Magazzeni}.} \bibinfo{year}{2022}\natexlab{}.
\newblock \showarticletitle{Global Counterfactual Explanations: Investigations,
  Implementations and Improvements}. In \bibinfo{booktitle}{\emph{ICLR Workshop
  on Privacy, Accountability, Interpretability, Robustness, Reasoning on
  Structured Data}}.
\newblock


\bibitem[\protect\citeauthoryear{Li, Liu, Wu, Xi, Cao, and Cao}{Li
  et~al\mbox{.}}{2021}]%
        {li-cfe-chest-xray-images}
\bibfield{author}{\bibinfo{person}{Yan Li}, \bibinfo{person}{Shasha Liu},
  \bibinfo{person}{Chunwei Wu}, \bibinfo{person}{Xidong Xi},
  \bibinfo{person}{Guitao Cao}, {and} \bibinfo{person}{Wenming Cao}.}
  \bibinfo{year}{2021}\natexlab{}.
\newblock \showarticletitle{DCFG: Discovering Directional CounterFactual
  Generation for Chest X-rays}. In \bibinfo{booktitle}{\emph{2021 IEEE
  International Conference on Bioinformatics and Biomedicine (BIBM)}}.
  \bibinfo{pages}{972--979}.
\newblock
\urldef\tempurl%
\url{https://doi.org/10.1109/BIBM52615.2021.9669770}
\showDOI{\tempurl}


\bibitem[\protect\citeauthoryear{Liu, Kailkhura, Loveland, and Han}{Liu
  et~al\mbox{.}}{2019}]%
        {liu_generative_2019}
\bibfield{author}{\bibinfo{person}{Shusen Liu}, \bibinfo{person}{Bhavya
  Kailkhura}, \bibinfo{person}{Donald Loveland}, {and} \bibinfo{person}{Yong
  Han}.} \bibinfo{year}{2019}\natexlab{}.
\newblock \showarticletitle{Generative Counterfactual Introspection for
  Explainable Deep Learning}. In \bibinfo{booktitle}{\emph{2019 IEEE Global
  Conference on Signal and Information Processing (GlobalSIP)}}.
  \bibinfo{pages}{1--5}.
\newblock
\urldef\tempurl%
\url{https://doi.org/10.1109/GlobalSIP45357.2019.8969491}
\showDOI{\tempurl}


\bibitem[\protect\citeauthoryear{Liu, Luo, Wang, and Tang}{Liu
  et~al\mbox{.}}{2014}]%
        {celebA_data}
\bibfield{author}{\bibinfo{person}{Ziwei Liu}, \bibinfo{person}{Ping Luo},
  \bibinfo{person}{Xiaogang Wang}, {and} \bibinfo{person}{Xiaoou Tang}.}
  \bibinfo{year}{2014}\natexlab{}.
\newblock \showarticletitle{Deep Learning Face Attributes in the Wild}.
\newblock  (\bibinfo{date}{11} \bibinfo{year}{2014}).
\newblock
\urldef\tempurl%
\url{https://doi.org/10.1109/ICCV.2015.425}
\showDOI{\tempurl}


\bibitem[\protect\citeauthoryear{Lucic, Haned, and de~Rijke}{Lucic
  et~al\mbox{.}}{2020a}]%
        {contrastive-xai-retail-forecast}
\bibfield{author}{\bibinfo{person}{Ana Lucic}, \bibinfo{person}{Hinda Haned},
  {and} \bibinfo{person}{Maarten de Rijke}.} \bibinfo{year}{2020}\natexlab{a}.
\newblock \showarticletitle{Why Does My Model Fail? Contrastive Local
  Explanations for Retail Forecasting}. In
  \bibinfo{booktitle}{\emph{Proceedings of the 2020 Conference on Fairness,
  Accountability, and Transparency}} \emph{(\bibinfo{series}{FAT* '20})}.
  \bibinfo{publisher}{Association for Computing Machinery},
  \bibinfo{address}{New York, NY, USA}, 9.
\newblock
\urldef\tempurl%
\url{https://doi.org/10.1145/3351095.3372824}
\showDOI{\tempurl}


\bibitem[\protect\citeauthoryear{Lucic, Oosterhuis, Haned, and de~Rijke}{Lucic
  et~al\mbox{.}}{2019}]%
        {lucic-actionable:2020-update1}
\bibfield{author}{\bibinfo{person}{Ana Lucic}, \bibinfo{person}{Harrie
  Oosterhuis}, \bibinfo{person}{Hinda Haned}, {and} \bibinfo{person}{Maarten de
  Rijke}.} \bibinfo{year}{2019}\natexlab{}.
\newblock \bibinfo{title}{FOCUS: Flexible Optimizable Counterfactual
  Explanations for Tree Ensembles}.
\newblock
\newblock
\urldef\tempurl%
\url{https://doi.org/10.48550/ARXIV.1911.12199}
\showDOI{\tempurl}


\bibitem[\protect\citeauthoryear{Lucic, Oosterhuis, Haned, and de~Rijke}{Lucic
  et~al\mbox{.}}{2020b}]%
        {lucic-actionable:2020}
\bibfield{author}{\bibinfo{person}{Ana Lucic}, \bibinfo{person}{Harrie
  Oosterhuis}, \bibinfo{person}{Hinda Haned}, {and} \bibinfo{person}{Maarten de
  Rijke}.} \bibinfo{year}{2020}\natexlab{b}.
\newblock \bibinfo{title}{Actionable {Interpretability} through {Optimizable}
  {Counterfactual} {Explanations} for {Tree} {Ensembles}}.
\newblock
\newblock
\urldef\tempurl%
\url{http://arxiv.org/abs/1911.12199}
\showURL{%
\tempurl}


\bibitem[\protect\citeauthoryear{Lucic, ter Hoeve, Tolomei, de~Rijke, and
  Silvestri}{Lucic et~al\mbox{.}}{2021}]%
        {CF-GNN2}
\bibfield{author}{\bibinfo{person}{Ana Lucic}, \bibinfo{person}{Maartje ter
  Hoeve}, \bibinfo{person}{Gabriele Tolomei}, \bibinfo{person}{Maarten de
  Rijke}, {and} \bibinfo{person}{Fabrizio Silvestri}.}
  \bibinfo{year}{2021}\natexlab{}.
\newblock \bibinfo{title}{CF-GNNExplainer: Counterfactual Explanations for
  Graph Neural Networks}.
\newblock
\newblock
\showeprint[arxiv]{cs.LG/2102.03322}


\bibitem[\protect\citeauthoryear{Lundberg and Lee}{Lundberg and Lee}{2017}]%
        {shap-paper}
\bibfield{author}{\bibinfo{person}{Scott~M Lundberg} {and}
  \bibinfo{person}{Su-In Lee}.} \bibinfo{year}{2017}\natexlab{}.
\newblock \showarticletitle{A Unified Approach to Interpreting Model
  Predictions}.
\newblock In \bibinfo{booktitle}{\emph{Advances in Neural Information
  Processing Systems 30}}. \bibinfo{publisher}{Curran Associates, Inc.},
  \bibinfo{pages}{4765--4774}.
\newblock


\bibitem[\protect\citeauthoryear{Mac}{Mac}{2019}]%
        {freddiemac-data}
\bibfield{author}{\bibinfo{person}{Freddie Mac}.}
  \bibinfo{year}{2019}\natexlab{}.
\newblock \bibinfo{title}{Single family loan-level dataset}.
\newblock
\newblock
\urldef\tempurl%
\url{https://www.freddiemac.com/research/datasets/sf-loanlevel-dataset}
\showURL{%
\tempurl}


\bibitem[\protect\citeauthoryear{Madaan, Padhi, Panwar, and Saha}{Madaan
  et~al\mbox{.}}{2021}]%
        {Madaan-cfe-text}
\bibfield{author}{\bibinfo{person}{Nishtha Madaan}, \bibinfo{person}{Inkit
  Padhi}, \bibinfo{person}{Naveen Panwar}, {and} \bibinfo{person}{Diptikalyan
  Saha}.} \bibinfo{year}{2021}\natexlab{}.
\newblock \showarticletitle{Generate Your Counterfactuals: Towards Controlled
  Counterfactual Generation for Text}.
\newblock \bibinfo{journal}{\emph{Proceedings of the AAAI Conference on
  Artificial Intelligence}}  \bibinfo{volume}{35} (\bibinfo{date}{May}
  \bibinfo{year}{2021}), \bibinfo{pages}{13516--13524}.
\newblock
\urldef\tempurl%
\url{https://ojs.aaai.org/index.php/AAAI/article/view/17594}
\showURL{%
\tempurl}


\bibitem[\protect\citeauthoryear{Mae}{Mae}{2020}]%
        {fannie-data}
\bibfield{author}{\bibinfo{person}{Fannie Mae}.}
  \bibinfo{year}{2020}\natexlab{}.
\newblock \bibinfo{title}{Fannie Mae dataset}.
\newblock
\newblock
\urldef\tempurl%
\url{https://www.fanniemae.com/portal/funding-the-market/data/loan-performance-data.html}
\showURL{%
\tempurl}


\bibitem[\protect\citeauthoryear{Magrini, di~Blasi, and Stefanini}{Magrini
  et~al\mbox{.}}{2017}]%
        {Sangiovese-data}
\bibfield{author}{\bibinfo{person}{Alessandro Magrini},
  \bibinfo{person}{Stefano di Blasi}, {and} \bibinfo{person}{Federico
  Stefanini}.} \bibinfo{year}{2017}\natexlab{}.
\newblock \showarticletitle{A conditional linear Gaussian network to assess the
  impact of several agronomic settings on the quality of Tuscan Sangiovese
  grapes}.
\newblock \bibinfo{journal}{\emph{Biometrical Letters}}  \bibinfo{volume}{54}
  (\bibinfo{date}{06} \bibinfo{year}{2017}), \bibinfo{pages}{25--42}.
\newblock
\urldef\tempurl%
\url{https://doi.org/10.1515/bile-2017-0002}
\showDOI{\tempurl}


\bibitem[\protect\citeauthoryear{Mahajan, Tan, and Sharma}{Mahajan
  et~al\mbox{.}}{2020}]%
        {mahajan_preserving_2020}
\bibfield{author}{\bibinfo{person}{Divyat Mahajan}, \bibinfo{person}{Chenhao
  Tan}, {and} \bibinfo{person}{Amit Sharma}.} \bibinfo{year}{2020}\natexlab{}.
\newblock \bibinfo{title}{Preserving {Causal} {Constraints} in {Counterfactual}
  {Explanations} for {Machine} {Learning} {Classifiers}}.
\newblock
\newblock
\urldef\tempurl%
\url{http://arxiv.org/abs/1912.03277}
\showURL{%
\tempurl}


\bibitem[\protect\citeauthoryear{Marchezini, Lacerda, Pappa, Meira, Miranda,
  Romano-Silva, Costa, and Diniz}{Marchezini et~al\mbox{.}}{2022}]%
        {Marchezini2022-dh-cfe-latent-mental-health}
\bibfield{author}{\bibinfo{person}{Guilherme~F Marchezini},
  \bibinfo{person}{Anisio~M Lacerda}, \bibinfo{person}{Gisele~L Pappa},
  \bibinfo{person}{Wagner Meira, Jr}, \bibinfo{person}{Debora Miranda},
  \bibinfo{person}{Marco~A Romano-Silva}, \bibinfo{person}{Danielle~S Costa},
  {and} \bibinfo{person}{Leandro~Malloy Diniz}.}
  \bibinfo{year}{2022}\natexlab{}.
\newblock \showarticletitle{Counterfactual inference with latent variable and
  its application in mental health care}.
\newblock \bibinfo{journal}{\emph{Data Min. Knowl. Discov.}}
  \bibinfo{volume}{36}, \bibinfo{number}{2} (\bibinfo{date}{Jan.}
  \bibinfo{year}{2022}), \bibinfo{pages}{811--840}.
\newblock


\bibitem[\protect\citeauthoryear{Martens and Provost}{Martens and
  Provost}{2014}]%
        {Provost1}
\bibfield{author}{\bibinfo{person}{David Martens} {and}
  \bibinfo{person}{Foster~J. Provost}.} \bibinfo{year}{2014}\natexlab{}.
\newblock \showarticletitle{Explaining Data-Driven Document Classifications}.
\newblock \bibinfo{journal}{\emph{MIS Q.}}  \bibinfo{volume}{38}
  (\bibinfo{year}{2014}), \bibinfo{pages}{73--99}.
\newblock


\bibitem[\protect\citeauthoryear{Mazzine, Goethals, Brughmans, and
  Martens}{Mazzine et~al\mbox{.}}{2021}]%
        {mazzine-CFE-employment}
\bibfield{author}{\bibinfo{person}{Raphael Mazzine}, \bibinfo{person}{Sofie
  Goethals}, \bibinfo{person}{Dieter Brughmans}, {and} \bibinfo{person}{David
  Martens}.} \bibinfo{year}{2021}\natexlab{}.
\newblock \showarticletitle{Counterfactual Explanations for Employment
  Services}. In \bibinfo{booktitle}{\emph{International workshop on Fair,
  Effective And Sustainable Talent management using data science}}.
  \bibinfo{pages}{1--7}.
\newblock


\bibitem[\protect\citeauthoryear{Mazzine and Martens}{Mazzine and
  Martens}{2021}]%
        {cfe-algos-quantitative-comparison-survey}
\bibfield{author}{\bibinfo{person}{Raphael Mazzine} {and}
  \bibinfo{person}{David Martens}.} \bibinfo{year}{2021}\natexlab{}.
\newblock \bibinfo{title}{A Framework and Benchmarking Study for Counterfactual
  Generating Methods on Tabular Data}.
\newblock
\newblock
\urldef\tempurl%
\url{https://doi.org/10.48550/ARXIV.2107.04680}
\showDOI{\tempurl}


\bibitem[\protect\citeauthoryear{Medeiros~Raimundo, Nonato, and
  Poco}{Medeiros~Raimundo et~al\mbox{.}}{2021}]%
        {cfe-for-monotone-obj}
\bibfield{author}{\bibinfo{person}{Marcos Medeiros~Raimundo},
  \bibinfo{person}{Luis Nonato}, {and} \bibinfo{person}{Jorge Poco}.}
  \bibinfo{year}{2021}\natexlab{}.
\newblock \bibinfo{title}{Mining Pareto-Optimal Counterfactual Antecedents With
  A Branch-And-Bound Model-Agnostic Algorithm}.
\newblock
\newblock
\urldef\tempurl%
\url{https://doi.org/10.21203/rs.3.rs-551661/v1}
\showDOI{\tempurl}


\bibitem[\protect\citeauthoryear{Mehedi~Hasan and Talbert}{Mehedi~Hasan and
  Talbert}{2022}]%
        {hasan-cfe-augmentation-proximity}
\bibfield{author}{\bibinfo{person}{Md~Golam~Moula Mehedi~Hasan} {and}
  \bibinfo{person}{Douglas Talbert}.} \bibinfo{year}{2022}\natexlab{}.
\newblock \showarticletitle{Data Augmentation using Counterfactuals: Proximity
  vs Diversity}.
\newblock \bibinfo{journal}{\emph{The International FLAIRS Conference
  Proceedings}}  \bibinfo{volume}{35} (\bibinfo{date}{May}
  \bibinfo{year}{2022}).
\newblock
\urldef\tempurl%
\url{https://doi.org/10.32473/flairs.v35i.130705}
\showDOI{\tempurl}


\bibitem[\protect\citeauthoryear{Mertes, Huber, Weitz, Heimerl, and
  André}{Mertes et~al\mbox{.}}{2022}]%
        {Mertes-GANterfactual-Medical-Images}
\bibfield{author}{\bibinfo{person}{Silvan Mertes}, \bibinfo{person}{Tobias
  Huber}, \bibinfo{person}{Katharina Weitz}, \bibinfo{person}{Alexander
  Heimerl}, {and} \bibinfo{person}{Elisabeth André}.}
  \bibinfo{year}{2022}\natexlab{}.
\newblock \showarticletitle{GANterfactual—Counterfactual Explanations for
  Medical Non-experts Using Generative Adversarial Learning}.
\newblock \bibinfo{journal}{\emph{Frontiers in Artificial Intelligence}}
  \bibinfo{volume}{5} (\bibinfo{year}{2022}).
\newblock
\urldef\tempurl%
\url{https://doi.org/10.3389/frai.2022.825565}
\showDOI{\tempurl}


\bibitem[\protect\citeauthoryear{Miller}{Miller}{2019}]%
        {Miller-xai:2019}
\bibfield{author}{\bibinfo{person}{Tim Miller}.}
  \bibinfo{year}{2019}\natexlab{}.
\newblock \showarticletitle{Explanation in artificial intelligence: Insights
  from the social sciences}.
\newblock \bibinfo{journal}{\emph{Artificial Intelligence}}
  (\bibinfo{date}{Feb} \bibinfo{year}{2019}), \bibinfo{pages}{1–38}.
\newblock
\urldef\tempurl%
\url{https://doi.org/10.1016/j.artint.2018.07.007}
\showDOI{\tempurl}


\bibitem[\protect\citeauthoryear{Mishra, Dutta, Long, and Magazzeni}{Mishra
  et~al\mbox{.}}{2021}]%
        {mishra-cfe-robustness-survey}
\bibfield{author}{\bibinfo{person}{Saumitra Mishra},
  \bibinfo{person}{Sanghamitra Dutta}, \bibinfo{person}{Jason Long}, {and}
  \bibinfo{person}{Daniele Magazzeni}.} \bibinfo{year}{2021}\natexlab{}.
\newblock \bibinfo{title}{A Survey on the Robustness of Feature Importance and
  Counterfactual Explanations}.
\newblock
\newblock
\urldef\tempurl%
\url{https://doi.org/10.48550/ARXIV.2111.00358}
\showDOI{\tempurl}


\bibitem[\protect\citeauthoryear{Miura, Hasegawa, and Shibahara}{Miura
  et~al\mbox{.}}{2021}]%
        {Miura2021MEGEX-privacy-attack-cfe}
\bibfield{author}{\bibinfo{person}{Takayuki Miura}, \bibinfo{person}{Satoshi
  Hasegawa}, {and} \bibinfo{person}{Toshiki Shibahara}.}
  \bibinfo{year}{2021}\natexlab{}.
\newblock \showarticletitle{MEGEX: Data-Free Model Extraction Attack against
  Gradient-Based Explainable AI}.
\newblock \bibinfo{journal}{\emph{ArXiv}}  \bibinfo{volume}{abs/2107.08909}
  (\bibinfo{year}{2021}).
\newblock


\bibitem[\protect\citeauthoryear{Mohammadi, Karimi, Barthe, and
  Valera}{Mohammadi et~al\mbox{.}}{2021}]%
        {scaling_Nearest_CFE}
\bibfield{author}{\bibinfo{person}{Kiarash Mohammadi},
  \bibinfo{person}{Amir-Hossein Karimi}, \bibinfo{person}{Gilles Barthe}, {and}
  \bibinfo{person}{Isabel Valera}.} \bibinfo{year}{2021}\natexlab{}.
\newblock \showarticletitle{Scaling Guarantees for Nearest Counterfactual
  Explanations}. In \bibinfo{booktitle}{\emph{Proceedings of the 2021 AAAI/ACM
  Conference on AI, Ethics, and Society}}. \bibinfo{publisher}{Association for
  Computing Machinery}, \bibinfo{address}{New York, NY, USA},
  \bibinfo{pages}{177–187}.
\newblock
\urldef\tempurl%
\url{https://doi.org/10.1145/3461702.3462514}
\showDOI{\tempurl}


\bibitem[\protect\citeauthoryear{Monteiro and Reynoso-Meza}{Monteiro and
  Reynoso-Meza}{2022}]%
        {another-genetic-algo-monteiro}
\bibfield{author}{\bibinfo{person}{Wellington~Rodrigo Monteiro} {and}
  \bibinfo{person}{Gilberto Reynoso-Meza}.} \bibinfo{year}{2022}\natexlab{}.
\newblock \showarticletitle{Counterfactual Generation Through Multi-objective
  Constrained Optimisation}.
\newblock  (\bibinfo{year}{2022}), 23.
\newblock
\urldef\tempurl%
\url{https://doi.org/10.21203/rs.3.rs-1325730/v1}
\showDOI{\tempurl}


\bibitem[\protect\citeauthoryear{Moro, Cortez, and Rita}{Moro
  et~al\mbox{.}}{2014}]%
        {Portuguese_Bank_data}
\bibfield{author}{\bibinfo{person}{Sérgio Moro}, \bibinfo{person}{Paulo
  Cortez}, {and} \bibinfo{person}{Paulo Rita}.}
  \bibinfo{year}{2014}\natexlab{}.
\newblock \showarticletitle{A data-driven approach to predict the success of
  bank telemarketing}.
\newblock \bibinfo{journal}{\emph{Decision Support Systems}}
  \bibinfo{volume}{62} (\bibinfo{year}{2014}), \bibinfo{pages}{22--31}.
\newblock
\urldef\tempurl%
\url{https://doi.org/10.1016/j.dss.2014.03.001}
\showDOI{\tempurl}


\bibitem[\protect\citeauthoryear{Mothilal, Sharma, and Tan}{Mothilal
  et~al\mbox{.}}{2020}]%
        {mothilal_explaining_2020}
\bibfield{author}{\bibinfo{person}{Ramaravind~K. Mothilal},
  \bibinfo{person}{Amit Sharma}, {and} \bibinfo{person}{Chenhao Tan}.}
  \bibinfo{year}{2020}\natexlab{}.
\newblock \showarticletitle{Explaining Machine Learning Classifiers through
  Diverse Counterfactual Explanations}. In
  \bibinfo{booktitle}{\emph{Proceedings of the Conference on Fairness,
  Accountability, and Transparency (FAccT)}} \emph{(\bibinfo{series}{FAT*
  '20})}. \bibinfo{publisher}{Association for Computing Machinery},
  \bibinfo{address}{New York, NY, USA}.
\newblock
\urldef\tempurl%
\url{https://doi.org/10.1145/3351095.3372850}
\showDOI{\tempurl}


\bibitem[\protect\citeauthoryear{Mueller, Weiner, Thal, Petersen, Jack, Jagust,
  Trojanowski, Toga, and Beckett}{Mueller et~al\mbox{.}}{2008}]%
        {alzheimer-data}
\bibfield{author}{\bibinfo{person}{Susanne~G. Mueller},
  \bibinfo{person}{Michael~W. Weiner}, \bibinfo{person}{Leon~J. Thal},
  \bibinfo{person}{Ronald~C. Petersen}, \bibinfo{person}{Clifford Jack},
  \bibinfo{person}{William Jagust}, \bibinfo{person}{John~Q. Trojanowski},
  \bibinfo{person}{Arthur~W. Toga}, {and} \bibinfo{person}{Laurel Beckett}.}
  \bibinfo{year}{2008}\natexlab{}.
\newblock \showarticletitle{Alzheimer's Disease Neuroimaging Initiative}. In
  \bibinfo{booktitle}{\emph{Advances in Alzheimer's and Parkinson's Disease}}.
  \bibinfo{publisher}{Springer US}, \bibinfo{address}{Boston, MA},
  \bibinfo{pages}{183--189}.
\newblock


\bibitem[\protect\citeauthoryear{Myers, Freed, Pardo, Furqan, Risi, and
  Zhu}{Myers et~al\mbox{.}}{2020}]%
        {cfe-model-bias1}
\bibfield{author}{\bibinfo{person}{Chelsea~M. Myers}, \bibinfo{person}{Evan
  Freed}, \bibinfo{person}{Luis Fernando~Laris Pardo}, \bibinfo{person}{Anushay
  Furqan}, \bibinfo{person}{Sebastian Risi}, {and} \bibinfo{person}{Jichen
  Zhu}.} \bibinfo{year}{2020}\natexlab{}.
\newblock \bibinfo{title}{Revealing Neural Network Bias to Non-Experts Through
  Interactive Counterfactual Examples}.
\newblock
\newblock
\urldef\tempurl%
\url{https://doi.org/10.48550/ARXIV.2001.02271}
\showDOI{\tempurl}


\bibitem[\protect\citeauthoryear{Naumann and Ntoutsi}{Naumann and
  Ntoutsi}{2021}]%
        {naumann2021consequenceaware}
\bibfield{author}{\bibinfo{person}{Philip Naumann} {and}
  \bibinfo{person}{Eirini Ntoutsi}.} \bibinfo{year}{2021}\natexlab{}.
\newblock \bibinfo{title}{Consequence-aware Sequential Counterfactual
  Generation}.
\newblock
\newblock
\showeprint[arxiv]{cs.LG/2104.05592}


\bibitem[\protect\citeauthoryear{Navas{-}Palencia}{Navas{-}Palencia}{2021}]%
        {CFE-scorecard}
\bibfield{author}{\bibinfo{person}{Guillermo Navas{-}Palencia}.}
  \bibinfo{year}{2021}\natexlab{}.
\newblock \bibinfo{title}{Optimal Counterfactual Explanations for Scorecard
  modelling}.
\newblock
\newblock
\urldef\tempurl%
\url{https://arxiv.org/abs/2104.08619}
\showURL{%
\tempurl}


\bibitem[\protect\citeauthoryear{Nemirovsky, Thiebaut, Xu, and
  Gupta}{Nemirovsky et~al\mbox{.}}{2021}]%
        {nemirovsky-cfe-gan-hiring-application}
\bibfield{author}{\bibinfo{person}{Daniel Nemirovsky}, \bibinfo{person}{Nicolas
  Thiebaut}, \bibinfo{person}{Ye Xu}, {and} \bibinfo{person}{Abhishek Gupta}.}
  \bibinfo{year}{2021}\natexlab{}.
\newblock \showarticletitle{Providing Actionable Feedback in Hiring
  Marketplaces Using Generative Adversarial Networks}. In
  \bibinfo{booktitle}{\emph{Proceedings of the 14th ACM International
  Conference on Web Search and Data Mining}}. \bibinfo{publisher}{Association
  for Computing Machinery}, \bibinfo{address}{New York, NY, USA}, 4.
\newblock
\urldef\tempurl%
\url{https://doi.org/10.1145/3437963.3441705}
\showDOI{\tempurl}


\bibitem[\protect\citeauthoryear{Nemirovsky, Thiebaut, Xu, and
  Gupta}{Nemirovsky et~al\mbox{.}}{2022}]%
        {nemirovsky-hired-people-cfe-countergan}
\bibfield{author}{\bibinfo{person}{Daniel Nemirovsky}, \bibinfo{person}{Nicolas
  Thiebaut}, \bibinfo{person}{Ye Xu}, {and} \bibinfo{person}{Abhishek Gupta}.}
  \bibinfo{year}{2022}\natexlab{}.
\newblock \showarticletitle{CounteRGAN: Generating counterfactuals for
  real-time recourse and interpretability using residual GANs}. In
  \bibinfo{booktitle}{\emph{Proceedings of the Thirty-Eighth Conference on
  Uncertainty in Artificial Intelligence}} \emph{(\bibinfo{series}{Proceedings
  of Machine Learning Research})}. \bibinfo{publisher}{PMLR},
  \bibinfo{pages}{1488--1497}.
\newblock
\urldef\tempurl%
\url{https://proceedings.mlr.press/v180/nemirovsky22a.html}
\showURL{%
\tempurl}


\bibitem[\protect\citeauthoryear{Nguyen, Quinn, Nguyen, and Tran}{Nguyen
  et~al\mbox{.}}{2021}]%
        {nguyen2021-cfe-dta}
\bibfield{author}{\bibinfo{person}{Tri~Minh Nguyen}, \bibinfo{person}{Thomas~P
  Quinn}, \bibinfo{person}{Thin Nguyen}, {and} \bibinfo{person}{Truyen Tran}.}
  \bibinfo{year}{2021}\natexlab{}.
\newblock \bibinfo{title}{Counterfactual Explanation with Multi-Agent
  Reinforcement Learning for Drug Target Prediction}.
\newblock
\newblock
\showeprint[arxiv]{cs.AI/2103.12983}


\bibitem[\protect\citeauthoryear{Numeroso and Bacciu}{Numeroso and
  Bacciu}{2021}]%
        {CF-GNN1}
\bibfield{author}{\bibinfo{person}{Danilo Numeroso} {and}
  \bibinfo{person}{Davide Bacciu}.} \bibinfo{year}{2021}\natexlab{}.
\newblock \bibinfo{title}{MEG: Generating Molecular Counterfactual Explanations
  for Deep Graph Networks}.
\newblock
\newblock


\bibitem[\protect\citeauthoryear{O'Brien and Kim}{O'Brien and Kim}{2021}]%
        {multi-agent-recourse}
\bibfield{author}{\bibinfo{person}{Andrew O'Brien} {and}
  \bibinfo{person}{Edward Kim}.} \bibinfo{year}{2021}\natexlab{}.
\newblock \bibinfo{title}{Multi-Agent Algorithmic Recourse}.
\newblock
\newblock
\urldef\tempurl%
\url{https://doi.org/10.48550/ARXIV.2110.00673}
\showDOI{\tempurl}


\bibitem[\protect\citeauthoryear{of~Commons}{of~Commons}{[n. d.]}]%
        {UK-XAI1}
\bibfield{author}{\bibinfo{person}{House of Commons}.} \bibinfo{year}{[n.
  d.]}\natexlab{}.
\newblock \bibinfo{title}{Algorithms in decision making}.
\newblock
  \bibinfo{howpublished}{\url{https://publications.parliament.uk/pa/cm201719/cmselect/cmsctech/351/351.pdf}}.
\newblock
\newblock
\shownote{Accessed: 2020-10-15.}


\bibitem[\protect\citeauthoryear{Oh, Yoon, and Suk}{Oh et~al\mbox{.}}{2020}]%
        {oh-cfe-images-BIN}
\bibfield{author}{\bibinfo{person}{Kwanseok Oh}, \bibinfo{person}{Jee~Seok
  Yoon}, {and} \bibinfo{person}{Heung-Il Suk}.}
  \bibinfo{year}{2020}\natexlab{}.
\newblock \bibinfo{title}{Born Identity Network: Multi-way Counterfactual Map
  Generation to Explain a Classifier's Decision}.
\newblock
\newblock
\urldef\tempurl%
\url{https://doi.org/10.48550/ARXIV.2011.10381}
\showDOI{\tempurl}


\bibitem[\protect\citeauthoryear{Oh, Yoon, and Suk}{Oh et~al\mbox{.}}{2021}]%
        {oh-cfe-alzheimer-images}
\bibfield{author}{\bibinfo{person}{Kwanseok Oh}, \bibinfo{person}{Jee~Seok
  Yoon}, {and} \bibinfo{person}{Heung-Il Suk}.}
  \bibinfo{year}{2021}\natexlab{}.
\newblock \bibinfo{title}{Learn-Explain-Reinforce: Counterfactual Reasoning and
  Its Guidance to Reinforce an Alzheimer's Disease Diagnosis Model}.
\newblock
\newblock
\urldef\tempurl%
\url{https://doi.org/10.48550/ARXIV.2108.09451}
\showDOI{\tempurl}


\bibitem[\protect\citeauthoryear{Olson, Khanna, Neal, Li, and Wong}{Olson
  et~al\mbox{.}}{2021}]%
        {CFE_for_rl}
\bibfield{author}{\bibinfo{person}{Matthew~L. Olson}, \bibinfo{person}{Roli
  Khanna}, \bibinfo{person}{Lawrence Neal}, \bibinfo{person}{Fuxin Li}, {and}
  \bibinfo{person}{Weng-Keen Wong}.} \bibinfo{year}{2021}\natexlab{}.
\newblock \showarticletitle{Counterfactual state explanations for reinforcement
  learning agents via generative deep learning}.
\newblock \bibinfo{journal}{\emph{Artificial Intelligence}}
  \bibinfo{volume}{295} (\bibinfo{year}{2021}), \bibinfo{pages}{103455}.
\newblock
\urldef\tempurl%
\url{https://doi.org/10.1016/j.artint.2021.103455}
\showDOI{\tempurl}


\bibitem[\protect\citeauthoryear{Parmentier and Vidal}{Parmentier and
  Vidal}{2021}]%
        {CFE-Tree-ensembles}
\bibfield{author}{\bibinfo{person}{Axel Parmentier} {and}
  \bibinfo{person}{Thibaut Vidal}.} \bibinfo{year}{2021}\natexlab{}.
\newblock \bibinfo{title}{Optimal Counterfactual Explanations in Tree
  Ensembles}.
\newblock
\newblock
\urldef\tempurl%
\url{https://arxiv.org/abs/2106.06631}
\showURL{%
\tempurl}


\bibitem[\protect\citeauthoryear{Pawelczyk, Agarwal, Joshi, Upadhyay, and
  Lakkaraju}{Pawelczyk et~al\mbox{.}}{2022a}]%
        {pawelczyk2021_CFE_AE_connection}
\bibfield{author}{\bibinfo{person}{Martin Pawelczyk}, \bibinfo{person}{Chirag
  Agarwal}, \bibinfo{person}{Shalmali Joshi}, \bibinfo{person}{Sohini
  Upadhyay}, {and} \bibinfo{person}{Himabindu Lakkaraju}.}
  \bibinfo{year}{2022}\natexlab{a}.
\newblock \showarticletitle{Exploring Counterfactual Explanations Through the
  Lens of Adversarial Examples: A Theoretical and Empirical Analysis}. In
  \bibinfo{booktitle}{\emph{Proceedings of The 25th International Conference on
  Artificial Intelligence and Statistics}} \emph{(\bibinfo{series}{Proceedings
  of Machine Learning Research})}, Vol.~\bibinfo{volume}{151}.
  \bibinfo{publisher}{PMLR}, \bibinfo{pages}{4574--4594}.
\newblock
\urldef\tempurl%
\url{https://proceedings.mlr.press/v151/pawelczyk22a.html}
\showURL{%
\tempurl}


\bibitem[\protect\citeauthoryear{Pawelczyk, Bielawski, van~den Heuvel, Richter,
  and Kasneci}{Pawelczyk et~al\mbox{.}}{2021}]%
        {pawelczyk2021CARLA-toolbox}
\bibfield{author}{\bibinfo{person}{Martin Pawelczyk}, \bibinfo{person}{Sascha
  Bielawski}, \bibinfo{person}{Johannes van~den Heuvel},
  \bibinfo{person}{Tobias Richter}, {and} \bibinfo{person}{Gjergji Kasneci}.}
  \bibinfo{year}{2021}\natexlab{}.
\newblock \bibinfo{title}{CARLA: A Python Library to Benchmark Algorithmic
  Recourse and Counterfactual Explanation Algorithms}.
\newblock
\newblock
\showeprint[arxiv]{cs.LG/2108.00783}


\bibitem[\protect\citeauthoryear{Pawelczyk, Broelemann, and Kasneci}{Pawelczyk
  et~al\mbox{.}}{2020a}]%
        {predictive_multiplicity}
\bibfield{author}{\bibinfo{person}{Martin Pawelczyk}, \bibinfo{person}{Klaus
  Broelemann}, {and} \bibinfo{person}{Gjergji. Kasneci}.}
  \bibinfo{year}{2020}\natexlab{a}.
\newblock \showarticletitle{On Counterfactual Explanations under Predictive
  Multiplicity}. In \bibinfo{booktitle}{\emph{Proceedings of Machine Learning
  Research}}. \bibinfo{publisher}{PMLR}, \bibinfo{address}{Virtual}, 9.
\newblock
\urldef\tempurl%
\url{http://proceedings.mlr.press/v124/pawelczyk20a.html}
\showURL{%
\tempurl}


\bibitem[\protect\citeauthoryear{Pawelczyk, Datta, van-den Heuvel, Kasneci, and
  Lakkaraju}{Pawelczyk et~al\mbox{.}}{2022b}]%
        {probabilistically-robust-cfe-hima-group}
\bibfield{author}{\bibinfo{person}{Martin Pawelczyk}, \bibinfo{person}{Teresa
  Datta}, \bibinfo{person}{Johannes van-den Heuvel}, \bibinfo{person}{Gjergji
  Kasneci}, {and} \bibinfo{person}{Himabindu Lakkaraju}.}
  \bibinfo{year}{2022}\natexlab{b}.
\newblock \bibinfo{title}{Probabilistically Robust Recourse: Navigating the
  Trade-offs between Costs and Robustness in Algorithmic Recourse}.
\newblock
\newblock
\urldef\tempurl%
\url{https://doi.org/10.48550/ARXIV.2203.06768}
\showDOI{\tempurl}


\bibitem[\protect\citeauthoryear{Pawelczyk, Haug, Broelemann, and
  Kasneci}{Pawelczyk et~al\mbox{.}}{2020b}]%
        {pawelczyk_learning_2020}
\bibfield{author}{\bibinfo{person}{Martin Pawelczyk}, \bibinfo{person}{Johannes
  Haug}, \bibinfo{person}{Klaus Broelemann}, {and} \bibinfo{person}{Gjergji
  Kasneci}.} \bibinfo{year}{2020}\natexlab{b}.
\newblock \bibinfo{title}{Learning Model-Agnostic Counterfactual Explanations
  for Tabular Data}.
\newblock , \bibinfo{numpages}{3126--3132}~pages.
\newblock
\urldef\tempurl%
\url{https://doi.org/10.1145/3366423.3380087}
\showDOI{\tempurl}


\bibitem[\protect\citeauthoryear{Pearl}{Pearl}{2000}]%
        {causality:Pearl}
\bibfield{author}{\bibinfo{person}{Judea Pearl}.}
  \bibinfo{year}{2000}\natexlab{}.
\newblock \bibinfo{booktitle}{\emph{Causality: Models, Reasoning, and
  Inference}}.
\newblock \bibinfo{publisher}{Cambridge University Press},
  \bibinfo{address}{USA}.
\newblock
\showISBNx{0521773628}


\bibitem[\protect\citeauthoryear{Pedapati, Balakrishnan, Shanmugan, and
  Dhurandhar}{Pedapati et~al\mbox{.}}{2020}]%
        {Dhurandhar-global-model-consistent-with-cfe}
\bibfield{author}{\bibinfo{person}{Tejaswini Pedapati},
  \bibinfo{person}{Avinash Balakrishnan}, \bibinfo{person}{Karthikeyan
  Shanmugan}, {and} \bibinfo{person}{Amit Dhurandhar}.}
  \bibinfo{year}{2020}\natexlab{}.
\newblock \showarticletitle{Learning Global Transparent Models Consistent with
  Local Contrastive Explanations}. In \bibinfo{booktitle}{\emph{Proceedings of
  the 34th International Conference on Neural Information Processing Systems}}
  \emph{(\bibinfo{series}{NIPS'20})}. \bibinfo{publisher}{Curran Associates
  Inc.}, \bibinfo{address}{Red Hook, NY, USA}, 11.
\newblock


\bibitem[\protect\citeauthoryear{Popescu, Shadaydeh, and Denzler}{Popescu
  et~al\mbox{.}}{2021}]%
        {cfe_knockoff_images}
\bibfield{author}{\bibinfo{person}{Oana-Iuliana Popescu}, \bibinfo{person}{Maha
  Shadaydeh}, {and} \bibinfo{person}{Joachim Denzler}.}
  \bibinfo{year}{2021}\natexlab{}.
\newblock \bibinfo{title}{Counterfactual Generation with Knockoffs}.
\newblock
\newblock
\urldef\tempurl%
\url{https://doi.org/10.48550/ARXIV.2102.00951}
\showDOI{\tempurl}


\bibitem[\protect\citeauthoryear{Poyiadzi, Sokol, Santos-Rodriguez, De~Bie, and
  Flach}{Poyiadzi et~al\mbox{.}}{2020}]%
        {poyiadzi_face_2020}
\bibfield{author}{\bibinfo{person}{Rafael Poyiadzi}, \bibinfo{person}{Kacper
  Sokol}, \bibinfo{person}{Raul Santos-Rodriguez}, \bibinfo{person}{Tijl
  De~Bie}, {and} \bibinfo{person}{Peter Flach}.}
  \bibinfo{year}{2020}\natexlab{}.
\newblock \bibinfo{title}{{FACE}: {Feasible} and {Actionable} {Counterfactual}
  {Explanations}}.
\newblock , \bibinfo{numpages}{344--350}~pages.
\newblock
\urldef\tempurl%
\url{https://doi.org/10.1145/3375627.3375850}
\showDOI{\tempurl}
\newblock
\shownote{arXiv: 1909.09369.}


\bibitem[\protect\citeauthoryear{Prado-Romero, Prenkaj, Stilo, and
  Giannotti}{Prado-Romero et~al\mbox{.}}{2022}]%
        {cfe-graph-nn-survey}
\bibfield{author}{\bibinfo{person}{Mario~Alfonso Prado-Romero},
  \bibinfo{person}{Bardh Prenkaj}, \bibinfo{person}{Giovanni Stilo}, {and}
  \bibinfo{person}{Fosca Giannotti}.} \bibinfo{year}{2022}\natexlab{}.
\newblock \bibinfo{title}{A Survey on Graph Counterfactual Explanations:
  Definitions, Methods, Evaluation}.
\newblock
\newblock
\urldef\tempurl%
\url{https://doi.org/10.48550/ARXIV.2210.12089}
\showDOI{\tempurl}


\bibitem[\protect\citeauthoryear{Qi and Chelmis}{Qi and Chelmis}{2021}]%
        {debugging-data-cfe2}
\bibfield{author}{\bibinfo{person}{Wenting Qi} {and}
  \bibinfo{person}{Charalampos Chelmis}.} \bibinfo{year}{2021}\natexlab{}.
\newblock \showarticletitle{Improving Algorithmic Decision–Making in the
  Presence of Untrustworthy Training Data}. In \bibinfo{booktitle}{\emph{2021
  IEEE International Conference on Big Data (Big Data)}}.
  \bibinfo{pages}{1102--1108}.
\newblock
\urldef\tempurl%
\url{https://doi.org/10.1109/BigData52589.2021.9671677}
\showDOI{\tempurl}


\bibitem[\protect\citeauthoryear{Ramakrishnan, Lee, and
  Albarghouthi}{Ramakrishnan et~al\mbox{.}}{2020}]%
        {ramakrishnan_synthesizing_2019}
\bibfield{author}{\bibinfo{person}{Goutham Ramakrishnan},
  \bibinfo{person}{Y.~C. Lee}, {and} \bibinfo{person}{Aws Albarghouthi}.}
  \bibinfo{year}{2020}\natexlab{}.
\newblock \showarticletitle{Synthesizing Action Sequences for Modifying Model
  Decisions}. In \bibinfo{booktitle}{\emph{Conference on Artificial
  Intelligence (AAAI)}}. \bibinfo{publisher}{AAAI press},
  \bibinfo{address}{California, USA}, 16.
\newblock
\urldef\tempurl%
\url{http://arxiv.org/abs/1910.00057}
\showURL{%
\tempurl}


\bibitem[\protect\citeauthoryear{Ramon, Martens, Provost, and Evgeniou}{Ramon
  et~al\mbox{.}}{2020}]%
        {ramon-cfe-comparison-text+behavior}
\bibfield{author}{\bibinfo{person}{Yanou Ramon}, \bibinfo{person}{David
  Martens}, \bibinfo{person}{Foster Provost}, {and} \bibinfo{person}{Theodoros
  Evgeniou}.} \bibinfo{year}{2020}\natexlab{}.
\newblock \showarticletitle{A Comparison of Instance-Level Counterfactual
  Explanation Algorithms for Behavioral and Textual Data: SEDC, LIME-C and
  SHAP-C}.
\newblock  \bibinfo{volume}{14}, \bibinfo{number}{4} (\bibinfo{year}{2020}),
  \bibinfo{pages}{801–819}.
\newblock
\urldef\tempurl%
\url{https://doi.org/10.1007/s11634-020-00418-3}
\showDOI{\tempurl}


\bibitem[\protect\citeauthoryear{Rasouli and Chieh~Yu}{Rasouli and
  Chieh~Yu}{2022}]%
        {rasouli2022CARE-CFE}
\bibfield{author}{\bibinfo{person}{Peyman Rasouli} {and}
  \bibinfo{person}{Ingrid Chieh~Yu}.} \bibinfo{year}{2022}\natexlab{}.
\newblock \showarticletitle{CARE: Coherent actionable recourse based on sound
  counterfactual explanations}.
\newblock \bibinfo{journal}{\emph{International Journal of Data Science and
  Analytics}} (\bibinfo{year}{2022}), \bibinfo{pages}{1--26}.
\newblock


\bibitem[\protect\citeauthoryear{Rasouli and Yu}{Rasouli and Yu}{2021}]%
        {rasouli-cfe-robustness-tabular}
\bibfield{author}{\bibinfo{person}{Peyman Rasouli} {and}
  \bibinfo{person}{Ingrid~Chieh Yu}.} \bibinfo{year}{2021}\natexlab{}.
\newblock \showarticletitle{Analyzing and Improving the Robustness of Tabular
  Classifiers using Counterfactual Explanations}. In
  \bibinfo{booktitle}{\emph{2021 20th IEEE International Conference on Machine
  Learning and Applications (ICMLA)}}. \bibinfo{pages}{1286--1293}.
\newblock
\urldef\tempurl%
\url{https://doi.org/10.1109/ICMLA52953.2021.00209}
\showDOI{\tempurl}


\bibitem[\protect\citeauthoryear{Rathi}{Rathi}{2019}]%
        {rathi-generating:2019}
\bibfield{author}{\bibinfo{person}{Shubham Rathi}.}
  \bibinfo{year}{2019}\natexlab{}.
\newblock \bibinfo{title}{Generating Counterfactual and Contrastive
  Explanations using SHAP}.
\newblock
\newblock
\urldef\tempurl%
\url{http://arxiv.org/abs/1906.09293}
\showURL{%
\tempurl}
\newblock
\shownote{arXiv: 1906.09293.}


\bibitem[\protect\citeauthoryear{Ravfogel, Prasad, Linzen, and
  Goldberg}{Ravfogel et~al\mbox{.}}{2021}]%
        {ravfogel-2021-cfe-text}
\bibfield{author}{\bibinfo{person}{Shauli Ravfogel}, \bibinfo{person}{Grusha
  Prasad}, \bibinfo{person}{Tal Linzen}, {and} \bibinfo{person}{Yoav
  Goldberg}.} \bibinfo{year}{2021}\natexlab{}.
\newblock \showarticletitle{Counterfactual Interventions Reveal the Causal
  Effect of Relative Clause Representations on Agreement Prediction}. In
  \bibinfo{booktitle}{\emph{Proceedings of the 25th Conference on Computational
  Natural Language Learning}}. \bibinfo{publisher}{Association for
  Computational Linguistics}, \bibinfo{pages}{194--209}.
\newblock
\urldef\tempurl%
\url{https://doi.org/10.18653/v1/2021.conll-1.15}
\showDOI{\tempurl}


\bibitem[\protect\citeauthoryear{Ravi, Yu, Santelices, Karray, and Fidan}{Ravi
  et~al\mbox{.}}{2021}]%
        {ravi-cfe-anomaly-detection}
\bibfield{author}{\bibinfo{person}{Ambareesh Ravi}, \bibinfo{person}{Xiaozhuo
  Yu}, \bibinfo{person}{Iara Santelices}, \bibinfo{person}{Fakhri Karray},
  {and} \bibinfo{person}{Baris Fidan}.} \bibinfo{year}{2021}\natexlab{}.
\newblock \showarticletitle{General Frameworks for Anomaly Detection
  Explainability: Comparative Study}. In \bibinfo{booktitle}{\emph{2021 IEEE
  International Conference on Autonomous Systems (ICAS)}}.
  \bibinfo{pages}{1--5}.
\newblock
\urldef\tempurl%
\url{https://doi.org/10.1109/ICAS49788.2021.9551129}
\showDOI{\tempurl}


\bibitem[\protect\citeauthoryear{Rawal, Kamar, and Lakkaraju}{Rawal
  et~al\mbox{.}}{2021}]%
        {rawal2021robust1}
\bibfield{author}{\bibinfo{person}{Kaivalya Rawal}, \bibinfo{person}{Ece
  Kamar}, {and} \bibinfo{person}{Himabindu Lakkaraju}.}
  \bibinfo{year}{2021}\natexlab{}.
\newblock \bibinfo{title}{Algorithmic Recourse in the Wild: Understanding the
  Impact of Data and Model Shifts}.
\newblock
\newblock
\showeprint[arxiv]{cs.LG/2012.11788}


\bibitem[\protect\citeauthoryear{Rawal and Lakkaraju}{Rawal and
  Lakkaraju}{2020}]%
        {hima-beyond-recourse-globalcfe}
\bibfield{author}{\bibinfo{person}{Kaivalya Rawal} {and}
  \bibinfo{person}{Himabindu Lakkaraju}.} \bibinfo{year}{2020}\natexlab{}.
\newblock \showarticletitle{Beyond Individualized Recourse: Interpretable and
  Interactive Summaries of Actionable Recourses}. In
  \bibinfo{booktitle}{\emph{Advances in Neural Information Processing
  Systems}}, Vol.~\bibinfo{volume}{33}. \bibinfo{publisher}{Curran Associates,
  Inc.}, \bibinfo{pages}{12187--12198}.
\newblock
\urldef\tempurl%
\url{https://proceedings.neurips.cc/paper/2020/file/8ee7730e97c67473a424ccfeff49ab20-Paper.pdf}
\showURL{%
\tempurl}


\bibitem[\protect\citeauthoryear{Redelmeier, Jullum, Aas, and
  Løland}{Redelmeier et~al\mbox{.}}{2021}]%
        {monte-carlo-cfe-technique}
\bibfield{author}{\bibinfo{person}{Annabelle Redelmeier},
  \bibinfo{person}{Martin Jullum}, \bibinfo{person}{Kjersti Aas}, {and}
  \bibinfo{person}{Anders Løland}.} \bibinfo{year}{2021}\natexlab{}.
\newblock \bibinfo{title}{MCCE: Monte Carlo sampling of realistic
  counterfactual explanations}.
\newblock
\newblock
\urldef\tempurl%
\url{https://doi.org/10.48550/ARXIV.2111.09790}
\showDOI{\tempurl}


\bibitem[\protect\citeauthoryear{Reed, Grieman, and Early}{Reed
  et~al\mbox{.}}{2021}]%
        {non-asimov-ai-regulation-paper}
\bibfield{author}{\bibinfo{person}{Chris Reed}, \bibinfo{person}{Keri Grieman},
  {and} \bibinfo{person}{Joseph Early}.} \bibinfo{year}{2021}\natexlab{}.
\newblock \showarticletitle{Non-Asimov Explanations Regulating AI Through
  Transparency}. In \bibinfo{booktitle}{\emph{Queen Mary Law Research Paper No.
  370/2021}}.
\newblock
\urldef\tempurl%
\url{https://ssrn.com/abstract=3970518}
\showURL{%
\tempurl}


\bibitem[\protect\citeauthoryear{Ribeiro, Singh, and Guestrin}{Ribeiro
  et~al\mbox{.}}{2016}]%
        {ribeiro_why_2016}
\bibfield{author}{\bibinfo{person}{Marco~Tulio Ribeiro},
  \bibinfo{person}{Sameer Singh}, {and} \bibinfo{person}{Carlos Guestrin}.}
  \bibinfo{year}{2016}\natexlab{}.
\newblock \showarticletitle{"Why Should I Trust You?": Explaining the
  Predictions of Any Classifier}. In \bibinfo{booktitle}{\emph{Proceedings of
  the 22nd ACM SIGKDD International Conference on Knowledge Discovery and Data
  Mining}} \emph{(\bibinfo{series}{KDD '16})}. \bibinfo{publisher}{Association
  for Computing Machinery}, \bibinfo{address}{New York, NY, USA}, 10.
\newblock
\urldef\tempurl%
\url{https://doi.org/10.1145/2939672.2939778}
\showDOI{\tempurl}


\bibitem[\protect\citeauthoryear{Ribeiro, Singh, and Guestrin}{Ribeiro
  et~al\mbox{.}}{2018}]%
        {Ribeiro2018Anchors}
\bibfield{author}{\bibinfo{person}{Marco~Tulio Ribeiro},
  \bibinfo{person}{Sameer Singh}, {and} \bibinfo{person}{Carlos Guestrin}.}
  \bibinfo{year}{2018}\natexlab{}.
\newblock \showarticletitle{Anchors: High-Precision Model-Agnostic
  Explanations}. In \bibinfo{booktitle}{\emph{Conference on Artificial
  Intelligence (AAAI)}}. \bibinfo{publisher}{AAAI press},
  \bibinfo{address}{California, USA}, 9.
\newblock
\urldef\tempurl%
\url{https://www.aaai.org/ocs/index.php/AAAI/AAAI18/paper/view/16982}
\showURL{%
\tempurl}


\bibitem[\protect\citeauthoryear{Robeer, Bex, and Feelders}{Robeer
  et~al\mbox{.}}{2021}]%
        {robeer-cfe-realistic-text}
\bibfield{author}{\bibinfo{person}{Marcel Robeer}, \bibinfo{person}{Floris
  Bex}, {and} \bibinfo{person}{Ad Feelders}.} \bibinfo{year}{2021}\natexlab{}.
\newblock \showarticletitle{Generating Realistic Natural Language
  Counterfactuals}. In \bibinfo{booktitle}{\emph{Findings of the Association
  for Computational Linguistics: EMNLP 2021}}. \bibinfo{publisher}{Association
  for Computational Linguistics}, \bibinfo{address}{Punta Cana, Dominican
  Republic}, \bibinfo{pages}{3611--3625}.
\newblock
\urldef\tempurl%
\url{https://doi.org/10.18653/v1/2021.findings-emnlp.306}
\showDOI{\tempurl}


\bibitem[\protect\citeauthoryear{Rodriguez, Caccia, Lacoste, Zamparo, Laradji,
  Charlin, and Vazquez}{Rodriguez et~al\mbox{.}}{2021}]%
        {beyond_trivial_cfe_images}
\bibfield{author}{\bibinfo{person}{Pau Rodriguez}, \bibinfo{person}{Massimo
  Caccia}, \bibinfo{person}{Alexandre Lacoste}, \bibinfo{person}{Lee Zamparo},
  \bibinfo{person}{Issam Laradji}, \bibinfo{person}{Laurent Charlin}, {and}
  \bibinfo{person}{David Vazquez}.} \bibinfo{year}{2021}\natexlab{}.
\newblock \bibinfo{title}{Beyond Trivial Counterfactual Explanations with
  Diverse Valuable Explanations}.
\newblock
\newblock
\urldef\tempurl%
\url{https://doi.org/10.48550/ARXIV.2103.10226}
\showDOI{\tempurl}


\bibitem[\protect\citeauthoryear{Ross, Lakkaraju, and Bastani}{Ross
  et~al\mbox{.}}{2021}]%
        {alexis-ross-training-methodology-cfe}
\bibfield{author}{\bibinfo{person}{Alexis Ross}, \bibinfo{person}{Himabindu
  Lakkaraju}, {and} \bibinfo{person}{Osbert Bastani}.}
  \bibinfo{year}{2021}\natexlab{}.
\newblock \showarticletitle{Learning Models for Actionable Recourse}. In
  \bibinfo{booktitle}{\emph{Advances in Neural Information Processing
  Systems}}, Vol.~\bibinfo{volume}{34}. \bibinfo{publisher}{Curran Associates,
  Inc.}, \bibinfo{pages}{18734--18746}.
\newblock
\urldef\tempurl%
\url{https://proceedings.neurips.cc/paper/2021/file/9b82909c30456ac902e14526e63081d4-Paper.pdf}
\showURL{%
\tempurl}


\bibitem[\protect\citeauthoryear{Ruben}{Ruben}{1992}]%
        {Ruben2004:phil3}
\bibfield{author}{\bibinfo{person}{David-Hillel Ruben}.}
  \bibinfo{year}{1992}\natexlab{}.
\newblock \bibinfo{booktitle}{\emph{Counterfactuals}}.
\newblock \bibinfo{publisher}{Routledge Publishers}.
\newblock
\urldef\tempurl%
\url{https://philarchive.org/archive/RUBEE-3}
\showURL{%
\tempurl}


\bibitem[\protect\citeauthoryear{Russell}{Russell}{2019}]%
        {russell_efficient_2019}
\bibfield{author}{\bibinfo{person}{Chris Russell}.}
  \bibinfo{year}{2019}\natexlab{}.
\newblock \showarticletitle{Efficient Search for Diverse Coherent
  Explanations}. In \bibinfo{booktitle}{\emph{Proceedings of the Conference on
  Fairness, Accountability, and Transparency (FAccT)}}
  \emph{(\bibinfo{series}{FAT* '19})}. \bibinfo{publisher}{Association for
  Computing Machinery}, \bibinfo{address}{New York, NY, USA},
  \bibinfo{pages}{20–28}.
\newblock
\showISBNx{9781450361255}
\urldef\tempurl%
\url{https://doi.org/10.1145/3287560.3287569}
\showDOI{\tempurl}


\bibitem[\protect\citeauthoryear{Sadler, Greene, and Archambault}{Sadler
  et~al\mbox{.}}{2021}]%
        {Sadler-cfe-community-detection}
\bibfield{author}{\bibinfo{person}{Sophie Sadler}, \bibinfo{person}{Derek
  Greene}, {and} \bibinfo{person}{Daniel~W. Archambault}.}
  \bibinfo{year}{2021}\natexlab{}.
\newblock \showarticletitle{A Study of Explainable Community-Level Features}.
  In \bibinfo{booktitle}{\emph{GEM: Graph Embedding and Mining ECML-PKDD 2021
  Workshop+Tutorial}}.
\newblock


\bibitem[\protect\citeauthoryear{Sajja, Mukherjee, Dwivedi, and Raykar}{Sajja
  et~al\mbox{.}}{2021}]%
        {semi-supervised-autoencoder-cfe}
\bibfield{author}{\bibinfo{person}{Surya Shravan~Kumar Sajja},
  \bibinfo{person}{Sumanta Mukherjee}, \bibinfo{person}{Satyam Dwivedi}, {and}
  \bibinfo{person}{Vikas~C. Raykar}.} \bibinfo{year}{2021}\natexlab{}.
\newblock \bibinfo{title}{Semi-supervised counterfactual explanations}.
\newblock
\newblock
\urldef\tempurl%
\url{https://openreview.net/forum?id=o6ndFLB1DST}
\showURL{%
\tempurl}


\bibitem[\protect\citeauthoryear{Samoilescu, Van~Looveren, and
  Klaise}{Samoilescu et~al\mbox{.}}{2021}]%
        {rl_cfe_approach2_amortized}
\bibfield{author}{\bibinfo{person}{Robert-Florian Samoilescu},
  \bibinfo{person}{Arnaud Van~Looveren}, {and} \bibinfo{person}{Janis Klaise}.}
  \bibinfo{year}{2021}\natexlab{}.
\newblock \bibinfo{title}{Model-agnostic and Scalable Counterfactual
  Explanations via Reinforcement Learning}.
\newblock
\newblock
\urldef\tempurl%
\url{https://doi.org/10.48550/ARXIV.2106.02597}
\showDOI{\tempurl}


\bibitem[\protect\citeauthoryear{Sanchez and Tsaftaris}{Sanchez and
  Tsaftaris}{2022}]%
        {sanchez-diffusion-causal-cfe-images}
\bibfield{author}{\bibinfo{person}{Pedro Sanchez} {and}
  \bibinfo{person}{Sotirios~A. Tsaftaris}.} \bibinfo{year}{2022}\natexlab{}.
\newblock \bibinfo{title}{Diffusion Causal Models for Counterfactual
  Estimation}.
\newblock
\newblock
\urldef\tempurl%
\url{https://doi.org/10.48550/ARXIV.2202.10166}
\showDOI{\tempurl}


\bibitem[\protect\citeauthoryear{Schleich, Geng, Zhang, and Suciu}{Schleich
  et~al\mbox{.}}{2021}]%
        {schleich2021geco}
\bibfield{author}{\bibinfo{person}{Maximilian Schleich},
  \bibinfo{person}{Zixuan Geng}, \bibinfo{person}{Yihong Zhang}, {and}
  \bibinfo{person}{Dan Suciu}.} \bibinfo{year}{2021}\natexlab{}.
\newblock \bibinfo{title}{GeCo: Quality Counterfactual Explanations in Real
  Time}.
\newblock
\newblock
\showeprint[arxiv]{cs.LG/2101.01292}


\bibitem[\protect\citeauthoryear{Schut, Key, McGrath, Costabello, Sacaleanu,
  Corcoran, and Gal}{Schut et~al\mbox{.}}{2021}]%
        {Epistemic_and_Aleatoric_uncertainty}
\bibfield{author}{\bibinfo{person}{Lisa Schut}, \bibinfo{person}{Oscar Key},
  \bibinfo{person}{Rory McGrath}, \bibinfo{person}{Luca Costabello},
  \bibinfo{person}{Bogdan Sacaleanu}, \bibinfo{person}{Medb Corcoran}, {and}
  \bibinfo{person}{Yarin Gal}.} \bibinfo{year}{2021}\natexlab{}.
\newblock \bibinfo{title}{Generating Interpretable Counterfactual Explanations
  By Implicit Minimisation of Epistemic and Aleatoric Uncertainties}.
\newblock
\newblock
\urldef\tempurl%
\url{https://doi.org/10.48550/ARXIV.2103.08951}
\showDOI{\tempurl}


\bibitem[\protect\citeauthoryear{{Selvaraju}, {Cogswell}, {Das}, {Vedantam},
  {Parikh}, and {Batra}}{{Selvaraju} et~al\mbox{.}}{2017}]%
        {grad-cam}
\bibfield{author}{\bibinfo{person}{R.~R. {Selvaraju}}, \bibinfo{person}{M.
  {Cogswell}}, \bibinfo{person}{A. {Das}}, \bibinfo{person}{R. {Vedantam}},
  \bibinfo{person}{D. {Parikh}}, {and} \bibinfo{person}{D. {Batra}}.}
  \bibinfo{year}{2017}\natexlab{}.
\newblock \showarticletitle{Grad-CAM: Visual Explanations from Deep Networks
  via Gradient-Based Localization}. In \bibinfo{booktitle}{\emph{IEEE
  International Conference on Computer Vision}}. \bibinfo{pages}{618--626}.
\newblock


\bibitem[\protect\citeauthoryear{Sennaar}{Sennaar}{2019}]%
        {hiring-ml}
\bibfield{author}{\bibinfo{person}{Kumba Sennaar}.}
  \bibinfo{year}{2019}\natexlab{}.
\newblock \bibinfo{title}{Machine Learning for Recruiting and Hiring – 6
  Current Applications}.
\newblock
  \bibinfo{howpublished}{\url{https://emerj.com/ai-sector-overviews/machine-learning-for-recruiting-and-hiring/}}.
\newblock
\newblock
\shownote{Accessed: 2020-10-15.}


\bibitem[\protect\citeauthoryear{Shang, Feng, and Shah}{Shang
  et~al\mbox{.}}{2022}]%
        {anna-cfe-recos}
\bibfield{author}{\bibinfo{person}{Ruoxi Shang}, \bibinfo{person}{K.~J.~Kevin
  Feng}, {and} \bibinfo{person}{Chirag Shah}.} \bibinfo{year}{2022}\natexlab{}.
\newblock \showarticletitle{Why Am I Not Seeing It? Understanding Users’
  Needs for Counterfactual Explanations in Everyday Recommendations}. In
  \bibinfo{booktitle}{\emph{2022 ACM Conference on Fairness, Accountability,
  and Transparency}} \emph{(\bibinfo{series}{FAccT '22})}.
  \bibinfo{publisher}{Association for Computing Machinery},
  \bibinfo{address}{New York, NY, USA}, 11.
\newblock
\urldef\tempurl%
\url{https://doi.org/10.1145/3531146.3533189}
\showDOI{\tempurl}


\bibitem[\protect\citeauthoryear{Shao and Kersting}{Shao and Kersting}{2022}]%
        {sum-product-networks-cfe}
\bibfield{author}{\bibinfo{person}{Xiaoting Shao} {and}
  \bibinfo{person}{Kristian Kersting}.} \bibinfo{year}{2022}\natexlab{}.
\newblock \bibinfo{title}{Gradient-based Counterfactual Explanations using
  Tractable Probabilistic Models}.
\newblock
\newblock
\urldef\tempurl%
\url{https://doi.org/10.48550/ARXIV.2205.07774}
\showDOI{\tempurl}


\bibitem[\protect\citeauthoryear{Sharma, Henderson, and Ghosh}{Sharma
  et~al\mbox{.}}{2019}]%
        {sharma_certifai_2019}
\bibfield{author}{\bibinfo{person}{Shubham Sharma}, \bibinfo{person}{Jette
  Henderson}, {and} \bibinfo{person}{Joydeep Ghosh}.}
  \bibinfo{year}{2019}\natexlab{}.
\newblock \bibinfo{title}{{CERTIFAI}: {Counterfactual} {Explanations} for
  {Robustness}, {Transparency}, {Interpretability}, and {Fairness} of
  {Artificial} {Intelligence} models}.
\newblock
\newblock
\urldef\tempurl%
\url{http://arxiv.org/abs/1905.07857}
\showURL{%
\tempurl}


\bibitem[\protect\citeauthoryear{Shokri, Strobel, and Zick}{Shokri
  et~al\mbox{.}}{2021}]%
        {modelprivacy-generalxai}
\bibfield{author}{\bibinfo{person}{Reza Shokri}, \bibinfo{person}{Martin
  Strobel}, {and} \bibinfo{person}{Yair Zick}.}
  \bibinfo{year}{2021}\natexlab{}.
\newblock \showarticletitle{On the Privacy Risks of Model Explanations}. In
  \bibinfo{booktitle}{\emph{Proceedings of the 2021 AAAI/ACM Conference on AI,
  Ethics, and Society}}. \bibinfo{publisher}{Association for Computing
  Machinery}, \bibinfo{address}{New York, NY, USA}, 11.
\newblock
\urldef\tempurl%
\url{https://doi.org/10.1145/3461702.3462533}
\showDOI{\tempurl}


\bibitem[\protect\citeauthoryear{Singh, Dourish, Howe, Miller, Sonenberg,
  Velloso, and Vetere}{Singh et~al\mbox{.}}{2021}]%
        {sequential-CFE}
\bibfield{author}{\bibinfo{person}{Ronal~Rajneshwar Singh},
  \bibinfo{person}{Paul Dourish}, \bibinfo{person}{Piers Howe},
  \bibinfo{person}{Tim Miller}, \bibinfo{person}{Liz Sonenberg},
  \bibinfo{person}{Eduardo Velloso}, {and} \bibinfo{person}{Frank Vetere}.}
  \bibinfo{year}{2021}\natexlab{}.
\newblock \bibinfo{title}{Directive Explanations for Actionable Explainability
  in Machine Learning Applications}.
\newblock
\newblock


\bibitem[\protect\citeauthoryear{Singla}{Singla}{2020}]%
        {credit-risk-ml}
\bibfield{author}{\bibinfo{person}{Saurav Singla}.}
  \bibinfo{year}{2020}\natexlab{}.
\newblock \bibinfo{title}{Machine Learning to Predict Credit Risk in Lending
  Industry}.
\newblock
  \bibinfo{howpublished}{\url{https://www.aitimejournal.com/@saurav.singla/machine-learning-to-predict-credit-risk-in-lending-industry}}.
\newblock
\newblock
\shownote{Accessed: 2020-10-15.}


\bibitem[\protect\citeauthoryear{Slack, Hilgard, Lakkaraju, and Singh}{Slack
  et~al\mbox{.}}{2021}]%
        {slack2021manipulated}
\bibfield{author}{\bibinfo{person}{Dylan Slack}, \bibinfo{person}{Sophie
  Hilgard}, \bibinfo{person}{Himabindu Lakkaraju}, {and}
  \bibinfo{person}{Sameer Singh}.} \bibinfo{year}{2021}\natexlab{}.
\newblock \bibinfo{title}{Counterfactual Explanations Can Be Manipulated}.
\newblock
\newblock
\showeprint[arxiv]{cs.LG/2106.02666}


\bibitem[\protect\citeauthoryear{Smith, Everhart, Dickson, Knowler, and
  Johannes}{Smith et~al\mbox{.}}{1988}]%
        {pima-diabetes-data}
\bibfield{author}{\bibinfo{person}{J.~W. Smith}, \bibinfo{person}{J. Everhart},
  \bibinfo{person}{W.~C. Dickson}, \bibinfo{person}{W. Knowler}, {and}
  \bibinfo{person}{R. Johannes}.} \bibinfo{year}{1988}\natexlab{}.
\newblock \showarticletitle{Using the ADAP Learning Algorithm to Forecast the
  Onset of Diabetes Mellitus}. In \bibinfo{booktitle}{\emph{Proceedings of the
  Annual Symposium on Computer Application in Medical Care}}.
  \bibinfo{publisher}{American Medical Informatics Association},
  \bibinfo{address}{Washington,D.C.}, \bibinfo{pages}{261–265}.
\newblock


\bibitem[\protect\citeauthoryear{Smith and Ramamoorthy}{Smith and
  Ramamoorthy}{2020}]%
        {Smith-CFE-images-robot-action}
\bibfield{author}{\bibinfo{person}{Simón~C. Smith} {and}
  \bibinfo{person}{Subramanian Ramamoorthy}.} \bibinfo{year}{2020}\natexlab{}.
\newblock \showarticletitle{Counterfactual Explanation and Causal Inference In
  Service of Robustness in Robot Control}. In \bibinfo{booktitle}{\emph{2020
  Joint IEEE 10th International Conference on Development and Learning and
  Epigenetic Robotics (ICDL-EpiRob)}}. \bibinfo{pages}{1--8}.
\newblock
\urldef\tempurl%
\url{https://doi.org/10.1109/ICDL-EpiRob48136.2020.9278061}
\showDOI{\tempurl}


\bibitem[\protect\citeauthoryear{Sokol and Flach}{Sokol and Flach}{2018}]%
        {glass-box-voice-assistant}
\bibfield{author}{\bibinfo{person}{Kacper Sokol} {and} \bibinfo{person}{Peter
  Flach}.} \bibinfo{year}{2018}\natexlab{}.
\newblock \showarticletitle{Glass-Box: Explaining AI Decisions with
  Counterfactual Statements through Conversation with a Voice-Enabled Virtual
  Assistant}. In \bibinfo{booktitle}{\emph{Proceedings of the 27th
  International Joint Conference on Artificial Intelligence}}
  \emph{(\bibinfo{series}{IJCAI'18})}. \bibinfo{publisher}{AAAI Press},
  \bibinfo{pages}{5868–5870}.
\newblock


\bibitem[\protect\citeauthoryear{Sokol and Flach}{Sokol and Flach}{2019}]%
        {sokol_desiderata_2019}
\bibfield{author}{\bibinfo{person}{Kacper Sokol} {and} \bibinfo{person}{Peter
  Flach}.} \bibinfo{year}{2019}\natexlab{}.
\newblock \showarticletitle{Desiderata for {Interpretability}: {Explaining}
  {Decision} {Tree} {Predictions} with {Counterfactuals}}.
\newblock \bibinfo{journal}{\emph{Conference on Artificial Intelligence
  (AAAI)}}  \bibinfo{volume}{33} (\bibinfo{date}{July} \bibinfo{year}{2019}).
\newblock
\urldef\tempurl%
\url{https://doi.org/10.1609/aaai.v33i01.330110035}
\showDOI{\tempurl}


\bibitem[\protect\citeauthoryear{Spooner, Dervovic, Long, Shepard, Chen, and
  Magazzeni}{Spooner et~al\mbox{.}}{2021}]%
        {cf-regression}
\bibfield{author}{\bibinfo{person}{Thomas Spooner}, \bibinfo{person}{Danial
  Dervovic}, \bibinfo{person}{Jason Long}, \bibinfo{person}{Jon Shepard},
  \bibinfo{person}{Jiahao Chen}, {and} \bibinfo{person}{Daniele Magazzeni}.}
  \bibinfo{year}{2021}\natexlab{}.
\newblock \bibinfo{title}{Counterfactual Explanations for Arbitrary Regression
  Models}.
\newblock
\newblock


\bibitem[\protect\citeauthoryear{State}{State}{2021}]%
        {cfe-logic-programming-laurastate}
\bibfield{author}{\bibinfo{person}{Laura State}.}
  \bibinfo{year}{2021}\natexlab{}.
\newblock \showarticletitle{Logic Programming for {XAI:} {A} Technical
  Perspective}. In \bibinfo{booktitle}{\emph{Proceedings of the International
  Conference on Logic Programming 2021 Workshops (ICLP 2021)}},
  Vol.~\bibinfo{volume}{2970}.
\newblock
\urldef\tempurl%
\url{http://ceur-ws.org/Vol-2970/meepaper1.pdf}
\showURL{%
\tempurl}


\bibitem[\protect\citeauthoryear{Stein}{Stein}{2021}]%
        {stein-cfe-agent-navigation}
\bibfield{author}{\bibinfo{person}{Gregory Stein}.}
  \bibinfo{year}{2021}\natexlab{}.
\newblock \showarticletitle{Generating High-Quality Explanations for Navigation
  in Partially-Revealed Environments}. In \bibinfo{booktitle}{\emph{Advances in
  Neural Information Processing Systems}}, Vol.~\bibinfo{volume}{34}.
  \bibinfo{publisher}{Curran Associates, Inc.}, \bibinfo{pages}{17493--17506}.
\newblock
\urldef\tempurl%
\url{https://proceedings.neurips.cc/paper/2021/file/926ec030f29f83ce5318754fdb631a33-Paper.pdf}
\showURL{%
\tempurl}


\bibitem[\protect\citeauthoryear{Sulem, Donini, Zafar, Aubet, Gasthaus,
  Januschowski, Das, Kenthapadi, and Archambeau}{Sulem et~al\mbox{.}}{2022}]%
        {sulem-cfe-time-series-anomaly-detection}
\bibfield{author}{\bibinfo{person}{Deborah Sulem}, \bibinfo{person}{Michele
  Donini}, \bibinfo{person}{Muhammad~Bilal Zafar},
  \bibinfo{person}{Francois-Xavier Aubet}, \bibinfo{person}{Jan Gasthaus},
  \bibinfo{person}{Tim Januschowski}, \bibinfo{person}{Sanjiv Das},
  \bibinfo{person}{Krishnaram Kenthapadi}, {and} \bibinfo{person}{Cedric
  Archambeau}.} \bibinfo{year}{2022}\natexlab{}.
\newblock \bibinfo{title}{Diverse Counterfactual Explanations for Anomaly
  Detection in Time Series}.
\newblock
\newblock
\urldef\tempurl%
\url{https://doi.org/10.48550/ARXIV.2203.11103}
\showDOI{\tempurl}


\bibitem[\protect\citeauthoryear{Tahoun and Kassis}{Tahoun and Kassis}{2020}]%
        {action-cfe-tahoun}
\bibfield{author}{\bibinfo{person}{Ezzeldin Tahoun} {and}
  \bibinfo{person}{Andre Kassis}.} \bibinfo{year}{2020}\natexlab{}.
\newblock \bibinfo{title}{Beyond Explanations: Recourse via Actionable
  Interpretability - Extended}.
\newblock
\newblock
\urldef\tempurl%
\url{https://doi.org/10.13140/RG.2.2.19076.14729}
\showDOI{\tempurl}


\bibitem[\protect\citeauthoryear{Tamagnini, Krause, Dasgupta, and
  Bertini}{Tamagnini et~al\mbox{.}}{2017}]%
        {Rivelo-visualcfe}
\bibfield{author}{\bibinfo{person}{Paolo Tamagnini}, \bibinfo{person}{Josua
  Krause}, \bibinfo{person}{Aritra Dasgupta}, {and} \bibinfo{person}{Enrico
  Bertini}.} \bibinfo{year}{2017}\natexlab{}.
\newblock \showarticletitle{Interpreting Black-Box Classifiers Using
  Instance-Level Visual Explanations}. In \bibinfo{booktitle}{\emph{Proceedings
  of the 2nd Workshop on Human-In-the-Loop Data Analytics}}.
  \bibinfo{publisher}{Association for Computing Machinery},
  \bibinfo{address}{New York, NY, USA}, 6.
\newblock
\urldef\tempurl%
\url{https://doi.org/10.1145/3077257.3077260}
\showDOI{\tempurl}


\bibitem[\protect\citeauthoryear{Tan, Xu, Ge, Li, Chen, and Zhang}{Tan
  et~al\mbox{.}}{2021}]%
        {cfe-reco-approach4}
\bibfield{author}{\bibinfo{person}{Juntao Tan}, \bibinfo{person}{Shuyuan Xu},
  \bibinfo{person}{Yingqiang Ge}, \bibinfo{person}{Yunqi Li},
  \bibinfo{person}{Xu Chen}, {and} \bibinfo{person}{Yongfeng Zhang}.}
  \bibinfo{year}{2021}\natexlab{}.
\newblock \showarticletitle{Counterfactual Explainable Recommendation}. In
  \bibinfo{booktitle}{\emph{Proceedings of the 30th ACM International
  Conference on Information \& Knowledge Management}}.
  \bibinfo{publisher}{Association for Computing Machinery},
  \bibinfo{address}{New York, NY, USA}, 10.
\newblock


\bibitem[\protect\citeauthoryear{Tan, Caruana, Hooker, and Lou}{Tan
  et~al\mbox{.}}{2018}]%
        {lendingclub-data}
\bibfield{author}{\bibinfo{person}{Sarah Tan}, \bibinfo{person}{Rich Caruana},
  \bibinfo{person}{Giles Hooker}, {and} \bibinfo{person}{Yin Lou}.}
  \bibinfo{year}{2018}\natexlab{}.
\newblock \showarticletitle{Distill-and-Compare: Auditing Black-Box Models
  Using Transparent Model Distillation}. In
  \bibinfo{booktitle}{\emph{Proceedings of the 2018 AAAI/ACM Conference on AI,
  Ethics, and Society}} \emph{(\bibinfo{series}{AIES '18})}.
  \bibinfo{publisher}{Association for Computing Machinery},
  \bibinfo{address}{New York, NY, USA}, 8.
\newblock
\urldef\tempurl%
\url{https://doi.org/10.1145/3278721.3278725}
\showDOI{\tempurl}


\bibitem[\protect\citeauthoryear{Tashea}{Tashea}{2017}]%
        {parole-ml}
\bibfield{author}{\bibinfo{person}{Jason Tashea}.}
  \bibinfo{year}{2017}\natexlab{}.
\newblock \bibinfo{title}{Courts Are Using AI to Sentence Criminals. That Must
  Stop Now}.
\newblock
  \bibinfo{howpublished}{\url{https://www.wired.com/2017/04/courts-using-ai-sentence-criminals-must-stop-now/}}.
\newblock
\newblock
\shownote{Accessed: 2020-10-15.}


\bibitem[\protect\citeauthoryear{Temraz and Keane}{Temraz and Keane}{2021}]%
        {CFE-classimbalance}
\bibfield{author}{\bibinfo{person}{Mohammed Temraz} {and}
  \bibinfo{person}{Mark~T. Keane}.} \bibinfo{year}{2021}\natexlab{}.
\newblock \bibinfo{title}{Solving the Class Imbalance Problem Using a
  Counterfactual Method for Data Augmentation}.
\newblock
\newblock
\urldef\tempurl%
\url{https://doi.org/10.48550/ARXIV.2111.03516}
\showDOI{\tempurl}


\bibitem[\protect\citeauthoryear{Temraz, Kenny, Ruelle, Shalloo, Smyth, and
  Keane}{Temraz et~al\mbox{.}}{2021}]%
        {temraz-cfe-data-augmentation-climate-change}
\bibfield{author}{\bibinfo{person}{Mohammed Temraz}, \bibinfo{person}{Eoin~M.
  Kenny}, \bibinfo{person}{Elodie Ruelle}, \bibinfo{person}{Laurence Shalloo},
  \bibinfo{person}{Barry Smyth}, {and} \bibinfo{person}{Mark~T. Keane}.}
  \bibinfo{year}{2021}\natexlab{}.
\newblock \showarticletitle{Handling Climate Change Using Counterfactuals:
  Using Counterfactuals in Data Augmentation to Predict Crop Growth in an
  Uncertain Climate Future}. In \bibinfo{booktitle}{\emph{Case-Based Reasoning
  Research and Development}}. \bibinfo{publisher}{Springer International
  Publishing}, \bibinfo{address}{Cham}, \bibinfo{pages}{216--231}.
\newblock


\bibitem[\protect\citeauthoryear{Teofili, Firmani, Koudas, Martello, Merialdo,
  and Srivastava}{Teofili et~al\mbox{.}}{2022}]%
        {cfe-for-entity-resolution}
\bibfield{author}{\bibinfo{person}{T. Teofili}, \bibinfo{person}{D. Firmani},
  \bibinfo{person}{N. Koudas}, \bibinfo{person}{V. Martello},
  \bibinfo{person}{P. Merialdo}, {and} \bibinfo{person}{D. Srivastava}.}
  \bibinfo{year}{2022}\natexlab{}.
\newblock \showarticletitle{Effective Explanations for Entity Resolution
  Models}. In \bibinfo{booktitle}{\emph{2022 IEEE 38th International Conference
  on Data Engineering (ICDE)}}. \bibinfo{publisher}{IEEE Computer Society},
  \bibinfo{address}{Los Alamitos, CA, USA}, \bibinfo{pages}{2709--2721}.
\newblock
\urldef\tempurl%
\url{https://doi.org/10.1109/ICDE53745.2022.00248}
\showDOI{\tempurl}


\bibitem[\protect\citeauthoryear{Thiagarajan, Narayanaswamy, Rajan, Liang,
  Chaudhari, and Spanias}{Thiagarajan et~al\mbox{.}}{2021}]%
        {cfe_images_onfly_scope_disc}
\bibfield{author}{\bibinfo{person}{Jayaraman Thiagarajan},
  \bibinfo{person}{Vivek~Sivaraman Narayanaswamy}, \bibinfo{person}{Deepta
  Rajan}, \bibinfo{person}{Jia Liang}, \bibinfo{person}{Akshay Chaudhari},
  {and} \bibinfo{person}{Andreas Spanias}.} \bibinfo{year}{2021}\natexlab{}.
\newblock \showarticletitle{Designing Counterfactual Generators using Deep
  Model Inversion}. In \bibinfo{booktitle}{\emph{Advances in Neural Information
  Processing Systems}}, Vol.~\bibinfo{volume}{34}. \bibinfo{publisher}{Curran
  Associates, Inc.}, \bibinfo{pages}{16873--16884}.
\newblock
\urldef\tempurl%
\url{https://proceedings.neurips.cc/paper/2021/file/8ca01ea920679a0fe3728441494041b9-Paper.pdf}
\showURL{%
\tempurl}


\bibitem[\protect\citeauthoryear{Tjoa and Guan}{Tjoa and Guan}{2019}]%
        {tjoa2019survey1}
\bibfield{author}{\bibinfo{person}{Erico Tjoa} {and} \bibinfo{person}{Cuntai
  Guan}.} \bibinfo{year}{2019}\natexlab{}.
\newblock \bibinfo{title}{A Survey on Explainable Artificial Intelligence
  (XAI): Towards Medical XAI}.
\newblock
\newblock
\showeprint[arxiv]{cs.LG/1907.07374}


\bibitem[\protect\citeauthoryear{Tolkachev, Mell, Zdancewic, and
  Bastani}{Tolkachev et~al\mbox{.}}{2022}]%
        {tolkachev-CFE-text}
\bibfield{author}{\bibinfo{person}{George Tolkachev}, \bibinfo{person}{Stephen
  Mell}, \bibinfo{person}{Stephan Zdancewic}, {and} \bibinfo{person}{Osbert
  Bastani}.} \bibinfo{year}{2022}\natexlab{}.
\newblock \showarticletitle{Counterfactual Explanations for Natural Language
  Interfaces}. In \bibinfo{booktitle}{\emph{Proceedings of the 60th Annual
  Meeting of the Association for Computational Linguistics (Volume 2: Short
  Papers)}}. \bibinfo{publisher}{Association for Computational Linguistics},
  \bibinfo{address}{Dublin, Ireland}, \bibinfo{pages}{113--118}.
\newblock
\urldef\tempurl%
\url{https://aclanthology.org/2022.acl-short.14}
\showURL{%
\tempurl}


\bibitem[\protect\citeauthoryear{Tolomei, Silvestri, Haines, and
  Lalmas}{Tolomei et~al\mbox{.}}{2017}]%
        {Tolomei2017:Interpretable}
\bibfield{author}{\bibinfo{person}{Gabriele Tolomei}, \bibinfo{person}{Fabrizio
  Silvestri}, \bibinfo{person}{Andrew Haines}, {and} \bibinfo{person}{Mounia
  Lalmas}.} \bibinfo{year}{2017}\natexlab{}.
\newblock \showarticletitle{Interpretable Predictions of Tree-Based Ensembles
  via Actionable Feature Tweaking}. In \bibinfo{booktitle}{\emph{International
  Conference on Knowledge Discovery and Data Mining (KDD)}}
  \emph{(\bibinfo{series}{KDD '17})}. \bibinfo{publisher}{Association for
  Computing Machinery}, \bibinfo{address}{New York, NY, USA}, 10.
\newblock
\urldef\tempurl%
\url{https://doi.org/10.1145/3097983.3098039}
\showDOI{\tempurl}


\bibitem[\protect\citeauthoryear{Tran, Ghazimatin, and Saha~Roy}{Tran
  et~al\mbox{.}}{2021}]%
        {cfe-reco-approach2}
\bibfield{author}{\bibinfo{person}{Khanh~Hiep Tran}, \bibinfo{person}{Azin
  Ghazimatin}, {and} \bibinfo{person}{Rishiraj Saha~Roy}.}
  \bibinfo{year}{2021}\natexlab{}.
\newblock \bibinfo{booktitle}{\emph{Counterfactual Explanations for Neural
  Recommenders}}.
\newblock \bibinfo{publisher}{Association for Computing Machinery},
  \bibinfo{address}{New York, NY, USA}, \bibinfo{pages}{1627--1631}.
\newblock
\showISBNx{9781450380379}
\urldef\tempurl%
\url{https://doi.org/10.1145/3404835.3463005}
\showURL{%
\tempurl}


\bibitem[\protect\citeauthoryear{Tsiakmaki and Ragos}{Tsiakmaki and
  Ragos}{2021}]%
        {student-performance-cfe}
\bibfield{author}{\bibinfo{person}{Maria Tsiakmaki} {and}
  \bibinfo{person}{Omiros Ragos}.} \bibinfo{year}{2021}\natexlab{}.
\newblock \showarticletitle{A Case Study of Interpretable Counterfactual
  Explanations for the Task of Predicting Student Academic Performance}. In
  \bibinfo{booktitle}{\emph{2021 25th International Conference on Circuits,
  Systems, Communications and Computers (CSCC)}}.
\newblock
\urldef\tempurl%
\url{https://doi.org/10.1109/CSCC53858.2021.00029}
\showDOI{\tempurl}


\bibitem[\protect\citeauthoryear{Tsirtsis, De, and Rodriguez}{Tsirtsis
  et~al\mbox{.}}{2021}]%
        {cfe-sequential-data-Tsirtsis}
\bibfield{author}{\bibinfo{person}{Stratis Tsirtsis}, \bibinfo{person}{Abir
  De}, {and} \bibinfo{person}{Manuel Rodriguez}.}
  \bibinfo{year}{2021}\natexlab{}.
\newblock \showarticletitle{Counterfactual Explanations in Sequential Decision
  Making Under Uncertainty}. In \bibinfo{booktitle}{\emph{Advances in Neural
  Information Processing Systems}}, Vol.~\bibinfo{volume}{34}.
  \bibinfo{publisher}{Curran Associates, Inc.}, \bibinfo{pages}{30127--30139}.
\newblock
\urldef\tempurl%
\url{https://proceedings.neurips.cc/paper/2021/file/fd0a5a5e367a0955d81278062ef37429-Paper.pdf}
\showURL{%
\tempurl}


\bibitem[\protect\citeauthoryear{Tsirtsis and Gomez-Rodriguez}{Tsirtsis and
  Gomez-Rodriguez}{2020}]%
        {Manuel:game-theory}
\bibfield{author}{\bibinfo{person}{Stratis Tsirtsis} {and}
  \bibinfo{person}{Manuel Gomez-Rodriguez}.} \bibinfo{year}{2020}\natexlab{}.
\newblock \bibinfo{title}{Decisions, Counterfactual Explanations and Strategic
  Behavior}.
\newblock
\newblock
\showeprint[arxiv]{cs.LG/2002.04333}


\bibitem[\protect\citeauthoryear{Turner}{Turner}{2016}]%
        {Turner2016_MES}
\bibfield{author}{\bibinfo{person}{Ryan Turner}.}
  \bibinfo{year}{2016}\natexlab{}.
\newblock \showarticletitle{A Model Explanation System: Latest Updates and
  Extensions}.
\newblock \bibinfo{journal}{\emph{ArXiv}}  \bibinfo{volume}{abs/1606.09517}
  (\bibinfo{year}{2016}).
\newblock


\bibitem[\protect\citeauthoryear{UNIVERSITY}{UNIVERSITY}{[n. d.]}]%
        {EU-XAIfund2}
\bibfield{author}{\bibinfo{person}{AALTO UNIVERSITY}.} \bibinfo{year}{[n.
  d.]}\natexlab{}.
\newblock \bibinfo{title}{The European Commission offers significant support to
  Europe's AI excellence}.
\newblock
  \bibinfo{howpublished}{\url{https://www.eurekalert.org/pub_releases/2020-03/au-tec031820.php}}.
\newblock
\newblock
\shownote{Accessed: 2020-10-15.}


\bibitem[\protect\citeauthoryear{Upadhyay, Joshi, and Lakkaraju}{Upadhyay
  et~al\mbox{.}}{2021}]%
        {upadhyay2021robust2}
\bibfield{author}{\bibinfo{person}{Sohini Upadhyay}, \bibinfo{person}{Shalmali
  Joshi}, {and} \bibinfo{person}{Himabindu Lakkaraju}.}
  \bibinfo{year}{2021}\natexlab{}.
\newblock \bibinfo{title}{Towards Robust and Reliable Algorithmic Recourse}.
\newblock
\newblock
\showeprint[arxiv]{cs.LG/2102.13620}


\bibitem[\protect\citeauthoryear{Ustun, Spangher, and Liu}{Ustun
  et~al\mbox{.}}{2019}]%
        {Ustun19:Actionable}
\bibfield{author}{\bibinfo{person}{Berk Ustun}, \bibinfo{person}{Alexander
  Spangher}, {and} \bibinfo{person}{Yang Liu}.}
  \bibinfo{year}{2019}\natexlab{}.
\newblock \showarticletitle{Actionable Recourse in Linear Classification}. In
  \bibinfo{booktitle}{\emph{Proceedings of the Conference on Fairness,
  Accountability, and Transparency (FAccT)}} \emph{(\bibinfo{series}{FAT*
  '19})}. \bibinfo{publisher}{Association for Computing Machinery},
  \bibinfo{address}{New York, NY, USA}, 10.
\newblock
\urldef\tempurl%
\url{https://doi.org/10.1145/3287560.3287566}
\showDOI{\tempurl}


\bibitem[\protect\citeauthoryear{Van~Looveren and Klaise}{Van~Looveren and
  Klaise}{2020}]%
        {van_looveren_interpretable_2020}
\bibfield{author}{\bibinfo{person}{Arnaud Van~Looveren} {and}
  \bibinfo{person}{Janis Klaise}.} \bibinfo{year}{2020}\natexlab{}.
\newblock \bibinfo{title}{Interpretable {Counterfactual} {Explanations}
  {Guided} by {Prototypes}}.
\newblock
\newblock
\urldef\tempurl%
\url{http://arxiv.org/abs/1907.02584}
\showURL{%
\tempurl}


\bibitem[\protect\citeauthoryear{Van~Looveren, Klaise, Vacanti, and
  Cobb}{Van~Looveren et~al\mbox{.}}{2021}]%
        {conditional_gan_cfe_looveren}
\bibfield{author}{\bibinfo{person}{Arnaud Van~Looveren}, \bibinfo{person}{Janis
  Klaise}, \bibinfo{person}{Giovanni Vacanti}, {and} \bibinfo{person}{Oliver
  Cobb}.} \bibinfo{year}{2021}\natexlab{}.
\newblock \bibinfo{title}{Conditional Generative Models for Counterfactual
  Explanations}.
\newblock
\newblock
\urldef\tempurl%
\url{https://doi.org/10.48550/ARXIV.2101.10123}
\showDOI{\tempurl}


\bibitem[\protect\citeauthoryear{Vandenhende, Mahajan, Radenovic, and
  Ghadiyaram}{Vandenhende et~al\mbox{.}}{2022}]%
        {vandenhende-cfe-visual-images}
\bibfield{author}{\bibinfo{person}{Simon Vandenhende}, \bibinfo{person}{Dhruv
  Mahajan}, \bibinfo{person}{Filip Radenovic}, {and} \bibinfo{person}{Deepti
  Ghadiyaram}.} \bibinfo{year}{2022}\natexlab{}.
\newblock \showarticletitle{Making Heads or Tails: Towards Semantically
  Consistent Visual Counterfactuals}. In \bibinfo{booktitle}{\emph{ECCV 2022}}.
\newblock


\bibitem[\protect\citeauthoryear{Verma, Dickerson, and Hines}{Verma
  et~al\mbox{.}}{2020}]%
        {first-version-cfesurvey}
\bibfield{author}{\bibinfo{person}{Sahil Verma}, \bibinfo{person}{John
  Dickerson}, {and} \bibinfo{person}{Keegan Hines}.}
  \bibinfo{year}{2020}\natexlab{}.
\newblock \bibinfo{title}{Counterfactual Explanations for Machine Learning: A
  Review}.
\newblock
\newblock
\urldef\tempurl%
\url{https://doi.org/10.48550/ARXIV.2010.10596}
\showDOI{\tempurl}


\bibitem[\protect\citeauthoryear{Verma, Dickerson, and Hines}{Verma
  et~al\mbox{.}}{2021a}]%
        {cfe-challenges-paper}
\bibfield{author}{\bibinfo{person}{Sahil Verma}, \bibinfo{person}{John
  Dickerson}, {and} \bibinfo{person}{Keegan Hines}.}
  \bibinfo{year}{2021}\natexlab{a}.
\newblock \bibinfo{title}{Counterfactual Explanations for Machine Learning:
  Challenges Revisited}.
\newblock
\newblock
\urldef\tempurl%
\url{https://doi.org/10.48550/ARXIV.2106.07756}
\showDOI{\tempurl}


\bibitem[\protect\citeauthoryear{Verma, Hines, and Dickerson}{Verma
  et~al\mbox{.}}{2021b}]%
        {verma2021amortized}
\bibfield{author}{\bibinfo{person}{Sahil Verma}, \bibinfo{person}{Keegan
  Hines}, {and} \bibinfo{person}{John~P. Dickerson}.}
  \bibinfo{year}{2021}\natexlab{b}.
\newblock \bibinfo{title}{Amortized Generation of Sequential Counterfactual
  Explanations for Black-box Models}.
\newblock
\newblock
\showeprint[arxiv]{cs.LG/2106.03962}


\bibitem[\protect\citeauthoryear{Verma and Rubin}{Verma and Rubin}{2018}]%
        {verma_fairness}
\bibfield{author}{\bibinfo{person}{Sahil Verma} {and} \bibinfo{person}{Julia
  Rubin}.} \bibinfo{year}{2018}\natexlab{}.
\newblock \showarticletitle{Fairness Definitions Explained}. In
  \bibinfo{booktitle}{\emph{Proceedings of the International Workshop on
  Software Fairness}} \emph{(\bibinfo{series}{FairWare ’18})}.
  \bibinfo{publisher}{Association for Computing Machinery},
  \bibinfo{address}{New York, NY, USA}, \bibinfo{pages}{1–7}.
\newblock
\urldef\tempurl%
\url{https://doi.org/10.1145/3194770.3194776}
\showDOI{\tempurl}


\bibitem[\protect\citeauthoryear{Vermeire, Brughmans, Goethals, de~Oliveira,
  and Martens}{Vermeire et~al\mbox{.}}{[n. d.]}]%
        {vermeire-explainable-images}
\bibfield{author}{\bibinfo{person}{Tom Vermeire}, \bibinfo{person}{Dieter
  Brughmans}, \bibinfo{person}{Sofie Goethals}, \bibinfo{person}{Raphael
  Mazzine~Barbossa de Oliveira}, {and} \bibinfo{person}{David Martens}.}
  \bibinfo{year}{[n. d.]}\natexlab{}.
\newblock \showarticletitle{Explainable Image Classification with Evidence
  Counterfactual}.
\newblock \bibinfo{journal}{\emph{Pattern Anal. Appl.}} \bibinfo{volume}{25},
  \bibinfo{number}{2} (\bibinfo{year}{[n. d.]}), 21.
\newblock
\urldef\tempurl%
\url{https://doi.org/10.1007/s10044-021-01055-y}
\showDOI{\tempurl}


\bibitem[\protect\citeauthoryear{VILLANI}{VILLANI}{[n. d.]}]%
        {Fr-XAI1}
\bibfield{author}{\bibinfo{person}{CÉDRIC VILLANI}.} \bibinfo{year}{[n.
  d.]}\natexlab{}.
\newblock \bibinfo{title}{FOR A MEANINGFUL ARTIFICIAL INTELLIGENCE}.
\newblock
  \bibinfo{howpublished}{\url{https://www.aiforhumanity.fr/pdfs/MissionVillani_Report_ENG-VF.pdf}}.
\newblock
\newblock
\shownote{Accessed: 2020-10-15.}


\bibitem[\protect\citeauthoryear{Virgolin and Fracaros}{Virgolin and
  Fracaros}{2022}]%
        {robustness-KC-score-is-good}
\bibfield{author}{\bibinfo{person}{Marco Virgolin} {and}
  \bibinfo{person}{Saverio Fracaros}.} \bibinfo{year}{2022}\natexlab{}.
\newblock \bibinfo{title}{On the Robustness of Sparse Counterfactual
  Explanations to Adverse Perturbations}.
\newblock
\newblock
\urldef\tempurl%
\url{https://doi.org/10.48550/ARXIV.2201.09051}
\showDOI{\tempurl}


\bibitem[\protect\citeauthoryear{von K{\"u}gelgen, Agarwal, Zeitler, Mastouri,
  and Sch{\"o}lkopf}{von K{\"u}gelgen et~al\mbox{.}}{2021}]%
        {bernhard-causal-cfe-confounded}
\bibfield{author}{\bibinfo{person}{J. von K{\"u}gelgen}, \bibinfo{person}{N.
  Agarwal}, \bibinfo{person}{J. Zeitler}, \bibinfo{person}{A. Mastouri}, {and}
  \bibinfo{person}{B. Sch{\"o}lkopf}.} \bibinfo{year}{2021}\natexlab{}.
\newblock \showarticletitle{Algorithmic recourse in partially and fully
  confounded settings through bounding counterfactual effects}. In
  \bibinfo{booktitle}{\emph{ICML 2021 Workshop on Algorithmic Recourse}}.
\newblock
\urldef\tempurl%
\url{https://sites.google.com/view/recourse21/home}
\showURL{%
\tempurl}


\bibitem[\protect\citeauthoryear{von K{\"u}gelgen, Bhatt, Karimi, Valera,
  Weller, and Scholkopf}{von K{\"u}gelgen et~al\mbox{.}}{2020}]%
        {vonKgelgen2020EqualRecourse2}
\bibfield{author}{\bibinfo{person}{Julius von K{\"u}gelgen},
  \bibinfo{person}{Umang Bhatt}, \bibinfo{person}{Amir-Hossein Karimi},
  \bibinfo{person}{Isabel Valera}, \bibinfo{person}{Adrian Weller}, {and}
  \bibinfo{person}{Bernhard Scholkopf}.} \bibinfo{year}{2020}\natexlab{}.
\newblock \bibinfo{title}{On the Fairness of Causal Algorithmic Recourse}.
\newblock
\newblock


\bibitem[\protect\citeauthoryear{Wachter, Mittelstadt, and Floridi}{Wachter
  et~al\mbox{.}}{2017a}]%
        {right-to-explanation-gdpr}
\bibfield{author}{\bibinfo{person}{Sandra Wachter}, \bibinfo{person}{Brent
  Mittelstadt}, {and} \bibinfo{person}{Luciano Floridi}.}
  \bibinfo{year}{2017}\natexlab{a}.
\newblock \showarticletitle{{Why a Right to Explanation of Automated
  Decision-Making Does Not Exist in the General Data Protection Regulation}}.
\newblock \bibinfo{journal}{\emph{International Data Privacy Law}}
  \bibinfo{volume}{7}, \bibinfo{number}{2} (\bibinfo{date}{06}
  \bibinfo{year}{2017}).
\newblock
\urldef\tempurl%
\url{https://doi.org/10.1093/idpl/ipx005}
\showDOI{\tempurl}


\bibitem[\protect\citeauthoryear{Wachter, Mittelstadt, and Russell}{Wachter
  et~al\mbox{.}}{2017b}]%
        {wachter_counterfactual_2017}
\bibfield{author}{\bibinfo{person}{Sandra Wachter}, \bibinfo{person}{Brent
  Mittelstadt}, {and} \bibinfo{person}{Chris Russell}.}
  \bibinfo{year}{2017}\natexlab{b}.
\newblock \showarticletitle{Counterfactual {Explanations} {Without} {Opening}
  the {Black} {Box}: {Automated} {Decisions} and the {GDPR}}.
\newblock \bibinfo{journal}{\emph{SSRN Electronic Journal}}
  \bibinfo{volume}{31}, \bibinfo{number}{2} (\bibinfo{year}{2017}).
\newblock
\urldef\tempurl%
\url{https://doi.org/10.2139/ssrn.3063289}
\showDOI{\tempurl}


\bibitem[\protect\citeauthoryear{Wang and Vasconcelos}{Wang and
  Vasconcelos}{2020}]%
        {SCOUT_cfe_images}
\bibfield{author}{\bibinfo{person}{Pei Wang} {and} \bibinfo{person}{Nuno
  Vasconcelos}.} \bibinfo{year}{2020}\natexlab{}.
\newblock \showarticletitle{SCOUT: Self-Aware Discriminant Counterfactual
  Explanations}. In \bibinfo{booktitle}{\emph{The IEEE/CVF Conference on
  Computer Vision and Pattern Recognition (CVPR)}}.
\newblock


\bibitem[\protect\citeauthoryear{Wang, Peng, Lu, Lu, Bagheri, and Summers}{Wang
  et~al\mbox{.}}{2017}]%
        {chest-xray-data}
\bibfield{author}{\bibinfo{person}{Xiaosong Wang}, \bibinfo{person}{Yifan
  Peng}, \bibinfo{person}{Le Lu}, \bibinfo{person}{Zhiyong Lu},
  \bibinfo{person}{Mohammadhadi Bagheri}, {and} \bibinfo{person}{Ronald~M.
  Summers}.} \bibinfo{year}{2017}\natexlab{}.
\newblock \showarticletitle{ChestX-ray8: Hospital-Scale Chest X-Ray Database
  and Benchmarks on Weakly-Supervised Classification and Localization of Common
  Thorax Diseases}. In \bibinfo{booktitle}{\emph{Proceedings of the IEEE
  Conference on Computer Vision and Pattern Recognition (CVPR)}}.
\newblock


\bibitem[\protect\citeauthoryear{Wang, Ding, Wang, Liu, Wu, Wang, Liu, and
  Miao}{Wang et~al\mbox{.}}{2021a}]%
        {skyline-cfe-interactive}
\bibfield{author}{\bibinfo{person}{Yongjie Wang}, \bibinfo{person}{Qinxu Ding},
  \bibinfo{person}{Ke Wang}, \bibinfo{person}{Yue Liu}, \bibinfo{person}{Xingyu
  Wu}, \bibinfo{person}{Jinglong Wang}, \bibinfo{person}{Yong Liu}, {and}
  \bibinfo{person}{Chunyan Miao}.} \bibinfo{year}{2021}\natexlab{a}.
\newblock \showarticletitle{The Skyline of Counterfactual Explanations for
  Machine Learning Decision Models}. In \bibinfo{booktitle}{\emph{Proceedings
  of the 30th ACM International Conference on Information \& Knowledge
  Management}}. \bibinfo{publisher}{Association for Computing Machinery},
  \bibinfo{address}{New York, NY, USA}, 10.
\newblock
\urldef\tempurl%
\url{https://doi.org/10.1145/3459637.3482397}
\showDOI{\tempurl}


\bibitem[\protect\citeauthoryear{Wang, Qian, and Miao}{Wang
  et~al\mbox{.}}{2022}]%
        {DualCF-model-extraction}
\bibfield{author}{\bibinfo{person}{Yongjie Wang}, \bibinfo{person}{Hangwei
  Qian}, {and} \bibinfo{person}{Chunyan Miao}.}
  \bibinfo{year}{2022}\natexlab{}.
\newblock \showarticletitle{DualCF: Efficient Model Extraction Attack from
  Counterfactual Explanations}. In \bibinfo{booktitle}{\emph{2022 ACM
  Conference on Fairness, Accountability, and Transparency}}
  \emph{(\bibinfo{series}{FAccT '22})}. \bibinfo{publisher}{Association for
  Computing Machinery}, \bibinfo{address}{New York, NY, USA}, 12.
\newblock
\urldef\tempurl%
\url{https://doi.org/10.1145/3531146.3533188}
\showDOI{\tempurl}


\bibitem[\protect\citeauthoryear{Wang, Samsten, Mochaourab, and
  Papapetrou}{Wang et~al\mbox{.}}{2021c}]%
        {wang_cfe-time-series-latentcf++}
\bibfield{author}{\bibinfo{person}{Zhendong Wang}, \bibinfo{person}{Isak
  Samsten}, \bibinfo{person}{Rami Mochaourab}, {and}
  \bibinfo{person}{Panagiotis Papapetrou}.} \bibinfo{year}{2021}\natexlab{c}.
\newblock \showarticletitle{Learning Time Series Counterfactuals via Latent
  Space Representations}. In \bibinfo{booktitle}{\emph{Discovery Science}}.
  \bibinfo{publisher}{Springer International Publishing},
  \bibinfo{address}{Cham}, \bibinfo{pages}{369--384}.
\newblock


\bibitem[\protect\citeauthoryear{Wang, Samsten, and Papapetrou}{Wang
  et~al\mbox{.}}{2021b}]%
        {wang-cfe-time-series-ICU-cardio}
\bibfield{author}{\bibinfo{person}{Zhendong Wang}, \bibinfo{person}{Isak
  Samsten}, {and} \bibinfo{person}{Panagiotis Papapetrou}.}
  \bibinfo{year}{2021}\natexlab{b}.
\newblock \showarticletitle{Counterfactual Explanations for Survival Prediction
  of Cardiovascular ICU Patients}. In \bibinfo{booktitle}{\emph{Artificial
  Intelligence in Medicine}}. \bibinfo{publisher}{Springer International
  Publishing}, \bibinfo{address}{Cham}, \bibinfo{pages}{338--348}.
\newblock
\showISBNx{978-3-030-77211-6}


\bibitem[\protect\citeauthoryear{Warren, Keane, and Byrne}{Warren
  et~al\mbox{.}}{2022}]%
        {user-study-cfe-causal-diff-keane}
\bibfield{author}{\bibinfo{person}{Greta Warren}, \bibinfo{person}{Mark~T
  Keane}, {and} \bibinfo{person}{Ruth M~J Byrne}.}
  \bibinfo{year}{2022}\natexlab{}.
\newblock \bibinfo{title}{Features of Explainability: How users understand
  counterfactual and causal explanations for categorical and continuous
  features in XAI}.
\newblock
\newblock
\urldef\tempurl%
\url{https://doi.org/10.48550/ARXIV.2204.10152}
\showDOI{\tempurl}


\bibitem[\protect\citeauthoryear{Wellawatte, Seshadri, and White}{Wellawatte
  et~al\mbox{.}}{2022}]%
        {CF-GNN5}
\bibfield{author}{\bibinfo{person}{Geemi~P. Wellawatte}, \bibinfo{person}{Aditi
  Seshadri}, {and} \bibinfo{person}{Andrew~D. White}.}
  \bibinfo{year}{2022}\natexlab{}.
\newblock \showarticletitle{Model agnostic generation of counterfactual
  explanations for molecules}.
\newblock \bibinfo{journal}{\emph{Chem. Sci.}}  \bibinfo{volume}{13}
  (\bibinfo{year}{2022}), \bibinfo{pages}{3697--3705}.
\newblock
\urldef\tempurl%
\url{https://doi.org/10.1039/D1SC05259D}
\showDOI{\tempurl}


\bibitem[\protect\citeauthoryear{{Wexler}, {Pushkarna}, {Bolukbasi},
  {Wattenberg}, {Viégas}, and {Wilson}}{{Wexler} et~al\mbox{.}}{2020}]%
        {what-if-tool}
\bibfield{author}{\bibinfo{person}{J. {Wexler}}, \bibinfo{person}{M.
  {Pushkarna}}, \bibinfo{person}{T. {Bolukbasi}}, \bibinfo{person}{M.
  {Wattenberg}}, \bibinfo{person}{F. {Viégas}}, {and} \bibinfo{person}{J.
  {Wilson}}.} \bibinfo{year}{2020}\natexlab{}.
\newblock \showarticletitle{The What-If Tool: Interactive Probing of Machine
  Learning Models}.
\newblock \bibinfo{journal}{\emph{IEEE Transactions on Visualization and
  Computer Graphics}} \bibinfo{volume}{26}, \bibinfo{number}{1}
  (\bibinfo{year}{2020}), \bibinfo{pages}{56--65}.
\newblock


\bibitem[\protect\citeauthoryear{White and Garcez}{White and Garcez}{2019}]%
        {white_measurable_2019}
\bibfield{author}{\bibinfo{person}{Adam White} {and}
  \bibinfo{person}{Artur~d'Avila Garcez}.} \bibinfo{year}{2019}\natexlab{}.
\newblock \bibinfo{title}{Measurable {Counterfactual} {Local} {Explanations}
  for {Any} {Classifier}}.
\newblock
\newblock
\urldef\tempurl%
\url{http://arxiv.org/abs/1908.03020}
\showURL{%
\tempurl}


\bibitem[\protect\citeauthoryear{White and Garcez}{White and Garcez}{2021}]%
        {white_measurable_2021_supporting}
\bibfield{author}{\bibinfo{person}{Adam White} {and}
  \bibinfo{person}{Artur~d'Avila Garcez}.} \bibinfo{year}{2021}\natexlab{}.
\newblock \bibinfo{title}{Counterfactual Instances Explain Little}.
\newblock
\newblock
\urldef\tempurl%
\url{https://doi.org/10.48550/ARXIV.2109.09809}
\showDOI{\tempurl}


\bibitem[\protect\citeauthoryear{White, Ngan, Phelan, Afgeh, Ryan,
  Reyes-Aldasoro, and Garcez}{White et~al\mbox{.}}{2021}]%
        {contrastive-cfe-images-CLEAR}
\bibfield{author}{\bibinfo{person}{Adam White}, \bibinfo{person}{Kwun~Ho Ngan},
  \bibinfo{person}{James Phelan}, \bibinfo{person}{Saman~Sadeghi Afgeh},
  \bibinfo{person}{Kevin Ryan}, \bibinfo{person}{Constantino~Carlos
  Reyes-Aldasoro}, {and} \bibinfo{person}{Artur~d'Avila Garcez}.}
  \bibinfo{year}{2021}\natexlab{}.
\newblock \bibinfo{title}{Contrastive Counterfactual Visual Explanations With
  Overdetermination}.
\newblock
\newblock
\urldef\tempurl%
\url{https://doi.org/10.48550/ARXIV.2106.14556}
\showDOI{\tempurl}


\bibitem[\protect\citeauthoryear{Wijekoon, Wiratunga, Nkisi-Orji, Martin,
  Palihawadana, and Corsar}{Wijekoon et~al\mbox{.}}{2021}]%
        {student-moodle-cfe-cbr-technique}
\bibfield{author}{\bibinfo{person}{Anjana Wijekoon}, \bibinfo{person}{Nirmalie
  Wiratunga}, \bibinfo{person}{Ikechukwu Nkisi-Orji}, \bibinfo{person}{Kyle
  Martin}, \bibinfo{person}{Chamath Palihawadana}, {and} \bibinfo{person}{David
  Corsar}.} \bibinfo{year}{2021}\natexlab{}.
\newblock \showarticletitle{Counterfactual explanations for student outcome
  prediction with Moodle footprints.} \bibinfo{publisher}{CEUR Workshop
  Proceedings}, \bibinfo{pages}{1--8}.
\newblock
\urldef\tempurl%
\url{https://rgu-repository.worktribe.com/output/1395861}
\showURL{%
\tempurl}


\bibitem[\protect\citeauthoryear{Wiratunga, Wijekoon, Nkisi-Orji, Martin,
  Palihawadana, and Corsar}{Wiratunga et~al\mbox{.}}{2021}]%
        {DisCERN-use-shap-for-cfe}
\bibfield{author}{\bibinfo{person}{Nirmalie Wiratunga}, \bibinfo{person}{Anjana
  Wijekoon}, \bibinfo{person}{Ikechukwu Nkisi-Orji}, \bibinfo{person}{Kyle
  Martin}, \bibinfo{person}{Chamath Palihawadana}, {and} \bibinfo{person}{David
  Corsar}.} \bibinfo{year}{2021}\natexlab{}.
\newblock \showarticletitle{DisCERN: Discovering Counterfactual Explanations
  using Relevance Features from Neighbourhoods}. In
  \bibinfo{booktitle}{\emph{2021 IEEE 33rd International Conference on Tools
  with Artificial Intelligence (ICTAI)}}. \bibinfo{pages}{1466--1473}.
\newblock
\urldef\tempurl%
\url{https://doi.org/10.1109/ICTAI52525.2021.00233}
\showDOI{\tempurl}


\bibitem[\protect\citeauthoryear{Woodward}{Woodward}{2003}]%
        {Woodward2003:phil4}
\bibfield{author}{\bibinfo{person}{James Woodward}.}
  \bibinfo{year}{2003}\natexlab{}.
\newblock \bibinfo{booktitle}{\emph{Making Things Happen: A Theory of Causal
  Explanation}}.
\newblock \bibinfo{publisher}{Oxford University Press}.
\newblock


\bibitem[\protect\citeauthoryear{Xiang and Lenskiy}{Xiang and Lenskiy}{2022}]%
        {Xiang2022-Realistic-VAE-CFE}
\bibfield{author}{\bibinfo{person}{Xintao Xiang} {and} \bibinfo{person}{Artem
  Lenskiy}.} \bibinfo{year}{2022}\natexlab{}.
\newblock \bibinfo{title}{Realistic Counterfactual Explanations by Learned
  Relations}.
\newblock
\newblock


\bibitem[\protect\citeauthoryear{Xu, Li, Liu, Fu, and Zhang}{Xu
  et~al\mbox{.}}{2020}]%
        {cfe-reco-approach5}
\bibfield{author}{\bibinfo{person}{Shuyuan Xu}, \bibinfo{person}{Yunqi Li},
  \bibinfo{person}{Shuchang Liu}, \bibinfo{person}{Zuohui Fu}, {and}
  \bibinfo{person}{Yongfeng Zhang}.} \bibinfo{year}{2020}\natexlab{}.
\newblock \bibinfo{title}{Learning Post-Hoc Causal Explanations for
  Recommendation}.
\newblock
\newblock


\bibitem[\protect\citeauthoryear{Yacoby, Green, Griffin, and Velez}{Yacoby
  et~al\mbox{.}}{2022}]%
        {CFE-user-study-judges}
\bibfield{author}{\bibinfo{person}{Yaniv Yacoby}, \bibinfo{person}{Ben Green},
  \bibinfo{person}{Christopher~L. Griffin}, {and} \bibinfo{person}{Finale~Doshi
  Velez}.} \bibinfo{year}{2022}\natexlab{}.
\newblock \bibinfo{title}{"If it didn't happen, why would I change my
  decision?": How Judges Respond to Counterfactual Explanations for the Public
  Safety Assessment}.
\newblock
\newblock
\urldef\tempurl%
\url{https://doi.org/10.48550/ARXIV.2205.05424}
\showDOI{\tempurl}


\bibitem[\protect\citeauthoryear{Yadav, Hase, and Bansal}{Yadav
  et~al\mbox{.}}{2021}]%
        {user-specific-cost}
\bibfield{author}{\bibinfo{person}{Prateek Yadav}, \bibinfo{person}{Peter
  Hase}, {and} \bibinfo{person}{Mohit Bansal}.}
  \bibinfo{year}{2021}\natexlab{}.
\newblock \bibinfo{title}{Low-Cost Algorithmic Recourse for Users With
  Uncertain Cost Functions}.
\newblock
\newblock
\urldef\tempurl%
\url{https://doi.org/10.48550/ARXIV.2111.01235}
\showDOI{\tempurl}


\bibitem[\protect\citeauthoryear{Yang, Alva, Chen, and Hu}{Yang
  et~al\mbox{.}}{2021a}]%
        {gan_cfe_amortized}
\bibfield{author}{\bibinfo{person}{Fan Yang}, \bibinfo{person}{Sahan~Suresh
  Alva}, \bibinfo{person}{Jiahao Chen}, {and} \bibinfo{person}{Xia Hu}.}
  \bibinfo{year}{2021}\natexlab{a}.
\newblock \showarticletitle{Model-Based Counterfactual Synthesizer for
  Interpretation}. In \bibinfo{booktitle}{\emph{Proceedings of the 27th ACM
  SIGKDD Conference on Knowledge Discovery \& Data Mining}}
  \emph{(\bibinfo{series}{KDD '21})}. \bibinfo{publisher}{Association for
  Computing Machinery}, \bibinfo{address}{New York, NY, USA},
  \bibinfo{pages}{1964--1974}.
\newblock
\urldef\tempurl%
\url{https://doi.org/10.1145/3447548.3467333}
\showDOI{\tempurl}


\bibitem[\protect\citeauthoryear{Yang, Liu, Du, and Hu}{Yang
  et~al\mbox{.}}{2021b}]%
        {yang-attribute-perturbation-text+images}
\bibfield{author}{\bibinfo{person}{Fan Yang}, \bibinfo{person}{Ninghao Liu},
  \bibinfo{person}{Mengnan Du}, {and} \bibinfo{person}{Xia Hu}.}
  \bibinfo{year}{2021}\natexlab{b}.
\newblock \showarticletitle{Generative Counterfactuals for Neural Networks via
  Attribute-Informed Perturbation}.
\newblock \bibinfo{journal}{\emph{SIGKDD Explor. Newsl.}}  \bibinfo{volume}{23}
  (\bibinfo{date}{may} \bibinfo{year}{2021}), 10.
\newblock
\urldef\tempurl%
\url{https://doi.org/10.1145/3468507.3468517}
\showDOI{\tempurl}


\bibitem[\protect\citeauthoryear{Yang, Kenny, Ng, Yang, Smyth, and Dong}{Yang
  et~al\mbox{.}}{2020}]%
        {yang-cfe-financial-text}
\bibfield{author}{\bibinfo{person}{Linyi Yang}, \bibinfo{person}{Eoin Kenny},
  \bibinfo{person}{Tin Lok~James Ng}, \bibinfo{person}{Yi Yang},
  \bibinfo{person}{Barry Smyth}, {and} \bibinfo{person}{Ruihai Dong}.}
  \bibinfo{year}{2020}\natexlab{}.
\newblock \showarticletitle{Generating Plausible Counterfactual Explanations
  for Deep Transformers in Financial Text Classification}. In
  \bibinfo{booktitle}{\emph{Proceedings of the 28th International Conference on
  Computational Linguistics}}. \bibinfo{publisher}{International Committee on
  Computational Linguistics}, \bibinfo{address}{Barcelona, Spain (Online)},
  \bibinfo{pages}{6150--6160}.
\newblock
\urldef\tempurl%
\url{https://doi.org/10.18653/v1/2020.coling-main.541}
\showDOI{\tempurl}


\bibitem[\protect\citeauthoryear{Yang, Kang, and Jung}{Yang
  et~al\mbox{.}}{2022}]%
        {cfe-text+images-DiscGrad}
\bibfield{author}{\bibinfo{person}{Nakyeong Yang}, \bibinfo{person}{Taegwan
  Kang}, {and} \bibinfo{person}{Kyomin Jung}.} \bibinfo{year}{2022}\natexlab{}.
\newblock \showarticletitle{Deriving Explainable Discriminative Attributes
  Using Confusion About Counterfactual Class}. In
  \bibinfo{booktitle}{\emph{ICASSP 2022}}. \bibinfo{pages}{1730--1734}.
\newblock
\urldef\tempurl%
\url{https://doi.org/10.1109/ICASSP43922.2022.9747693}
\showDOI{\tempurl}


\bibitem[\protect\citeauthoryear{Yao, Wang, and Li}{Yao et~al\mbox{.}}{2022}]%
        {yao-cfe-evaluate-recsys}
\bibfield{author}{\bibinfo{person}{Yuanshun Yao}, \bibinfo{person}{Chong Wang},
  {and} \bibinfo{person}{Hang Li}.} \bibinfo{year}{2022}\natexlab{}.
\newblock \bibinfo{title}{Counterfactually Evaluating Explanations in
  Recommender Systems}.
\newblock
\newblock
\urldef\tempurl%
\url{https://doi.org/10.48550/ARXIV.2203.01310}
\showDOI{\tempurl}


\bibitem[\protect\citeauthoryear{Yousefzadeh and O’Leary}{Yousefzadeh and
  O’Leary}{2019}]%
        {debugging-ml-models-cfe}
\bibfield{author}{\bibinfo{person}{Roozbeh Yousefzadeh} {and}
  \bibinfo{person}{Dianne~P. O’Leary}.} \bibinfo{year}{2019}\natexlab{}.
\newblock \bibinfo{title}{DEBUGGING TRAINED MACHINE LEARNING MODELS USING FLIP
  POINTS}.
\newblock
\newblock
\urldef\tempurl%
\url{https://debug-ml-iclr2019.github.io/cameraready/DebugML-19_paper_11.pdf}
\showURL{%
\tempurl}


\bibitem[\protect\citeauthoryear{Yuan, Zhu, Zhang, Huang, Ye, and Xiong}{Yuan
  et~al\mbox{.}}{2021}]%
        {CFE-earning-call-data-augment}
\bibfield{author}{\bibinfo{person}{Zixuan Yuan}, \bibinfo{person}{Yada Zhu},
  \bibinfo{person}{Wei Zhang}, \bibinfo{person}{Ziming Huang},
  \bibinfo{person}{Guangnan Ye}, {and} \bibinfo{person}{Hui Xiong}.}
  \bibinfo{year}{2021}\natexlab{}.
\newblock \bibinfo{title}{Multi-Domain Transformer-Based Counterfactual
  Augmentation for Earnings Call Analysis}.
\newblock
\newblock
\urldef\tempurl%
\url{https://doi.org/10.48550/ARXIV.2112.00963}
\showDOI{\tempurl}


\bibitem[\protect\citeauthoryear{Zhang and Lim}{Zhang and Lim}{2022}]%
        {Zhang-emotion-vocal-recognition-cfe}
\bibfield{author}{\bibinfo{person}{Wencan Zhang} {and} \bibinfo{person}{Brian~Y
  Lim}.} \bibinfo{year}{2022}\natexlab{}.
\newblock \showarticletitle{Towards Relatable Explainable {AI} with the
  Perceptual Process}. \bibinfo{publisher}{{ACM}}.
\newblock
\urldef\tempurl%
\url{https://doi.org/10.1145/3491102.3501826}
\showDOI{\tempurl}


\bibitem[\protect\citeauthoryear{Zhang., McAreavey., and Liu.}{Zhang.
  et~al\mbox{.}}{2022}]%
        {Zhang-CFE-user-study}
\bibfield{author}{\bibinfo{person}{Yuhao Zhang.}, \bibinfo{person}{Kevin
  McAreavey.}, {and} \bibinfo{person}{Weiru Liu.}}
  \bibinfo{year}{2022}\natexlab{}.
\newblock \showarticletitle{Developing and Experimenting on Approaches to
  Explainability in AI Systems}. In \bibinfo{booktitle}{\emph{Proceedings of
  the 14th International Conference on Agents and Artificial Intelligence -
  Volume 2: ICAART,}}. INSTICC, \bibinfo{publisher}{SciTePress},
  \bibinfo{pages}{518--527}.
\newblock
\urldef\tempurl%
\url{https://doi.org/10.5220/0010900300003116}
\showDOI{\tempurl}


\bibitem[\protect\citeauthoryear{Zhao}{Zhao}{2020}]%
        {fast_cfe_images}
\bibfield{author}{\bibinfo{person}{Yunxia Zhao}.}
  \bibinfo{year}{2020}\natexlab{}.
\newblock \bibinfo{title}{Fast Real-time Counterfactual Explanations}.
\newblock
\newblock
\urldef\tempurl%
\url{https://doi.org/10.48550/ARXIV.2007.05684}
\showDOI{\tempurl}


\bibitem[\protect\citeauthoryear{Zhong and Negre}{Zhong and Negre}{2022}]%
        {cfe-reco-approach8}
\bibfield{author}{\bibinfo{person}{Jinfeng Zhong} {and} \bibinfo{person}{Elsa
  Negre}.} \bibinfo{year}{2022}\natexlab{}.
\newblock \showarticletitle{Shap-Enhanced Counterfactual Explanations for
  Recommendations}. In \bibinfo{booktitle}{\emph{Proceedings of the 37th
  ACM/SIGAPP Symposium on Applied Computing}}. \bibinfo{publisher}{Association
  for Computing Machinery}, \bibinfo{address}{New York, NY, USA},
  \bibinfo{pages}{1365–1372}.
\newblock
\urldef\tempurl%
\url{https://doi.org/10.1145/3477314.3507029}
\showDOI{\tempurl}


\bibitem[\protect\citeauthoryear{{Zhou}, {Khosla}, {Lapedriza}, {Oliva}, and
  {Torralba}}{{Zhou} et~al\mbox{.}}{2016}]%
        {Khosla_cam}
\bibfield{author}{\bibinfo{person}{B. {Zhou}}, \bibinfo{person}{A. {Khosla}},
  \bibinfo{person}{A. {Lapedriza}}, \bibinfo{person}{A. {Oliva}}, {and}
  \bibinfo{person}{A. {Torralba}}.} \bibinfo{year}{2016}\natexlab{}.
\newblock \showarticletitle{Learning Deep Features for Discriminative
  Localization}. In \bibinfo{booktitle}{\emph{CVPR}}.
  \bibinfo{publisher}{IEEE}, \bibinfo{address}{New York, USA},
  \bibinfo{pages}{2921--2929}.
\newblock


\bibitem[\protect\citeauthoryear{Zhou, Wang, He, and Wang}{Zhou
  et~al\mbox{.}}{2021}]%
        {cfe-reco-approach6}
\bibfield{author}{\bibinfo{person}{Yao Zhou}, \bibinfo{person}{Haonan Wang},
  \bibinfo{person}{Jingrui He}, {and} \bibinfo{person}{Haixun Wang}.}
  \bibinfo{year}{2021}\natexlab{}.
\newblock \bibinfo{title}{From Intrinsic to Counterfactual: On the
  Explainability of Contextualized Recommender Systems}.
\newblock
\newblock
\urldef\tempurl%
\url{https://doi.org/10.48550/ARXIV.2110.14844}
\showDOI{\tempurl}


\bibitem[\protect\citeauthoryear{Zien, Krämer, Sonnenburg, and Rätsch}{Zien
  et~al\mbox{.}}{2009}]%
        {Zien_exp9}
\bibfield{author}{\bibinfo{person}{Alexander Zien}, \bibinfo{person}{Nicole
  Krämer}, \bibinfo{person}{Sören Sonnenburg}, {and} \bibinfo{person}{Gunnar
  Rätsch}.} \bibinfo{year}{2009}\natexlab{}.
\newblock \showarticletitle{The Feature Importance Ranking Measure}. In
  \bibinfo{booktitle}{\emph{Machine Learning and Knowledge Discovery in
  Databases}}, Vol.~\bibinfo{volume}{5782}. \bibinfo{publisher}{Springer Berlin
  Heidelberg}, \bibinfo{address}{Berlin, Heidelberg}.
\newblock
\urldef\tempurl%
\url{https://doi.org/10.1007/978-3-642-04174-7_45}
\showDOI{\tempurl}


\end{thebibliography}
\clearpage

\appendix
\section{Full Table}
\label{sec:full-table}
Initially, we categorized the set of papers with more columns and in a much larger table. 
We selected the most critical columns and put them in \cref{tab:main-table}. 
The full table is available \href{https://docs.google.com/spreadsheets/d/1sz14O-yDBSnNalJqhIsEaVaLeQ-C5SPqAFSMErU9hGM/edit?usp=sharing}{here}. 

\section{Methodology}
\label{sec:method}
% We describe the method for our paper -- how we collected papers. 

% End date of lit review (sometime in June 2020?)

%\subsection{Transition into meat of paper}  Next, we do X then Y then Z ...

\subsection{How we collected the paper to review?}
We collected a set of \papers papers. This section provides the exact procedure used to arrive at this set of papers. 
For the first version of this survey paper, we had started from a seed set of papers recommended by other people~\cite{mahajan_preserving_2020,mothilal_explaining_2020,ramakrishnan_synthesizing_2019,Ustun19:Actionable,wachter_counterfactual_2017}, followed by snowballing their references. 
For this updated (second) version of the paper, we collected papers that cited the first paper that proposed CFEs for ML, i.e., \citet{wachter_counterfactual_2017} and the first version of this CFE survey paper~\citep{first-version-cfesurvey}. 

For an even complete search, we searched for ``counterfactual explanations'', ``recourse'', and ``inverse classification'' on two popular search engines for scholarly articles, Semantic Scholar and Google scholar. 
We looked for papers published in the last five years on both search engines. 
This is a reasonable time frame since the paper that started the discussion of counterfactual explanations in the context of machine learning (specifically for tabular data) was published in 2017~\citep{wachter_counterfactual_2017}. 
We collect papers that were published before 31st May 2022. 
The papers we collected were published at conferences like KDD, IJCAI, FAccT, AAAI, WWW, NeurIPS, WHI, or uploaded to Arxiv. %\SV{Do we want to show the statistics of the conference vs. arxiv papers (as bar plot, for example)? This kind of shows how nascent the field is. }

\subsection{Scope of the review}
Even though the first paper we reviewed was published online in 2017, and most other papers we review cite it \citep{wachter_counterfactual_2017} as the seminal paper that started the discussion around counterfactual explanations, we do not claim that this is an entirely new idea. 
Communities from data mining~\cite{fernandez-loria_explaining_2020,Provost1}, causal inference~\cite{causality:Pearl}, and even software engineering~\cite{metamorphic-testing} have explored similar ideas to identify the principal cause of a prediction, an effect, and a bug, respectively. 
Even before the emergence of counterfactual explanations in applied fields, they have been the topic of discussion in fields like social sciences~\citep{Miller-xai:2019}, philosophy~\citep{Kment:phil1, Lewis1973:phil2, Ruben2004:phil3}, psychology~\citep{Byrne:psycho1,Byrne2019:psycho2,Kahneman1986:psycho3}. 
In this review paper, we restrict our discussion to recent papers that discuss counterfactual explanations in machine learning, specifically classification settings. 
These papers have been inspired by the emerging trend of FATE and the legal requirements pertaining to explainability in tasks automated by machine learning algorithms. 
     % methodology to collect the papers
\section{Burgeoning legal frameworks around explanations in AI}
\label{sec:legal}
% Right to Explanation~\citep{wachter_counterfactual_2017}
To increase the accountability of automated decision systems---specifically, AI systems---laws and regulations regarding the decisions produced by such systems have been proposed and implemented across the globe~\citep{AI-Accountability:2017}. 
The most recent version of the European Union's General Data Protection Regulation (GDPR), enforced starting on May 25, 2018, offered a right to information about the existence, logic, and envisaged consequences of such a system~\citep{Euro-GDPR2}. This also includes the right to not be a subject of an automated decision-making system. 
Although the closeness of this law to ``right to explanation'' is debatable and ambiguous~\citep{right-to-explanation-gdpr}, the official interpretation by Working Party for Article 29 has concluded that the GDPR requires explanations of specific decisions, and therefore counterfactual explanations are apt. 
In the US, the Equal Credit Opportunity Act (ECOA) and the Fair Credit Reporting Act (FCRA) require the creditor to inform the reasons for an adverse action, such as rejection of a loan request~\citep{ECOA1,ECOA2}. They generally compare the applicant's feature to the average value in the population to arrive at the principal reasons. 
Government reports from the United Kingdom~\citep{UK-XAI1} and France~\citep{Fr-XAI1,Fr-XAI2} also touched on the issue of explainability in AI systems. 
In the US, Defense Advanced Research Projects Agency (DARPA) launched the Explainable AI (XAI) program in 2016 to encourage research into designing explainable models, understanding the psychological requirements of explanations, and the design of explanation interfaces~\citep{DARPA1}. 
The European Union has taken similar initiatives as well~\citep{EU-XAIfund1,EU-XAIfund2}. 
The US White House recently put forward the Blueprint for an AI Bill of Rights \citep{blueprints-bill-of-rights} to modulate decisions from automated systems. 
The Bill outlines five principles for operating such systems: 1) safe and effective systems, 2) algorithmic discrimination protections, 3) data privacy, 4) explanations for decisions made using such systems, and 5) discussion about human alternatives. 
While many techniques have been proposed for explainable machine learning, it is yet unclear if and how these specific techniques can help address the letter of the law. Future collaboration between AI researchers, regulators, the legal community, and consumer watchdog groups will help ensure the development of trustworthy AI.

\end{document}